%% file: arXiv.tex
\documentclass[preprint,12pt]{elsarticle}

\usepackage{epsfig}
\usepackage{helvet}
\usepackage{courier}
\usepackage{graphicx}
\usepackage{epstopdf}
\usepackage{amsfonts}
\usepackage{amsmath}
\usepackage{amsthm}
\usepackage{amssymb}
\usepackage{times}
\usepackage{latexsym} 
\usepackage{verbatim}
\usepackage[all,knot]{xy}

\usepackage{tabstackengine}
\usepackage{enumitem}
\usepackage{multirow}
\usepackage{hyperref}
\usepackage{bm}
\usepackage{footmisc}

\newcommand{\lracF}{\ensuremath{\langle\mathcal{L}, \mathcal{R}, 
 \mathcal{A}, \mathcal{C}\rangle}}
\def\sentset{\ensuremath{S}}

\newcommand{\af}{\emph{AF}}

\def\asmset{\ensuremath{\Delta}}

\def\alng{\ensuremath{\mc{L}}}
\def\lng{\ensuremath{\mc{L}^c}}
\def\sent{\ensuremath{\sigma}}
\def\sents{\ensuremath{\Sigma}}
\def\probParameter{\ensuremath{\theta}}
\def\learnRate{\ensuremath{\eta}}
\def\argA{\ensuremath{\mt{A}}}
\newcommand\argAi[1]{\ensuremath{\mt{A}_{\mt{#1}}}}
\def\argB{\ensuremath{\mt{B}}}
\newcommand\argBi[1]{\ensuremath{\mt{B}_{\mt{#1}}}}
\def\argC{\ensuremath{\mt{C}}}
\newcommand\argX[1]{\ensuremath{\mt{#1}}}
\newcommand\argXi[2]{\ensuremath{\mt{#1}_{\mt{#2}}}}
\def\args{\ensuremath{\mt{As}}}
\def\argsB{\ensuremath{\mt{Bs}}}

\def\aapd{\ensuremath{\mt{AA2PD}}}
\def\abapd{\ensuremath{\mt{ABA2PD}}}

\def\pdf{\ensuremath{\langle \mc{L}, \mc{R}\rangle}}

\newcommand\pdfs[1]{\ensuremath{\langle \mc{L}_{\mt{#1}}, \mc{R}_{\mt{#1}}\rangle}}

\def\PrC{\ensuremath{\mathrm{Pr_c}}}
\def\PrO{\ensuremath{\mathrm{Pr_o}}}

\def\AAF{\ensuremath{\langle \mc{A}, \mc{T} \rangle}}
\def\ABAF{\ensuremath{\langle \mc{L}, \mc{R}, \mc{A}, \mc{C} \rangle}}

\newcommand\argAn[1]{\ensuremath{A_{#1}}}

\DeclareMathOperator*{\argmin}{arg\,min}

\newcommand{\mc}[1]{\ensuremath{\mathcal{#1}}}
\newcommand{\mt}[1]{\ensuremath{\mathtt{#1}}}
\newcommand{\argu}[2]{#1 \vdash #2} 
\newcommand{\arguD}[2]{\ensuremath{#1 \vdash_{\mathtt{D}} #2}} 
\newcommand{\arguR}[3]{\ensuremath{#1 \vdash^{#3} #2}} 
\newcommand{\arguABA}[2]{\ensuremath{#1 \vdash_{\mathtt{A}} #2}} 
\newcommand{\prule}[2]{\ensuremath{#1}:[\ensuremath{#2}]}

\newcommand{\wrt}{with respect to}
\newcommand{\ifaf}{if and only if}
\newcommand{\respectively}{respectively}
\newcommand{\st}{such that}

\newtheorem{preex}{Example}[section]
\newenvironment{example}
{\begin{preex} \begin{rm}}
{\end{rm} \end{preex}}

\newtheorem{predfn}{Definition}[section]
\newenvironment{definition}
{\begin{predfn} \begin{rm}}
{\end{rm} \end{predfn}}

\newtheorem{prethm}{Theorem}[section]
\newenvironment{theorem}
{\begin{prethm} \begin{rm}}
{\end{rm} \end{prethm}}

\newtheorem{precol}{Corollary}[section]
\newenvironment{corollary}
{\begin{precol} \begin{rm}}
{\end{rm} \end{precol}}

\newtheorem{prelem}{Lemma}[section]
\newenvironment{lemma}
{\begin{prelem} \begin{rm}}
{\end{rm} \end{prelem}}

\newtheorem{prepro}{Proposition}[section]
\newenvironment{proposition}
{\begin{prepro} \begin{rm}}
{\end{rm} \end{prepro}}

\begin{document}

\begin{frontmatter}

\title{Probabilistic Deduction: an Approach to Probabilistic
  Structured Argumentation}

\author{Xiuyi Fan}

\address{Nanyang Technological University, Singapore}


\begin{abstract}
This paper introduces Probabilistic Deduction (PD) as an approach to
probabilistic structured argumentation. A PD framework is composed of
probabilistic rules (p-rules). As rules in classical structured
argumentation frameworks, p-rules form deduction systems. In addition,
p-rules also represent conditional probabilities that 
define joint probability distributions. With PD frameworks, one
performs probabilistic reasoning by solving Rule-Probabilistic
Satisfiability. At the same time, one can obtain an argumentative
reading to the probabilistic reasoning with arguments and attacks. In
this work, we introduce a probabilistic 
version of the Closed-World Assumption (P-CWA) and prove that
our probabilistic approach coincides with the complete extension in
classical argumentation under P-CWA and with maximum entropy
reasoning. We present several approaches to compute the joint
probability distribution from p-rules for achieving a practical proof
theory for PD. PD provides a framework to unify probabilistic
reasoning with argumentative reasoning. This is the first work in
probabilistic structured argumentation where the joint distribution is
not assumed form external sources.
\end{abstract}


\end{frontmatter}

\newpage
\tableofcontents{}

\newpage

\section{Introduction}
\label{Introduction}

The field of argumentation has been in rapid development in the past
three decades. In argumentation, information forms arguments; one
argument attacks another if the former is in conflict with the
latter. As stated in Dung's landmark paper \cite{Dung95}, {\em 
  ``whether or not a rational agent believes in a statement depends on
  whether or not the argument supporting this statement can be
  successfully defended against the counterarguments''}, 
argumentation analyses statement acceptability by studying
attack relations amongst arguments. As a reasoning paradigm in
multi-agent settings, especially for reasoning under uncertainty or
with conflict information, argumentation has seen its applications in
e.g. medical (see
e.g. \cite{Craven12,Fox07,Labrie2014,Fan13-CLIMA,Kokciyan2021,Cyras18,Gainsburg2016}),
legal (see \cite{bongiovanni2018} for an overview), and engineering
(see e.g. \cite{WilsonLopez20} for an overview) domains.

Arguments in Dung's abstract argumentation (AA) are atomic without
internal structure. Also, in AA, there is no specification of what is an
argument or an attack as these notions are assumed to be
given. To have a more detailed formalisation of arguments
than is available with AA, one turns to structured
argumentation - using some forms of logic, arguments are built
from a formal language, which serves as a representation of
information; attacks are also derived from some notion representing
conflicts in the underlying logic and language \cite{tutor}. Both
abstract argumentation and structured argumentation are seen as
powerful reasoning paradigms with extensive theoretical results and
practical applications (see e.g. \cite{Baroni2018} for an overview). 

As reasoning with probabilistic information is considered a pertinent
issue in many application areas, several different probabilistic
argumentation frameworks have been developed in the literature to join
probability with argumentation. As summarised in
Hunter~\cite{Hunter13}, two main approaches to probabilistic
argumentation exist today: the epistemic and the constellations  
approaches. Quoting Hunter \& Thimm \cite{Hunter17} on this
distinction:
\begin{quote}
  {\em In the constellations approach, the uncertainty is in the
    topology of the graph [of arguments]. \ldots
  In the epistemic approach, the topology of the argument graph is
  fixed, but there is uncertainty about whether an argument is
  believed.}
\end{quote}
In other words, in a constellations approach,
probabilities are defined over sets (extensions) of arguments,
representing the uncertainty on whether sets of arguments exist in
a given context; whereas in an epistemic approach, probabilities
are defined over arguments, representing uncertainty on whether
arguments are true. Both approaches have seen many successful
development. For instance,
\cite{Dung10,Li11,Rienstra12,Hunter12,Dondio14,Polberg14,Doder14,Sun15b,FazzingaFP16,Fazzinga16,Cyras21} 
are works taking the constellations approach; and \cite{Thimm12,Hunter14,Kafer22,Hunter20,Hunter13,Hunter17,Hunter20b,Hunter22} are works
taking the epistemic approach. 

As in non-probabilistic or classical argumentation, arguments in
probabilistic argumentation can either be atomic or structured. For
instance, amongst the works mentioned above, 
\cite{Dung10,Fazzinga16,Rienstra12,Hunter13,Hunter20,Hunter22,Cyras21} are the
ones studying structured arguments whereas the rest are the ones
studying non-structured arguments. Within the group of works that
studying probabilistic structured argumentation with the epistemic
approach, i.e. \cite{Hunter13,Hunter20,Hunter22}, it is assumed that a
probability distribution over the language is given. An implication
of this assumption is that the logic component is detached from the
probability component in the sense that one first performs logic
operations to form arguments, and then view them through a lens of
probability. In other words, there is a ``logic information''
component describing some knowledge that is used to construct
arguments; separately, there is ``probability information'' which acts
as an perspective filter to augment arguments.

\begin{table}
  \begin{center}
  \caption{Examples of Different Probabilistic Argumentation Approaches.}
  \begin{tabular}{|l|c|c|}
    \hline
    & Abstract & Structured \\
    \hline
    Constellations &
    \cite{Li11,Hunter12,Dondio14,Polberg14,Doder14,Sun15b,FazzingaFP16,Cyras21} &
    \cite{Dung10,Rienstra12,Fazzinga16}
    \\
    Epistemic &
    \cite{Thimm12,Hunter14,Kafer22,Hunter17,Hunter20b} &
    \cite{Hunter13,Hunter20,Hunter22}
    \\
    \hline    
  \end{tabular}
  \end{center}  
\end{table}

This work aims to provide an alternative approach to epistemic
probabilistic structured argumentation. Instead of assuming the
duality of logic and probability, we consider all information 
being probabilistic and represented in the form of {\em Probabilistic
  Deduction (PD)} frameworks composed of {\em probability rules
  (p-rules)}. Being the sole representation in our work, p-rules
describe both probability and logic information at the same time as
p-rules can be read as both conditional probabilities and production
rules. Instead of taking a probability distribution from some
external source,
p-rules
define probability distributions; at the same time, when reading them
as production rules, p-rules form a deduction system that can be used
to build arguments and attacks as in classical structured
argumentation.


\begin{example}
  \label{exp:opening}

  Consider a hypothetical university admission example with the
  following information.

  \begin{itemize}
    \item
    A student is likely to receive good exam scores if he studies hard. \\
    \prule{\mt{GoodExamScore} \gets \mt{HardStudy}}{0.8}

  \item
    A student is likely to receive good exam scores if he has high IQ.
    \\
    \prule{\mt{GoodExamScore} \gets \mt{HighIQ}}{0.6}

  \item
    A student is likely to be admitted to university if he has good
    exam scores.\\
    \prule{\mt{Admission} \gets \mt{GoodExamScore}}{0.7}

  \item
    A student is likely not to be admitted if he does not have
    extracurricular experience.\\
    \prule{\mt{\neg Admission} \gets \mt{\neg
        ExtraExp}}{0.7}

  \item
    A student will have extracurricular experience if he
    has both time and interest for it.\\
    \prule{\mt{ExtraExp} \gets \mt{TimeForExtraExp,
        InterestInExtraExp}}{1} 

  \item
    A student may or may not have time for extracurricular experience.\\
    \prule{\mt{TimeForExtraExp} \gets}{0.5}

  \item
    A student is likely interest in having extracurricular
    experience.\\
    \prule{\mt{InterestInExtraExp} \gets}{0.8}

  \item
    A student may or may not have high IQ.\\
    \prule{\mt{HighIQ} \gets}{0.5}

  \item
    A student will not study hard if he is lazy.\\
    \prule{\neg \mt{HardStudy}\gets\mt{Lazy}}{1}    
  \end{itemize}
  Each of these statements is represented with a probabilistic rule
  (p-rule) denoting conditional probability. For instance,
  \begin{center}
    \prule{\mt{GoodExamScore} \gets \mt{HardStudy}}{0.8} \\
    is read as\\
    $\Pr(\mt{GoodExamScore}|\mt{HardStudy}) = 0.8$; \\
    and \\
    \prule{\mt{TimeForExtraExp} \gets}{0.5}\\
    is read as\\
    $\Pr(\mt{TimeForExtraExp}) = 0.5$.
  \end{center}
  From these p-rules, we can build arguments and specify attacks using
  the approach we will describe in Section~\ref{sec:argAtt}. Some
  arguments and their attacks are shown in
  Figure~\ref{fig:openingExp}; and readings of these arguments are
  summarised in Table~\ref{table:opening}. With a
  probability calculation approach we will introduce in
  Section~\ref{sec:prules}, we compute probabilities for {\em
    literals} as follows:
  
  \begin{tabular}{ll}
    $\Pr(\mt{GoodExamScore}) = 0.744$, & $\Pr(\mt{HardStudy}) = 0.735$, \\
    $\Pr(\mt{HighIQ}) = 0.5$, & $\Pr(\mt{Admission}) = 0.521$, \\
    $\Pr(\mt{ExtraExp}) = 0.315$, & $\Pr(\mt{TimeForExtraExp}) = 0.5$, \\
    $\Pr(\mt{InterestInExtraExp}) = 0.8$, & $\Pr(\mt{Lazy}) = 0.265$.
  \end{tabular}

  \noindent
  With the PD framework we will introduce in Section~\ref{sec:argAtt},
  we compute {\em arguments} probabilities as follows.
  
  \begin{tabular}{llll}
    $\Pr(\argX{A}) = 0.411$, & $\Pr(\argX{B}) = 0.201$, & $\Pr(\argX{C}) = 0.5$, &
    $\Pr(\argX{D}) = 0.315$.
  \end{tabular}

  \begin{figure}
    \begin{small}
    \[\xymatrix@1@C=0pt@R=20pt{
      *+[F]{\argX{A} = \argu{\{\mt{HS,GES,Adm}\}}{\mt{Adm}}} & &
      *+[F]{\argX{B} = \argu{\{\mt{HIQ,GES,Adm}\}}{\mt{Adm}}} \\
      &*+[F]{\argX{C} = \argu{\{\neg \mt{EE}, \neg \mt{Adm}\}}{\neg \mt{Adm}}}\ar[lu] \ar[ru]\\
      &*+[F]{\argX{D} = \argu{\{\mt{TFEE,IIEE,EE}\}}{\mt{EE}}}\ar[u] \\
    }\]
    \end{small}    
    \caption{Some arguments and attacks in
      Example~\ref{exp:opening}. In this example, \mt{HS, GES, Adm,
        HIQ, EE}, \mt{TFEE}, \mt{IIEE} are short hands for
      \mt{HardStudy, GoodExamScore, Admission}, \mt{High IQ,
        ExtraExp}, \mt{TimeForExtraExp}, \mt{InterestInExtraExp},
      \respectively.\label{fig:openingExp}} 
  \end{figure}
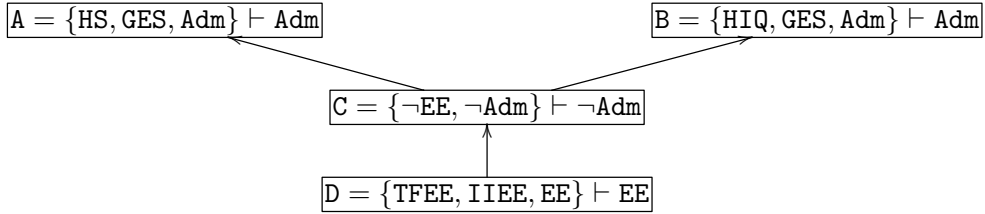

  \begin{table}
    \begin{small}
    \begin{center}
    \caption{Arguments and their readings in
      Figure~\ref{fig:openingExp}.\label{table:opening}} 
    \begin{tabular}{|c|l|}
      \hline
      Argument & Reading \\
      \hline
      \multirow{2}*{\argX{A}} & With hard study, a student will score well in
                 exams.\\
                 & Thus they will be admitted to university. \\
      \hline
      \multirow{2}*{\argX{B}} & With high IQ, a student will score well in exams.\\
      & Thus they will be admitted to university. \\
      \hline
      \multirow{2}*{\argX{C}} & Without extracurricular experience, a student will not \\
      & be admitted to university.\\
      \hline
      \argX{D} & With time and interest, student will have
      extracurricular experience.  \\    
      \hline      
    \end{tabular}
    \end{center}
    \end{small}    
  \end{table}  
\end{example}

As illustrated in Example~\ref{exp:opening},  one can view PD as a
representation for probabilistic information supported by a well
defined probability semantics. (We will show in
Section~\ref{sec:prules} that the probability semantics is developed
from Nilssons' probabilistic satisfiability (PSAT) \cite{Nilsson86}.)
At the same time, there is an argumentative interpretation to PD
frameworks in which information can be arranged for presentation with
the notions of arguments and attacks. This spirit is inline with
contemporary approaches on argumentation for explainable 
AI (see e.g., \cite{Cyras0ABT21} for a survey). In these works,
there is a ``computational layer'' for carrying out the
computation using any suitable (numerical) techniques, such as machine
learning or optimization, and an ``argumentation layer'' built on top
of the ``computational layer'', in which the argumentation layer is
responsible for producing explanations with argumentation notions such
as arguments and attacks. A key property of our work, as we
will show in Section~\ref{sec:argAtt}, is that the two layers
reconcile with each other in the sense that when there is no
uncertainty with p-rules, i.e. when all p-rules derived from classical
argumentation having probability 1, the probability computation
coincides with the complete semantics \cite{Dung95} in classical
abstract argumentation. 

As we intend to let PD frameworks to have practical value, effort has
been put into proof theories of PD as we will present in
Section~\ref{sec:solveRule-PSAT}. In a nutshell, our approach works by
viewing each p-rules as a  constraint imposed on the probability space
defined by the language. We then find a solution in the feasible
region as the  joint probability distribution. For each literal, we
define its probability as the marginal probability computed from the
joint probability distribution. For each argument, we define its
probability as the sum of probabilities of all models entailing all
literals in the argument. The core to our probability computation is
finding the joint probability distribution. To this end, we have
developed approaches using linear programming, quadratic programming
and stochastic gradient descent.

A quick summary of the rest of this paper is as follows.

We introduce the notion of probabilistic rules (p-rules) in
Section~\ref{subsec:pruleSAT}, describing its syntax and how p-rules
define joint probability distributions. We introduce 
probability computation of literals in Section~\ref{subsec:probLit}
with two key concepts, probabilistic open-world assumption
  (P-OWA) and probabilistic closed-world assumption (P-CWA),
mimicking their counterparts in non-probabilistic logic. We
introduce Maximum Entropy Reasoning in Section~\ref{subsec:MES},
which give a unique joint probability distribution on each sets of
p-rules. As maximum entropy solutions distribute probability ``as
equally as possible'', we also show an important result
(Lemma~\ref{lemma:MEomega}) that the probability of a possible
world is not zero unless zero is the only value it can take. As both
P-CWA and maximum entropy reasoning impose constraints on the joint
distribution, we clarify their relations in Section~\ref{subsec:PCWA-MES}.

In Section~\ref{sec:argAtt}, we first give an overview to AA in
Section~\ref{subsec:BKG}. We then formally introduce PD framework in
Section~\ref{subsec:PDF}, presenting definitions of arguments and
attacks in PD frameworks. We connect PD with AA in
Section~\ref{subsec:PDAA}, showing 
how AA frameworks can be mapped to PD frameworks in which all p-rules
have probability 1. We also present the result that on such mapping,
the probability semantics of PD (under P-CWA and Maximum Entropy
Reasoning) coincides with the complete extension
(Theorem~\ref{thm:probCompleteLabel}).

Section~\ref{sec:solveRule-PSAT} presents proof theories of PD
frameworks, focusing on the joint probability distribution
calculation. Section~\ref{subset:LP} gives the basic approach using 
linear programming. This is modelled after Nilsson's PSAT approach
\cite{Nilsson86,Fan22}. Section~\ref{subsec:ComputePCWA} and
\ref{subsec:ComputeMES} present approaches for computing joint
distributions under P-CWA and with maximum entropy reasoning,
respectively. The chief contributions are an algorithm that computes
P-CWA ``locally'' (Theorem~\ref{thm:localForGlobal}) and using linear
entropy in place of von Neumann entropy,
respectively. Section~\ref{subsec:SGD} presents the 
stochastic gradient descent (SGD) approach for computing joint
probabilities. As we show in the performance study in
Section~\ref{subsec:sgdImplementation}, SGD (with its GPU
implementation) is the most practical approach for reasoning with PD
frameworks. 

This paper builds upon our prior work \cite{Fan22} as follows. The
concepts of p-rules and Rule-PSAT (Section~\ref{subsec:pruleSAT}) as
well as the linear programming approach for calculating joint probability
(Section~\ref{subset:LP}) have been presented in \cite{Fan22}. In this
paper, we have significantly expanded the theoretical presentation of
\cite{Fan22} by introducing the PD framework
with a probabilistic version of the closed-world assumption and
connecting PD to existing argumentation frameworks. We have also
present several scalable techniques for calculating joint
probabilities, which pave the way to practical probabilistic
structured argumentation.

Proofs of all theoretical results are shown in \ref{sec:proof}.

\section{Probabilistic Rules}
\label{sec:prules}

In this section, we introduce probabilistic rules (p-rules) and
satisfiability as the cornerstone of this work. We introduce
probabilistic versions of closed-world and open-world assumptions and
show how probability of literals can be computed from p-rules with
these two assumptions. We introduce the concept of maximum
entropy reasoning in the context of p-rules as well.

\subsection{Probabilistic Rules and Satisfiability}
\label{subsec:pruleSAT}

Given $n$ {\em atoms} $\sent_0, \ldots, \sent_{n-1}$ forming a
language $\alng = \{\sent_0, \ldots, \sent_{n-1}\}$, we let
$\lng$ be the closure of $\alng$ under the classical negation
$\neg$ (namely if $\sent \in \alng$, then $\sent, \neg \sent \in
\lng$).\footnote{In this work, symbols $\neg$, $\wedge$, $\vee$ and
$\models$ take their standard meaning as in classical logic.} The core
representation of this work, {\em probabilistic rule (p-rule)}, 
is defined as follows.

\begin{definition}
  \label{dfn:prule}\cite{Fan22} 
Given a language $\alng$, a {\em probabilistic rule (p-rule)} is of
the form
$$\prule{\sent_0 \gets \sent_1, \ldots, \sent_k}{\probParameter}$$
for $k \geq 0$, $\sent_i \in \lng$, and $0 \leq \probParameter \leq
1$.

$\sent_0$ is referred to as the {\em head} of the p-rule, $\sent_1,
\ldots, \sent_k$ the {\em body}, and $\probParameter$ the {\em
  probability.}
\end{definition}

The p-rule in Definition~\ref{dfn:prule} states that the probability of
$\sent_0$, when $\sent_1 \ldots \sent_k$ all hold, is
$\probParameter$. In other words, this rule states that $\Pr(\sent_0 |
\sent_1, \ldots, \sent_k) = \probParameter$.
Without loss of generality, we only consider $\probParameter > 0$ in
this work. In other words, for $\prule{\sent_0 \gets \_}{0}$, one
writes $\prule{\neg \sent_0 \gets \_}{1}$.\footnote{Throughout, $\_$
stands for an anonymous variable as in Prolog.}

\begin{definition}\cite{Henderson20}
Given a language $\alng$ with $n$ atoms, the {\em Complete
  Conjunction Set (CC Set)} $\Omega$ of $\alng$ is the set of $2^n$
conjunction of literals \st{} each conjunction contains $n$ distinct
atoms.

Each $\omega \in \Omega$ is referred to as an {\em atomic
  conjunction.}
\end{definition}

$\Omega$ represent the set of all {\em possible worlds} and each
$\omega \in \Omega$ is one of them. For instance, for $\alng =
\{\sent_0, \sent_1\}$, the CC set of $\alng$ is
$$\Omega = \{\neg \sent_0 \wedge \neg \sent_1, \neg \sent_0
\wedge \sent_1, \sent_0 \wedge \neg 
\sent_1, \sent_0 \wedge \sent_1\}.$$
The four atomic conjunctions are:
$\neg \sent_0 \wedge \neg \sent_1, \neg \sent_0
\wedge \sent_1, \sent_0 \wedge \neg \sent_1,$ and $\sent_0 \wedge
\sent_1$.

\begin{definition}\cite{Fan22}
\label{dfn:consistency}  
  Given a language $\alng$ and a set of p-rules $\mc{R}$, let
  $\Omega$ be the CC set of $\alng$.
  A function $\pi : \Omega \rightarrow [0,1]$ is a {\em consistent
  probability distribution} with respect to $\mc{R}$ on $\alng$ for
  $\Omega$ \ifaf:
  \begin{enumerate}
  \item
    For all $\omega_i \in \Omega$,
    \begin{equation}
      \label{eqn:dfnq1}
      0 \leq \pi(\omega_i) \leq 1;
    \end{equation}
  \item
    It holds that:
    \begin{equation}
      \label{eqn:dfnq2}      
      \sum_{\omega_i \in \Omega} \pi(\omega_i) = 1.
    \end{equation}
  \item
    For each p-rule
    $\prule{\sent_0 \gets }{\probParameter} \in \mc{R}$, it holds that:
    \begin{equation}
      \label{eqn:dfnq3}      
      \probParameter = \sum_{\omega_i \in \Omega, \omega_i
          \models \sent_0} \pi(\omega_i).
    \end{equation}
  \item
    For each p-rule
    $\prule{\sent_0 \gets \sent_1, \ldots, \sent_k}{\probParameter}
    \in \mc{R}$, $(k > 0)$, it holds that:
    \begin{equation}
      \label{eqn:dfnq4}
      \probParameter = \frac{\sum_{\omega_i \in \Omega, \omega_i
          \models \sent_0 \wedge \ldots \wedge \sent_k}
        \pi(\omega_i)}{\sum_{\omega_i \in \Omega, \omega_i \models \sent_1
          \wedge \ldots \wedge \sent_k} \pi(\omega_i)}.
    \end{equation}    
  \end{enumerate}
\end{definition}

Our notion of consistency as given in Definition~\ref{dfn:consistency}
consists of two parts. Equations~\ref{eqn:dfnq1} and \ref{eqn:dfnq2}
assert $\pi$ being a probability distribution over the CC set of
$\alng$ with each $\pi(\omega_i)$ between 0 and 1, and the sum
of all $\pi(\omega_i)$ is 1, respectively.
Equations~\ref{eqn:dfnq3} and
\ref{eqn:dfnq4} assert that each p-rule should be viewed as defining 
conditional probabilities for which the probability of the head of the
p-rule conditioned on the body is the probability. When the body is
empty, the head is conditioned on the universe. In other words,
Equation~\ref{eqn:dfnq3} asserts $\Pr(\sent_0) = \theta$, whereas
Equation~\ref{eqn:dfnq4} asserts $\Pr(\sent_0|\sent_1,\ldots,\sent_k)
= \theta$.

\begin{example}
  \label{exp:consistency}

  Let $\alng = \{\sent_0, \sent_1\}$,
  $\mc{R} = \{\prule{\sent_0 \gets \sent_1}{\alpha};
    \prule{\sent_1  \gets}{\beta}\}$.
    The CC set is 
  $\Omega = \{ \neg \sent_0 \wedge \neg \sent_1,
  \sent_0 \wedge \neg \sent_1,
  \neg \sent_0 \wedge      \sent_1,
       \sent_0 \wedge      \sent_1
  \}$. 
  From $\prule{\sent_0 \gets \sent_1}{\alpha}$, applying
  Equation~\ref{eqn:dfnq3}, we have
  \begin{equation}
    \label{eqn:eq1}
    \alpha = \frac{\pi(\sent_0 \wedge \sent_1)}{
      \pi(\neg \sent_0 \wedge \sent_1) +
      \pi(\sent_0 \wedge \sent_1)}.
  \end{equation}
  From $\prule{\sent_1 \gets}{\beta}$, applying
  Equation~\ref{eqn:dfnq4}, we have 
  \begin{equation}
    \label{eqn:eq2}    
    \beta = \pi(\neg \sent_0 \wedge \sent_1) +
      \pi(\sent_0 \wedge \sent_1).
  \end{equation}
  Applying Equation~\ref{eqn:dfnq2} on $\Omega$, we have
  \begin{equation}
    \label{eqn:sum1}
    \pi(\neg \sent_0 \wedge \neg \sent_1) + 
    \pi(\sent_0 \wedge \neg \sent_1) + 
    \pi(\neg \sent_0 \wedge \sent_1) + 
    \pi(\sent_0 \wedge \sent_1) = 1.
  \end{equation}
  Lastly, check the inequality given in Equation~\ref{eqn:dfnq1},
  \begin{equation}
    \label{eqn:0to1}
    0 \leq     \pi(\neg \sent_0 \wedge \neg \sent_1), 
    \pi(\sent_0 \wedge \neg \sent_1),
    \pi(\neg \sent_0 \wedge \sent_1),
    \pi(\sent_0 \wedge \sent_1)
    \leq 1.
  \end{equation}
  $\pi$ is a consistent probability distribution \ifaf{}
  there is at least one solution to
  Equations~\ref{eqn:eq1}-\ref{eqn:0to1}.
\end{example}

With consistency defined, we are ready to define Rule-PSAT as follows.

\begin{definition}
  \label{dfn:Rule-PSAT} \cite{Fan22}
  The {\em Rule Probabilistic Satisfiability (Rule-PSAT)} problem is to
  determine for a set of p-rules $\mc{R}$ on a language $\alng$,
  whether there exists a consistent probability distribution for the
  CC set of $\alng$ with respect to $\mc{R}$.

  If a consistent probability distribution exists, then $\mc{R}$ is
  {\em Rule-PSAT}; otherwise, it is not.
\end{definition}

We illustrate Rule-PSAT with Example~\ref{exp:Rule-PSAT}.

\begin{example}
  \label{exp:Rule-PSAT}

  (Example~\ref{exp:consistency} continued.) To test whether $\mc{R}$
  is Rule-PSAT on $\mc{L}$, we need to solve
  Equations~\ref{eqn:eq1}-\ref{eqn:0to1} for $\pi$ as $\mc{R}$ is
  Rule-PSAT \ifaf{} a solution exists. It is easy to see that this is
  the case as:
  \begin{align*}
    \pi(\sent_0 \wedge \sent_1) &= \alpha \beta \\
    \pi(\neg \sent_0 \wedge \sent_1) &= \beta - \alpha \beta \\
    \pi(\sent_0 \wedge \neg \sent_1) + \pi(\neg \sent_0 \wedge \neg
    \sent_1) &= 1 - \beta 
  \end{align*}
  Since $0 \leq \alpha, \beta \leq 1$, we have $0 \leq \pi(\sent_0 \wedge
  \sent_1), \pi(\neg \sent_0 \wedge \sent_1) \leq 1$.
  We can let $\pi(\sent_0 \wedge \neg \sent_1) = 0$,
  $\pi(\neg \sent_0 \wedge \neg \sent_1) = 1 - \beta$ and obtain one 
  solution for $\pi$. As the system is underspecified with four
  unknowns and three equations, we have infinitely many solutions to
  $\pi(\sent_0 \wedge \neg \sent_1)$ and $\pi(\neg \sent_0 \wedge \neg
  \sent_1)$ in the range of [0, $1-\beta$].
\end{example}

The next example gives a set of p-rules that is not Rule-PSAT.
\begin{example}
  \label{exp:non-rule-PSAT}

  Let $\mc{R}$ be a set of three p-rules:
  $$\{\prule{\sent_0 \gets \sent_1}{0.9}, \prule{\sent_0 \gets}{0.8},
  \prule{\sent_1 \gets}{0.9}\}.$$
  From $\prule{\sent_0 \gets \sent_1}{0.9}$ and
  Equation~\ref{eqn:dfnq4}, we have
  \begin{equation}
    \label{eqn:exp:non-rule-PSAT:1}
    0.9 = \frac{\pi(\sent_0 \wedge \sent_1)}{\pi(\sent_0 \wedge
      \sent_1) + \pi(\neg \sent_0 \wedge \sent_1)}.
  \end{equation}
  From $\prule{\sent_1 \gets}{0.9}$, we have
  \begin{equation}
    \label{eqn:exp:non-rule-PSAT:2}
    0.9 = \pi(\sent_0 \wedge \sent_1) + \pi(\neg \sent_0 \wedge \sent_1).
  \end{equation}
  Subsitute (\ref{eqn:exp:non-rule-PSAT:2}) in
  (\ref{eqn:exp:non-rule-PSAT:1}), we have $\pi(\sent_0 \wedge \sent_1) =
  0.81$.

  From $\prule{\sent_0 \gets}{0.8}$, we have
  \begin{equation*}
    \label{eqn:exp:non-rule-PSAT:3}
    0.8 = \pi(\sent_0 \wedge \sent_1) + \pi(\sent_0 \wedge \neg \sent_1).
  \end{equation*}
  Thus, $\pi(\sent_0 \wedge \neg \sent_1) = -0.01$, which does not
  satisfy $0 \leq \pi(\omega_i) \leq 1$.
\end{example}

Note that there is no restriction imposed on the form of p-rules other
than the ones given in Definition~\ref{dfn:prule}, as
illustrated in the next two examples, Examples~\ref{exp:form} and
\ref{exp:twoRulesSameHead}, a set of p-rules can be 
consistent even if there are rules in this set forming cycles or
having two rules with the same head.

\begin{example}
  \label{exp:form}

  Consider a set of p-rules
  $$\{\prule{\sent_0 \gets \sent_1}{0.7},
  \prule{\sent_1 \gets \sent_0}{0.6},
  \prule{\sent_1 \gets }{0.5}\}.$$
  We can see that there is a cycle between $\sent_1$ and
  $\sent_0$ as one can deduce $\sent_0$ from $\sent_1$ and 
    deduce $\sent_1$ from $\sent_0$.
  However, we can still compute a (unique) solution
  for $\pi$ over the CC set of $\{\sent_0, \sent_1\}$.
  Using Equations~\ref{eqn:dfnq2} to \ref{eqn:dfnq4}, we have:
  \begin{align*}
    0.7 &= \pi(\sent_0 \wedge \sent_1) / (\pi(\sent_0 \wedge
      \sent_1) + \pi(\neg \sent_0 \wedge \sent_1)), \\
    0.6 &= \pi(\sent_0 \wedge \sent_1) / (\pi(\sent_0 \wedge
      \sent_1) + \pi(\sent_0 \wedge \neg \sent_1)), \\
    0.5 &= \pi(\sent_0 \wedge \sent_1) + \pi(\neg \sent_0 \wedge
      \sent_1), \\
    1 &= \pi(\neg \sent_0 \wedge \neg \sent_1) + \pi(\neg \sent_0 \wedge
      \sent_1) + \pi(\sent_0 \wedge \neg \sent_1) + \pi(\sent_0 \wedge
      \sent_1).
  \end{align*}
  A solution is:
  $\pi(\neg \sent_0 \wedge \neg \sent_1) = 0.27$,
  $\pi(\sent_0 \wedge \neg \sent_1) = 0.23$,
  $\pi(\neg \sent_0 \wedge \sent_1) = 0.15$,
  $\pi(\sent_0 \wedge \sent_1) = 0.35$.
\end{example}

\begin{example}
  \label{exp:twoRulesSameHead}

  Consider a set of p-rules
  $$\{
  \prule{\sent_0 \gets \sent_1}{0.6},
  \prule{\sent_0 \gets \sent_2}{0.5},
  \prule{\sent_1 \gets}{0.7},
  \prule{\sent_2 \gets}{0.6}\}.$$
  There are two p-rules with head $\sent_0$, namely,
  \begin{center}
  $\prule{\sent_0 \gets \sent_1}{0.6}$ and 
  $\prule{\sent_0 \gets \sent_2}{0.5}$.
  \end{center}
  These two p-rules have different
  bodies and probabilities. We set up
  equations as follows.\footnote{To simplify the presentation, Boolean
  values are used as shorthand for the literals. E.g., 111, 011, and 001
  denote
  $\sent_0 \wedge \sent_1 \wedge \sent_2$,
  $\neg \sent_0 \wedge \sent_1 \wedge \sent_2$, and
  $\neg \sent_0 \wedge \neg \sent_1 \wedge \sent_2$,
  respectively.}
  \begin{align*}
    0.6 &= (\pi(111) + \pi(110)) / (\pi(010) + \pi(011) + \pi(110) + \pi(111)), \\
    0.5 &= (\pi(101) + \pi(111)) / (\pi(001) + \pi(011) + \pi(101) + \pi(111)), \\
    0.7 &= \pi(010) + \pi(011) + \pi(110) + \pi(111), \\
    0.6 &= \pi(001) + \pi(011) + \pi(101) + \pi(111), \\
    1   &= \pi(000) + \pi(001) + \pi(010) + \pi(011) \\
        &+ \pi(100) + \pi(101) + \pi(110) + \pi(111).
  \end{align*}
  Solve these, a solution found
  is follows:
  
  \begin{center}
  \begin{tabular}{llll}
  $\pi(000) = 0$, & $\pi(001) = 0.02$, & $\pi(010) = 0$, & $\pi(011) = 0.28$, \\
  $\pi(100) = 0.15$, & $\pi(101) = 0.13$, & $\pi(110) = 0.25$, & $\pi(111) = 0.17$. \\
  \end{tabular}
  \end{center}
\end{example}

\subsection{Probability of Literals}
\label{subsec:probLit}

So far, we have defined probability distribution over the CC set of a
language. To discuss probabilities of literals in the language,
there are two distinct views we can take:
{\em probabilistic open-world assumptions (P-OWA)} and 
{\em probabilistic closed-world assumptions (P-CWA)}, explained as
follows.

With P-OWA, from a set of p-rules, we take the stand that:
\begin{quote}
  The probability of a literal is determined by the p-rules
  deducing the literal in conjunction with some unspecified factors
  that are not described by all p-rules that are known.
\end{quote}  
P-CWA is the opposite of the P-OWA, such that:
\begin{quote}
  The probability of a literal is determined by the known
  p-rules deducing the literal.
\end{quote}

P-OWA and P-CWA can be viewed as probabilistic counterparts to Reiter's
classic OWA and CWA \cite{Reiter77} in the following way.
\begin{itemize}
\item
  P-OWA and OWA assume that things which cannot be
    deduced from known information can still be true;
\item
  P-CWA and CWA both assume that the information available is
  ``complete'' for reasoning.
\end{itemize}
We start our discussion with P-OWA. To define literal probability
with P-OWA, we need the following result.

\begin{proposition}
\label{prop:margin}

Given a set of p-rules $\mc{R}$ over a language $\alng$, if there is
a consistent probability distribution $\pi$ for $\Omega$ with respect
to $\mc{R}$, then for any $\sent \in \lng$, it is the case that:

\begin{align}
  \label{eqn:prop1}
  \sum_{\omega_i \in \Omega, \omega_i \models \sent}
  \pi(\omega_i) &\geq 0, \\
  \label{eqn:prop2}  
  \sum_{\omega_i \in \Omega, \omega_i \models \sent} \pi(\omega_i) +
  \sum_{\omega_i \in \Omega, \omega_i \models \neg \sent}
  \pi(\omega_i) &= 1.
\end{align}    
\end{proposition}

With Proposition~\ref{prop:margin}, we can define probability of
literals under P-OWA. Given a set of p-rules $\mc{R}$, if
there is consistent probability distribution $\pi$ for $\Omega$ with
respect to $\mc{R}$, then for any $\sent \in \lng$, the probability
of $\sent$ under P-OWA is $\PrO(x)$ \st:

\begin{equation}
  \label{eqn:sentProb}
  \PrO(\sent) = \sum_{\omega_i \in \Omega, \omega_i \models \sent} \pi(\omega_i).
\end{equation}

Under P-OWA, the literal probability is as defined in
\cite{Fan22}. From a Rule-PSAT solution, which characterises a
probability distribution over the CC set, one can compute literal
probabilities by summing up $\pi(\omega_i)$. We illustrate literal
probability computation in P-OWA in the example below.

\begin{example}
  \label{exp:powa}

  Consider the set of p-rules
  $$\{\prule{\sent_0 \gets \neg \sent_1}{1}, \prule{\sent_1
    \gets}{1}\},$$
  which states that $\sent_0$ holds if $\neg \sent_1$ does, and
  $\sent_1$ holds.  
  With P-OWA, we have the following equations:
  \begin{align}
    \label{eqn:expCwaVSowa1}
    \pi(\sent_0 \wedge \neg \sent_1) / (\pi(\sent_0 \wedge \neg
    \sent_1) + \pi(\neg \sent_0 \wedge \neg \sent_1) &= 1 \\
    \label{eqn:expCwaVSowa2}    
    \pi(\sent_0 \wedge \sent_1) + \pi(\neg \sent_0 \wedge \sent_1) &=
    1 \\
    \label{eqn:expCwaVSowa3}
    \pi(\neg \sent_0 \wedge \neg \sent_1) + \pi(\neg \sent_0 \wedge
    \sent_1) + \pi(\sent_0 \wedge \neg \sent_1) + \pi(\sent_0 \wedge
    \sent_1) &= 1
  \end{align}
  Solve these, a solution to the joint distribution is
  \begin{center}
  \begin{tabular}{ll}
    $\pi(\neg \sent_0 \wedge \neg \sent_1) = 0$, &
    $\pi(\neg \sent_0 \wedge \sent_1) = 0.5$, \\
    $\pi(\sent_0 \wedge \neg \sent_1) = 0$, &
    $\pi(\sent_0 \wedge \sent_1) = 0.5$.
  \end{tabular}
  \end{center}
  Use Equation~\ref{eqn:sentProb} to calculate literal probability,
  we have 
  \begin{align*}
    \PrO(\sent_0) &= \pi(\sent_0 \wedge \sent_1) + \pi(\sent_0 \wedge
    \neg \sent_1)  = 0.5,\\
    \PrO(\sent_1) &= \pi(\sent_0 \wedge \sent_1) + \pi(\neg \sent_0
    \wedge \sent_1) = 1.
  \end{align*}
  Thus, we read these as:
  \begin{enumerate}
    \item
      $\sent_1$ holds as we have asserted with the
      p-rule $$\prule{\sent_1 \gets}{1};$$
    \item
      yet there is a 50/50 chance that $\sent_0$ holds as well,
      despite that we have the knowledge that {\em $\sent_0$
        would hold if $\sent_1$ does not,} captured with 
      the p-rule $$\prule{\sent_0 \gets \neg \sent_1}{1}.$$
  \end{enumerate}
  Thus, we see that the world is ``open'' as $\sent_0$ has a chance to
  hold, even though we have no way to deduce that with the p-rules
  we have.
\end{example}

P-CWA asserts more constraints to literal probabilities than P-OWA.
With P-CWA, the probability of a literal is determined by all ways of
deducing the literal. To define literal probability under P-OWA, we
formalize {\em deduction} with p-rules using the same notion defined
in Assumption-based Argumentation (ABA) \cite{Toni14}, as follows.

\begin{definition}
  \label{dfn:deduction}
  Given a language $\alng$ and a set of p-rules $\mc{R}$, a {\em
    deduction} for $\sent \in \lng$ with $S \subseteq \lng$, denoted
  $\arguD{S}{\sent}$, is a finite tree with nodes labelled by literals
  in $\lng$ or by $\tau$\footnote{$\tau \not\in \alng$ represents
  ``true'' and stands for the empty body of rules. In other words,
  each rule $\sent \gets$ can be interpreted as $\sent \gets \tau$ for
  the purpose of presenting deductions as trees.}, the root labelled
  by $\sent$, leaves either $\tau$ or literals in $S$, non-leaves
  $\sent$, as children, the elements of the body of some rules in
  $\mc{R}$ with head $\sent$. 
\end{definition}

With deduction defined, we can define literal probabilities under
P-CWA. Formally, let $\{\arguD{\sents_1}{\sent}, \ldots, 
\arguD{\sents_m}{\sent}\}$ be all maximal deductions for
$\sent$,\footnote{A deduction $\arguD{S}{\sent}$ is maximal when
there is no $\arguD{S'}{\sent}$ \st{} $S\subset S'$.} where 
$\sents_1 = \{\sent_1^1,\ldots, \sent_{k1}^1\}, \ldots, \sents_m =
\{\sent_1^m,\ldots, \sent_{km}^m\}$. Let 

\begin{equation}
  \label{eqn:bigS}
  S = \bigwedge\limits_{i=1}^{k1} \sent_i^1 \vee \ldots \vee
  \bigwedge\limits_{i=1}^{km} \sent_i^m.
\end{equation}

\noindent
Then,
\begin{equation}
  \label{eqn:sent_prob_cwa}
  \PrC(\sent) = \sum_{\omega_i \in \Omega, \omega_i \models \sent}
  \pi(\omega_i) = \sum_{\omega_i \in \Omega, \omega_i \models S}
  \pi(\omega_i).
\end{equation}

The difference between P-OWA and P-CWA is illustrated in the following
example.
\begin{example}
  \label{exp:cwaVSowa}
  (Example~\ref{exp:powa} continued.)
  There are maximal deductions
\begin{center}
  $\arguD{\{\neg \sent_1,\sent_0\}}{\sent_0}$ and
  $\arguD{\{\sent_1\}}{\sent_1}$
\end{center}
  for $\sent_0$ and $\sent_1$, respectively.
  With P-CWA, in addition to Equation~\ref{eqn:expCwaVSowa1} -
  \ref{eqn:expCwaVSowa3}, we have an additional constraint derived
  from $\arguD{\{\neg \sent_1,\sent_0\}}{\sent_0}$:
\begin{equation}
  \label{eqn:cwaEqn}
  \pi(\sent_0 \wedge \sent_1) + \pi(\sent_0 \wedge \neg \sent_1) =
  \pi(\sent_0 \wedge \neg \sent_1).
\end{equation}
This is the case as the LHS sums up probabilities on atomic
conjunctions that entail $\sent_0$; and the RHS is the atomic
conjunction that entails $S = \sent_0 \wedge \neg \sent_1$.
The (unique) solution to $\pi$ is 
  \begin{center}
  \begin{tabular}{ll}
    $\pi(\neg \sent_0 \wedge \neg \sent_1) = 0$, &
    $\pi(\neg \sent_0 \wedge \sent_1) = 1$, \\
    $\pi(\sent_0 \wedge \neg \sent_1) = 0$, &
    $\pi(\sent_0 \wedge \sent_1) = 0$.
  \end{tabular}
  \end{center}

\noindent  
With these, we have $\PrC(\sent_0) = 0, \PrC(\sent_1) = 1$.

Such results match with our intuition:
\begin{itemize}
  \item
    With P-CWA, we assume that the only way to obtain $\sent_0$ is by
    having $\neg \sent_1$ (with the p-rule \prule{\sent_0 \gets \neg 
  \sent_1}{1}). However, since we know $\sent_1$ without any 
    doubt (from the p-rule \prule{\sent_1 \gets}{1}), there is no room to
    believe $\neg \sent_1$; thus we cannot not deduce $\sent_0$.
  \item
    On the other hand, with P-OWA, we assume that although we can
    obtain $\sent_0$ from $\neg \sent_1$, but there are other possible
    ways of deriving $\sent_0$ that we are unaware of, thus not having
    $\neg \sent_1$ does not suggest that we can rule out a possibility
    of $\sent_0$.
\end{itemize}
\end{example}

Another way to look at P-OWA and P-CWA is from
Equation~\ref{eqn:sent_prob_cwa}. Given $\sent \in \lng$, let $S$ be
as defined in Equation~\ref{eqn:bigS}, then for any $\omega_i \in
\Omega$, $\omega_i \models \sent$ and $\omega_i \not\models S$, it
holds that: 
\begin{equation}
  \label{eqn:omega0}
  \pi(\omega_i) = 0.
\end{equation}

From Equation~\ref{eqn:omega0}, as demonstrated in
Example~\ref{exp:cwaVSowa}, it is clear that P-CWA imposes additional
  constraints on the joint distribution $\pi$. Thus, from a set of
  p-rules that is Rule-PSAT, literal probabilities under P-CWA may be
  undefined. Consider the following example.

\begin{example}
  \label{exp:cwaPSAT}

  Consider the set of p-rules
  $$R = \{\prule{\sent_0 \gets \neg \sent_1}{1}, \prule{\sent_1
    \gets}{1}, \prule{\sent_0 \gets}{0.5}\}.$$
  Clearly, $R$ is Rule-PSAT with the solution as given in
  Example~\ref{exp:powa}, computed using
  Equations~\ref{eqn:expCwaVSowa1}-\ref{eqn:expCwaVSowa3}. However,
  if Equation~\ref{eqn:cwaEqn} is added to assert P-CWA, then
  there is no solution to $\pi$. Thus, literal probabilities under
  P-CWA such as $\PrC(\sent_0)$ and $\PrC(\sent_1)$ are undefined in
  this example.
\end{example}

We introduce the concept of {\em P-CWA consistency} as follows.
\begin{definition}
  \label{dfn:cwaConsistency}

  A set of p-rules defined with a language $\alng$ is {\em P-CWA
    consistent} \ifaf{} it is consistent and for each $\sent \in
  \lng$, $0 \leq \PrC(\sent) \leq 1.$
\end{definition}

P-CWA differs from assumptions several standard assumptions made 
in handling probabilistic systems such as {\em independence} (discussed
in e.g.,~\cite{Henderson20}) or {\em mutual exclusivity} (discussed in
e.g.,~\cite{Williamson02}), as illustrated in
Example~\ref{exp:PCWAvsIndep} below.

\begin{example}
  \label{exp:PCWAvsIndep}

  Consider a p-rule:
  \begin{center}
    \prule{\sent_0 \gets \sent_1}{\probParameter}.
  \end{center}
  By the definition of p-rule, it holds that
  \begin{equation*}
    \Pr(\sent_0 | \sent_1) = \frac{\Pr(\sent_0 \wedge
      \sent_1)}{\Pr(\sent_1)} = \probParameter. 
  \end{equation*}

  \begin{itemize}
  \item  
    With P-CWA, from the deduction $\arguD{\{\sent_0,
      \sent_1\}}{\sent_0}$, we have
    $$\Pr(\sent_0) = \Pr(\sent_0 \wedge 
  \sent_1).$$ Thus, $\probParameter = \Pr(\sent_0)/\Pr(\sent_1)$.

  \item
  With the independence assumption, assuming that $\sent_0$ and
  $\sent_1$ are independent, we have $$\Pr(\sent_0 \wedge \sent_1) =
  \Pr(\sent_0)\Pr(\sent_1).$$ Thus, $\probParameter = \Pr(\sent_0)$.

  \item  
  With the mutual exclusivity assumption, assuming that $\sent_0$ and
  $\sent_1$ are mutually exclusive, we have
  $$\Pr(\sent_0 \wedge \sent_1) = 0.$$
  Thus, $\probParameter = 0$.
  \end{itemize}  
\end{example}

It is easy to see that P-CWA also differs from conditional
independence~\cite{philip79}, which is the main assumption enabling
Bayesian network~\cite{Russell2009}, as illustrated in
Example~\ref{exp:PCWAvsCInd} below.

\begin{example}
  \label{exp:PCWAvsCInd}

  Consider two p-rules:
  \begin{center}
    \prule{\sent_0 \gets \sent_1}{\probParameter_1}, \hspace{10pt}
    \prule{\sent_1 \gets \sent_2}{\probParameter_2}.
  \end{center}
  By the definition of p-rule, we have
  \begin{equation*}
    \Pr(\sent_0 | \sent_1) = \probParameter_1, 
    \Pr(\sent_1 | \sent_2) = \probParameter_2.
  \end{equation*}
  Use the chain rule, 
  \begin{equation*}
    \Pr(\sent_0 \wedge \sent_1 \wedge \sent_2) =
    \Pr(\sent_0 | \sent_1 \wedge \sent_2)\Pr(\sent_1|\sent_2)\Pr(\sent_2).
  \end{equation*}
  
  With conditional independence, assuming that $\sent_0$ and
  $\sent_2$ are conditionally independent given $\sent_1$, we have
    $\Pr(\sent_0 | \sent_1 \wedge \sent_2) = \Pr(\sent_0 | \sent_1).$
  Thus,
  \begin{equation}
    \label{eqn:indepJoint}
    \Pr(\sent_0 \wedge \sent_1 \wedge \sent_2) = \Pr(\sent_0 |
    \sent_1)\Pr(\sent_1|\sent_2)\Pr(\sent_2) = \probParameter_1
    \probParameter_2 \Pr(\sent_2). 
  \end{equation}
  
  With P-CWA, from the deduction $\arguD{\{\sent_1,
    \sent_2\}}{\sent_1}$ we have
  $$\Pr(\sent_1) = \Pr(\sent_1 \wedge \sent_2).$$

  From
  $\Pr(\sent_1|\sent_2) = \probParameter_2$, we have
  $$\frac{\Pr(\sent_1 \wedge \sent_2)}{\Pr(\sent_2)} =
  \probParameter_2.$$

  Thus, $\Pr(\sent_1) =
  \probParameter_2\Pr(\sent_2)$.
  From $\Pr(\sent_0|\sent_1) = \probParameter_1$, we have
  \begin{equation*}
    \frac{\Pr(\sent_0 \wedge \sent_1)}{\Pr(\sent_1)} =
    \frac{\Pr(\sent_0 \wedge \sent_1 \wedge \sent_2)+\Pr(\sent_0
      \wedge \sent_1 \wedge \neg
      \sent_2)}{\probParameter_2\Pr(\sent_2)} = \probParameter_1.
  \end{equation*}
  Since $\Pr(\sent_1) =
  \Pr(\sent_1 \wedge \sent_2)$,
  $$\Pr(\sent_1 \wedge \neg \sent_2) = 0.$$
  Since $\Pr(\sent_0 \wedge \sent_1 \wedge \neg
  \sent_2) \leq \Pr(\sent_1 \wedge \neg \sent_2)$, we have 
  $$\Pr(\sent_0 \wedge \sent_1 \wedge \neg \sent_2) = 0.$$
  Therefore, with P-CWA we also obtain 
  \begin{equation*}
    \Pr(\sent_0 \wedge \sent_1 \wedge \sent_2) = \probParameter_1
    \probParameter_2 \Pr(\sent_2)
  \end{equation*}
  as with the conditional independence assumption.
  
  However, with P-CWA, from the deduction $\arguD{\{\sent_0, \sent_1,
    \sent_2\}}{\sent_0}$, we also have
  $$\Pr(\sent_0) = \Pr(\sent_0 \wedge \sent_1 \wedge \sent_2),$$ 
  which infers that
  $$\Pr(\sent_0 \wedge \sent_1 \wedge \neg \sent_2) =
  \Pr(\sent_0 \wedge \neg \sent_1 \wedge \sent_2) = \Pr(\sent_0 \wedge
  \neg \sent_1 \wedge \neg \sent_2) = 0.$$
  These do not hold in general with the conditional independence
  assumption.
\end{example}

\subsection{Maximum Entropy Solutions}
\label{subsec:MES}

As we are solving systems derived from p-rules to compute the joint
distribution $\pi$, when the system is underdetermined, multiple
solutions exist, as illustrated in the next example.

\begin{example}
  \label{exp:underdertermined}

  (Example~\ref{exp:twoRulesSameHead} continued.) This example shows a
  system with five equations and eight unknowns. Thus the system is 
  underdetermined. In addition to the solution shown previously, the
  following is another solution:
  
  \begin{center}
  \begin{tabular}{llll}
    $\pi(\omega_1) = 0.14$, & $\pi(\omega_2) = 0.16$, & $\pi(\omega_3) = 0.14$, & $\pi(\omega_4) = 0.14$, \\
    $\pi(\omega_5) = 0$, & $\pi(\omega_6) = 0$, & $\pi(\omega_7) = 0.12$, & $\pi(\omega_8) = 0.3$. \\
  \end{tabular}
  \end{center}
\end{example}

If a set of p-rules $\mc{R}$ is satisfiable (or P-CWA consistent), but
the solution to $\pi$ is not unique, then the range of the probability
of any literal in $\lng$ can be found with optimization. The upper
bound of the probability of a literal $\sent \in \lng$ can be 
found by maximising $\Pr(\sent)$ as defined in
Equation~\ref{eqn:sentProb} (with P-OWA) or \ref{eqn:sent_prob_cwa}
(with P-CWA), subject to constraints given by the systems derived from 
the p-rules. The lower bound of $\Pr(\sent)$ can be found by
minimising these equations accordingly.

\begin{example}
  \label{exp:maxMin}

  (Example~\ref{exp:underdertermined} continued.) The solution shown
  in Example~\ref{exp:twoRulesSameHead} maximises $\Pr(\sent_0)$
  ($\Pr(\sent_0)=0.7$); whereas the solution in
  Example~\ref{exp:underdertermined} minimises it $(\Pr(\sent_0)=0.42).$
\end{example}

In addition to choosing a solution that maximizes or minimizes the
probability of a literal, we can also choose the solution that
maximizes the entropy of the joint distribution. The principle of
maximum entropy is commonly used in probabilistic reasoning
\cite{jaynes2003,paris1994}, including in probabilistic argumentation
as discussed in e.g., \cite{Hunter20} and \cite{Thimm12}. It states
that
\begin{quote}
{\em amongst the set of distributions that characterize the known
information equally well, the distribution with the maximum entropy
should be chosen \cite{Jaynes57}.}
\end{quote}
The entropy of a discrete
probability distribution $\{p_1, p_2,\ldots\}$ is  
\begin{equation*}
  H(p_1, p_2, \ldots) = -\sum_i p_i \log(p_i).\footnote{Throughout, we
  consider log base 2 as in standard entropy definition. Although it
  makes no difference on the choice of log bases for our discussion.}
\end{equation*}
In our context, given a language with $n$ atoms, the maximum entropy
distribution can be found by maximising 
\begin{equation}
  \label{eqn:maxEntropyObj}
  H(\pi_1,\ldots,\pi_{2^n}) = -\sum_{i=1}^{2^n} \pi(\omega_i)
  \log(\pi(\omega_i)), 
\end{equation}
subject to the system derived from p-rules.

\begin{example}
  \label{exp:maxEntropy}

  (Example~\ref{exp:maxMin} continued.) The maximum entropy
  distribution solution found in this example is:

  \begin{center}
  \begin{tabular}{llll}
  $\pi(\omega_1) = 0.058$, & $\pi(\omega_2) = 0.114$, & $\pi(\omega_3) = 0.094$, & $\pi(\omega_4) = 0.186$, \\
  $\pi(\omega_5) = 0.058$, & $\pi(\omega_6) = 0.07$, & $\pi(\omega_7) = 0.19$, & $\pi(\omega_8) = 0.23$. \\
  \end{tabular}
  \end{center}

  \noindent
  With this $\pi$, $\Pr(\sent_0) = 0.55$.
\end{example}

By the definition of Rule-PSAT and P-CWA consistency, it is easy to
see that maximum entropy solution exists and is unique for satisfiable
p-rules. Formally,

\begin{proposition}
  \label{prop:MEunique}

  Given a set of consistent p-rules, the maximum entropy solution
  $\pi^{m}$ exists and is uniquely determined. 
\end{proposition}

A maximum entropy solution is as unbiased as possible amongst all
solutions
\cite{Thimm12}. A useful result on maximum entropy solution
that we use in Section~\ref{sec:argAtt} is the following.

\begin{lemma}
  \label{lemma:MEomega}

  Given a set of consistent p-rules, for each $\omega \in \Omega$,
  consider some constant $\alpha_{\omega}$ \st{} $[0,
    \alpha_{\omega}]$ is the feasible region for $\pi(\omega)$. 
  Let $\pi^{m}$ be the maximum entropy solution. If $\alpha_{\omega} >
  0$, then $\pi^{m}(\omega) > 0$.
\end{lemma}

Lemma~\ref{lemma:MEomega} sanctions that a maximum entropy
solution asserting a non-zero probability to each $\omega \in \Omega$
if constraints given by p-rules allow such allocation. In other
words, with maximum entropy reasoning, $\pi^m(\omega)$ is 0 only if
there is no other solution $\pi(\omega)$ exists. Consequentially, with
maximum entropy solution, literal probabilities are not 0 unless
explicitly set by p-rules. Formally,

\begin{corollary}
  \label{coro:MEsent}

  Given a set of consistent p-rules, for each $\sent \in \mc{L}$,
  let $\alpha_{\sent} \in [0,1]$ be a constant \st{} 
  $\Pr(\sent) = \Pr_x(\sent) \leq \alpha_{\sent}$ (for $x \in
  \{o,c\}$). With the maximum entropy solution $\pi$, if
  $\alpha_{\sent} > 0$, then $\Pr(\sent) > 0.$
\end{corollary}

\subsection{Relation between P-CWA and Maximum Entropy Solutions}
\label{subsec:PCWA-MES}

Although both P-CWA and maximum entropy reasoning restrict the joint
probability distribution we can take on the CC set, these are
orthogonal concepts. In other words, one can choose to apply either
P-CWA, maximum entropy reasoning individually, or both at the same
time. They can all lead to different distributions. We illustrate them
with the following example.

\begin{example}
  \label{exp:pcwaVSme}

  Given a language $\alng = \{\sent_0, \sent_1, \sent_2\}$ and a set
  of two p-rules:
  \begin{center}
    $\{\prule{\sent_0 \gets \sent_1}{0.5}, \prule{\neg \sent_1
      \gets \sent_2}{0.5}\}$.
  \end{center}
  With P-OWA, to compute the joint probability
  distribution on the CC set, we set up three equations over the $2^3 = 8$ unknowns:
  \begin{align*}
    0.5 &= (\pi(110) + \pi(111))/(\pi(010) + \pi(011) + \pi(110) + \pi(111)),\\
    0.5 &= (\pi(001) + \pi(101))/(\pi(001) + \pi(011) + \pi(101) + \pi(111)),\\
    1   &= \pi(000) + \pi(001) + \pi(010) + \pi(011)  \\
        &+ \pi(100) + \pi(101) + \pi(110) + \pi(111).
  \end{align*}
  With P-CWA, we must consider two deductions:
  \begin{center}
  $\arguD{\{\sent_0, \sent_1\}}{\sent_0}$ and $\arguD{\{\neg \sent_1,
      \sent_2\}}{\neg \sent_1}$.
  \end{center}
  These assert that:
  \begin{align*}
    \sum_{\omega \in \Omega, \omega \models \sent_0} \pi(\omega) &=
    \sum_{\omega \in \Omega, \omega \models \sent_0 \wedge \sent_1}
    \pi(\omega),\\
    \sum_{\omega \in \Omega, \omega \models \neg \sent_1} \pi(\omega) &=
    \sum_{\omega \in \Omega, \omega \models \neg \sent_1 \wedge \sent_2}
    \pi(\omega),    
  \end{align*}
  which translate to 
  \begin{align*}
    \sum_{\omega \in \Omega, \omega \models \sent_0 \wedge \neg
      \sent_1} \pi(\omega) &= \pi(101) + \pi(100) = 0,\\
    \sum_{\omega \in \Omega, \omega \models \neg \sent_1 \wedge \neg
      \sent_2} \pi(\omega) &= \pi(000) + \pi(100) = 0.    
  \end{align*}

  The resulting distribution over the CC set is as summarised in
  Table~\ref{table:pcwaVSme}. Note that unlike maximum entropy
  reasoning, which gives us unique solutions, the two systems used to
  compute P-OWA and P-CWA without maximum entropy reasoning are
  underdetermined, so infinitely many solutions exist. 

  \begin{table}
    \begin{center}
    \caption{P-CWA and Maximum Entropy (ME) Solutions for the P-Rules in
      Example~\ref{exp:pcwaVSme}. \label{table:pcwaVSme}}
    \begin{tabular}{|c|cccc|}
      \hline
      \hline
      & $\pi(000)$ & $\pi(001)$ & $\pi(010)$ & $\pi(011)$ \\
      \hline
      \hline
      P-OWA without ME & 1 & 0 & 0 & 0  \\
      P-OWA with ME & 0.125 & 0.125 & 0.125 & 0.125 \\
      P-CWA without ME & 0 & 0.5 & 0 & 0.25  \\
      P-CWA with ME & 0 & 0.293 & 0.207 & 0.146\\
      \hline
      \hline
      & $\pi(100)$ & $\pi(101)$ & $\pi(110)$ & $\pi(111)$\\      
      \hline
      \hline
      P-OWA without ME & 0 & 0 & 0 & 0  \\
      P-OWA with ME & 0.125 & 0.125 & 0.125 & 0.125 \\
      P-CWA without ME & 0 & 0 & 0 & 0.25  \\
      P-CWA with ME & 0 & 0 & 0.207 & 0.146 \\      
      \hline
    \end{tabular}
    \end{center}
  \end{table}
\end{example}

In this section, we have introduced p-rule as the core building block
of this work. Its probability semantics is defined with the joint
probability distribution over the CC set of the language. We introduce
Rule-PSAT to describe consistent p-rule sets. Literal probability is
defined as the sum of probabilities of conjunctions that are models of
the literal.

We then introduce Probabilistic
Open-World and Closed-World assumptions, P-OWA and P-CWA,
respectively, modelling their counterparts in non-monotonic logic. We
show how literal probability can be computed \wrt{} both P-OWA and
P-CWA. We make a few remarks that
P-CWA differs from other common probabilistic assumptions such as
independence, mutual exclusivity and conditional independence. We
finish this section with an introduction to maximum entropy solutions
and show that when the joint distribution is computed with maximum
entropy reasoning, literals will not take 0 probability unless that is
the only solution they have.

\section{Argumentation with P-Rules}
\label{sec:argAtt}

Thus far, we have introduced p-rules as the basic building block in
probabilistic deduction. In this section, we show how probabilistic
arguments can be built with p-rules and how attacks can be defined
between arguments. To this end, we formally define {\em Probabilistic
Deduction (PD) Framework} composed of p-rules. We then show how PD
admits
Abstract Argumentation~\cite{Dung95}
as instances.

Note that in this section, we assume p-rules in discussion are
Rule-PSAT consistent. Thus there exists a consistent joint probability
distribution $\pi$ for the CC set of $\mc{L}$. We will discuss several
methods for computing joint distributions $\pi$ from a set of p-rules
in Section~\ref{sec:solveRule-PSAT}. We also assume that P-CWA can be
imposed. Thus, unless specified otherwise, we use $\Pr(\_)$ to
denote $\PrC(\_)$ in this section.

\subsection{Background: Abstract Argumentation}
\label{subsec:BKG}

We briefly review concepts from abstract argumentation (AA).

An {\em Abstract Argumentation (AA) \em frameworks} \cite{Dung95} are 
pairs $\AAF$, consisting of a set of \emph{abstract arguments},
$\mathcal{A}$, and a binary {\it attack} relation, $\mc{T}$. 
Given an AA framework $\af = \AAF$, a set of arguments (or {\em
  extension}) $E \subseteq \mc{A}$ is
\begin{itemize}
  \item
{\it admissible} (in $\af$) \ifaf{} $\forall \argA, \argB \in E$,
$(\argA, \argB) \not \in \mc{T}$ (i.e. $E$ is \emph{conflict-free})
and for any $\argA \in E$, if $(\argC, \argA) \in \mc{T}$, then there
exists some $\argB \in E$ \st{} $(\argB, \argC) \in \mc{R}$;
\item
{\em complete} \ifaf{} $E$ is admissible and contains all arguments it
defends, where $E$ {\em defends} some $\argA' \in \mc{A}$ iff $E$
attacks all arguments that attack $\argA'$.
\end{itemize}

Given $\AAF$ and a set of labels $\Lambda = \{\mt{in, out, undec}\}$, a
{\em labelling} is a total function $\mc{A} \mapsto \Lambda$. Given a
labelling on some argumentation framework,
\begin{itemize}
\item
an \mt{in}-labelled argument is said to be legally \mt{in} \ifaf{} all
its attackers are labelled \mt{out};
  
\item
an \mt{out}-labelled argument is said to be legally \mt{out} \ifaf{} it
has at least one attacker that is labelled \mt{in};

\item
an \mt{undec}-labelled argument is said to be legally \mt{undec}
\ifaf{} not all its attackers are labelled \mt{out} and 
it does not have an attacker that is labelled \mt{in}.
\end{itemize}

A {\em complete} labelling is a labelling where every \mt{in}-labelled
argument is legally \mt{in}, every \mt{out}-labelled argument is
legally \mt{out} and every \mt{undec} labelled argument is legally
\mt{undec}.
An important result that connects {\em complete extensions} and {\em
  complete labelling} is that arguments that are labelled \mt{in} with
a complete-labelling belong to a complete extension. \cite{Baroni11}

\subsection{Probabilistic Deduction Framework}
\label{subsec:PDF}

We define Probabilistic Deduction framework as a set of P-CWA
consistent p-rules constructed on a language, as follows.

\begin{definition}
  \label{dfn:PD}

  A {\em Probabilistic Deduction (PD)} framework is a pair \pdf{}
  where $\alng$ is the language, \mc{R} is a set of p-rules \st{}

  \begin{itemize}
  \item
    for all $\rho \in \mc{R}$, literals in $\rho$ are in $\lng$,
  \item
    \mc{R} is P-CWA consistent.
  \end{itemize}

\end{definition}

With a PD framework, we can build arguments as deductions. 

\begin{definition}
  \label{dfn:arg}

  Given a PD framework \pdf, an argument for $\sent \in \mc{L}$
  supported by $S \subseteq \mc{L}$, $R \subseteq \mc{R}$, denoted
  $\argu{S}{\sent}$ is \st{} there is a deduction
  $\argA = \arguD{S}{\sent}$ in which for each leaf node $N$ in
  $\argA$, either
  \begin{enumerate}
    \item
      $N$ is labelled by $\tau$, or 
    \item
      $N$ is labelled by some $\sent' \in \mc{L}$, 
      $\prule{\neg \sent' \gets \_}{\cdot} \in \mc{R}$
      and
      $|S| > 1$.
  \end{enumerate}

\end{definition}

The condition $|S| > 1$ in Definition~\ref{dfn:arg} is put in
place to remove ``negative singleton argument'' resulted from
p-rules. For instance, consider a PD framework $\pdf$ with $\alng =
\{\sent_0\}$, $\mc{R} = \{\prule{\sent_0 \gets}{1}\}.$
Without this condition, we would admit $\arguD{\{\neg \sent_0\}}{\neg
  \sent_0}$ as an argument, because 
\begin{enumerate}
  \item
    $\neg \sent_0 \in \lng$ forms a tree by itself in which the root
    and the leaf are both $\neg \sent_0$,  
  \item
    $\neg \neg \sent_0 = \sent_0$ and $\prule{\sent_0 \gets}{1}$ is a
    p-rule in $\mc{R}$.
\end{enumerate}
Intuitively, in this definition of argument, we want to assert that
\begin{enumerate}
\item
  if there is only a single literal $\sent$ in the deduction,  
  then there must exist a p-rule $\prule{\sent \gets}{\cdot}$ in the
  set of rules; otherwise, 
\item
  there must be some reason to acknowledge each leaf, either directly
  through a rule without body, or the existence of some information
  about the negation of the leaf. 
\end{enumerate}

Example~\ref{exp:arg} presents a few deductions for illustration.
\begin{example}
  \label{exp:arg}

  Consider a language $\alng = \{\sent_0, \sent_1, \sent_2, \sent_3\}$ and four
  sets of p-rules $R_1, \ldots, R_4$, where 
  \begin{itemize}
  \item
    $R_1 = \{\prule{\sent_0 \gets}{0.7}\}$,

  \item
    $R_2 = \{\prule{\sent_1 \gets}{0.8}, \prule{\sent_0 \gets \sent_1}{0.4}\}$,

  \item
    $R_3 = \{\prule{\sent_1 \gets}{0.7}, \prule{\neg \sent_2 \gets}{0.8},
    \prule{\sent_0 \gets \sent_1, \neg \sent_2}{0.8}\}$,

  \item
    $R_4 = \{\prule{\neg \sent_2 \gets \sent_3}{0.7}, \prule{\sent_1 \gets \sent_2}{0.8},
    \prule{\neg \sent_0 \gets \sent_1}{0.8}\}$.
  \end{itemize}

  Figure~\ref{fig:argExp} shows some examples of arguments built with
  these sets of p-rules.
\end{example}

\begin{figure}
  \[\xymatrix@1@C=24pt@R=12pt{
       R_1     && R_2           &&               & R_3                   &&& R_4 \\   
               && \sent_0       &&               & \sent_0               &&& \neg \sent_0\\
    \sent_0    && \sent_1\ar[u] && \sent_1\ar[ur] && \neg \sent_2\ar[ul]  &&    \sent_1\ar[u] \\
    \tau\ar[u] && \tau\ar[u]   &&  \tau\ar[u]    && \tau\ar[u]           &&    \sent_2\ar[u]
  }\]
  \caption{Argument examples: $\argu{\{\sent_0\}}{\sent_0}$ (built
    with $R_1$), $\argu{\{\sent_0, \sent_1\}}{\sent_0}$ (built with
    $R_2$), $\argu{\{\sent_0, \sent_1, \neg \sent_2\}}{\sent_0}$
    (built with $R_3$) and $\argu{\{\sent_2, \sent_1, \neg
      \sent_0\}}{\sent_0}$ (built with $R_4$) in
    Example~\ref{exp:arg}. \label{fig:argExp}}
\end{figure}

\begin{example}
  \label{exp:argNotDec}

  (Example~\ref{exp:pcwaVSme} continued.)
  With these two p-rules,
  \begin{center}
    $\{\prule{\sent_0 \gets \sent_1}{0.5}, \prule{\neg \sent_1
      \gets \sent_2}{0.5}\}$.
  \end{center}
  although both
  $\arguD{\{\sent_0,\sent_1\}}{\sent_0}$ and
  $\arguD{\{\neg \sent_1, \sent_2\}}{\neg \sent_1}$ are
  deductions; only $\argu{\{\sent_0,\sent_1\}}{\sent_0}$ is a PD
  argument. 
\end{example}

\begin{definition}
  \label{dfn:attack}
  For two arguments $\argA = \argu{\_}{\sent}$ and $\argB =
  \argu{\sents}{\_}$ in some PD framework, $\argA$ attacks $\argB$
  if $\neg \sent \in \sents$.
\end{definition}

\begin{example}
  \label{exp:attack}

  Consider a PD framework \pdf{} with
  $$\mc{R} = \{\prule{\sent_0 \gets }{0.8}, \prule{\sent_1 \gets \neg
    \sent_0}{0.9}\}.$$
  Two arguments
  $\argA = \argu{\{\sent_0\}}{\sent_0}$ and $\argB = \argu{\{\neg
    \sent_0, \sent_1\}}{\sent_1}$ can be built with 
  $\mc{R}$ \st{} $\argA$ attacks $\argB$, as illustrated in
  Figure~\ref{fig:attackExp}. 
\end{example}

\begin{figure}[b]
  \[\xymatrix@1@C=26pt@R=20pt{
    *+[F]{\argA = \argu{\{\sent_0\}}{\sent_0}} \ar[r] &
    *+[F]{\argB = \argu{\{\neg \sent_0, \sent_1\}}{\sent_1}} 
  }\]
  \caption{Argument $\argA = \argu{\{\sent_0\}}{\sent_0}$
    attacks $\argB = \argu{\{\neg \sent_0, \sent_1\}}{\sent_1}$ in
    Example~\ref{exp:attack} and
    \ref{exp:twoAttack}. \label{fig:attackExp}}
\end{figure}

At the core of PD semantics is the argument probability, defined as
follows.

\begin{definition}
  \label{dfn:argProb}

Given an argument $\argA = \argu{S}{\sent}$, in which $S =
\{s_1,\ldots,s_k\}$, the {\em probability of $\argA$} is:

\begin{equation}
  \label{eqn:argProb}
  \Pr(\argA) = \Pr(s_i \wedge \ldots \wedge s_k) = \sum_{\omega_i \in
    \Omega, \omega_i \models s_1 \wedge \ldots \wedge s_k} \pi(\omega_i).
\end{equation}
\end{definition}

\noindent
Trivially, $0 \leq \Pr(\argA) \leq 1$. We illustrate argument
probability in Example~\ref{exp:argProb}.

\begin{example}
  \label{exp:argProb}

  (Example~\ref{exp:attack} continued.)
  Consider the following joint distribution computed from the
  p-rules:
  $$\pi(00) = 0.02, \pi(01) = 0.18, \pi(10) = 0.8, \pi(11) = 0.$$
  Note that the joint distribution is unique. With P-CWA, from
  $\arguD{\{\sent_1, \neg \sent_0\}}{\sent_1}$, we have
  \begin{equation*}
    \sum_{\omega \in \Omega, \omega \models \sent_1} \pi(\omega) =
    \sum_{\omega \in \Omega, \omega \models \sent_1 \wedge \neg \sent_0} \pi(\omega).    
  \end{equation*}
  This implies $\pi(\sent_0 \wedge \sent_1) = \pi(11) = 0$.
  
  From the joint distribution,   we compute literal and argument
  probabilities using Equations~\ref{eqn:sentProb} and
  \ref{eqn:argProb}, respectively, as follows. 
\begin{align*}
    \Pr(\sent_0) &= \pi(10) + \pi(11) = 0.8, \\
    \Pr(\neg \sent_0) &= \pi(00) + \pi(01) = 0.2, \\
    \Pr(\sent_1) &= \pi(01) + \pi(11) = 0.18, \\
    \Pr(\neg \sent_1) &= \pi(00) + \pi(10) = 0.82, \\
    \Pr(\argu{\{\sent_0\}}{\sent_0}) &= \pi(10) + \pi(11) = 0.8, \\
    \Pr(\argu{\{\neg \sent_0, \sent_1\}}{\sent_1}) &= \pi(01) = 0.18.
\end{align*}
\end{example}

Probabilities of arguments that forming attack cycles can be computed
without any special treatment, as illustrated in the following
example.

\begin{example}
  \label{exp:attackCycle}

  Consider a PD framework with a set of p-rules
  $$\{\prule{\sent_0 \gets \neg \sent_1}{0.9},\prule{\sent_1 \gets
    \neg \sent_0}{0.6}\}.$$
  Two arguments $\argA = \argu{\{\neg \sent_1, \sent_0\}}{\sent_0}$
  and $\argB = \argu{\{\neg \sent_0, \sent_1\}}{\sent_1}$ can be built
  \st{} $\argA$ attacks $\argB$ and $\argB$ attacks $\argA$, as
  illustrated in Figure~\ref{fig:cycleAttack}.

  We compute the joint probability distribution as:
  $$\pi(00) = 0.087, \pi(01) = 0.13, \pi(10) = 0.783, \pi(11) = 0.$$
  Note that the joint probability distribution is again unique. With
  P-CWA, we have 
  \begin{equation*}
    \sum_{\omega \in \Omega, \omega \models \sent_0} \pi(\omega) =
    \sum_{\omega \in \Omega, \omega \models \sent_0 \wedge \neg \sent_1} \pi(\omega).    
  \end{equation*}
  and 
  \begin{equation*}
    \sum_{\omega \in \Omega, \omega \models \sent_1} \pi(\omega) =
    \sum_{\omega \in \Omega, \omega \models \sent_1 \wedge \neg \sent_0} \pi(\omega).    
  \end{equation*}
  Both imply $\pi(\sent_0 \wedge \sent_1) = \pi(11) = 0$.
  Thus there are four equations and four unknowns, so the solution is
  unique.
  
  With these, 
  we compute literal and argument probabilities:
  \begin{center}
    $\Pr(\argA) = \pi(10) = 0.783,$ \\
    $\Pr(\argB) = \pi(01) = 0.13.$
  \end{center}
\end{example}

\begin{figure}
  \[\xymatrix@1@C=26pt@R=20pt{
    *+[F]{\argA = \argu{\{\neg \sent_1, \sent_0\}}{\sent_0}} \ar@/^1.5pc/[r] &
    *+[F]{\argB = \argu{\{\neg \sent_0, \sent_1\}}{\sent_1}}  \ar@/^1.5pc/[l]
  }\]
  \caption{Arguments $\argA = \argu{\{\neg \sent_1, \sent_0\}}{\sent_0}$
    and $\argB = \argu{\{\neg \sent_0, \sent_1\}}{\sent_1}$ attack each
    other in Example~\ref{exp:attackCycle}. \label{fig:cycleAttack}}
\end{figure}

A few observations can be made with our notions of arguments and
attacks in PD frameworks as follows.

\begin{itemize}
\item
  Arguments are defined syntactically in that arguments
  are deductions, which are trees with nodes being
  literals and edges defined with p-rules.

\item
  Attacks are also defined syntactically, without referring to
  either literal or argument probabilities, \st{} argument $\argA$
  attacks argument $\argB$ \ifaf{} the claim of the $\argA$ is the
  negation of some literals in $\argB$. In this process, we make no
  distinction between ``undercut'' or ``rebuttal'' as done in some
  other constructions (see e.g., \cite{Eemeren17} for some discussion
  on these concepts).

\item
  The probability semantics of PD framework given in
  Equation~\ref{eqn:argProb} is based on solving the joint
  distribution $\pi$ over the CC set of the language $\mc{L}$ under
  the P-CWA assumption. Thus, it is by design a ``global'' semantics
  in that it requires a sense of ``global consistency'' as given in
  Definition~\ref{dfn:cwaConsistency}. We give a brief discussion in
  \ref{sec:inconsistPD} about reasoning with PD frameworks that
  are not P-CWA consistent.

\item
  Given a PD framework, since its joint distribution $\pi$ may not
  be unique (unless maximum entropy reasoning is enforced), argument
  probabilities may not be unique, as well (subject to the same
  maximum entropy reasoning condition). We discuss the calculation
  of the joint distribution in detail in
  Section~\ref{sec:solveRule-PSAT}. 
\end{itemize}

A few results concerning the argument probability semantics are as
follows. Arguments containing both a literal and its negation have 0
probability. 
\begin{proposition}
  \label{prop:inconsistentSent0Prob}

  For any argument $\argA = \argu{\sents}{\_}$, if $\sent, \neg \sent
  \in \sents$, then $\Pr(\argA) = 0$.
\end{proposition}

Self-attacking arguments have 0 probability. 
\begin{proposition}
  \label{prop:selfAttack0Prob}

  For any argument $\argA = \argu{\sents}{\sent}$, if $\neg \sent \in
  \sents$, then $\Pr(\argA) = 0$.
\end{proposition}

An argument's probability is no higher than the probability of its
claim. 
\begin{proposition}
  \label{prop:argProbLowerThanSentProb}

  For any argument $\argA = \argu{\sents}{\sent}$, $\Pr(\argA) \leq
  \Pr(\sent)$. 
\end{proposition}

If an argument's probability equals the probability of its claim, then
there is one and only one argument for the claim.
\begin{proposition}
  \label{prop:argProbEqSentProb}

  For any argument $\argA = \argu{\sents}{\sent}$, $\Pr(\argA) =
  \Pr(\sent)$ \ifaf{} there is no $\argB \neq \argA$ \st{} $\argB =
  \argu{\sents'}{\sent}$ and $\Pr(\argB) \neq 0$.
\end{proposition}

If an argument is attacked by another argument, then the sum of the
probability of the two arguments is no more than 1.
\begin{proposition}
  \label{prop:attackerAttackeeNoMoreThan1}

  For any two arguments $\argA$ and $\argB$, if $\argA$ attacks
  $\argB$, then $\Pr(\argA) + \Pr(\argB) \leq 1$.
\end{proposition}
Proposition~\ref{prop:attackerAttackeeNoMoreThan1} is the first of the
two conditions of {\em p-justifiable} introduced in
\cite{Thimm12} (Definition 4), also known as the ``coherence
criterion'' introduced in \cite{Hunter14}. Note that the second
condition of p-justifiable, the sum of probabilities of an argument and
its attackers must be no less than 1 (also known as the ``optimistic 
criterion'' in \cite{Hunter14}), does not hold in PD in general, as
illustrated in the following example.

\begin{example}
  \label{exp:2ndCondPjustifiable}

  Consider a PD framework with two p-rules
  $$\prule{\sent_0 \gets \neg \sent_1}{0.1},
  \prule{\sent_1 \gets}{0.1}.$$
  Let $\argA = \argu{\{\sent_1\}}{\sent_1}$
  and $\argB = \argu{\{\sent_0, \neg \sent_1\}}{\sent_0}$. We compute
  $\Pr(\argA) = 0.1$ and $\Pr(\argB) = 0.09$. Clearly, $\argA$ attacks
  $\argB$, yet $\Pr(\argA) + \Pr(\argB) < 1$. This example represents
  a case where a weak argument attacks another weak argument, and
  the sum of the probabilities of the two is less than 1.  
\end{example}

With p-rules, PD naturally support reasoning with both knowledge with
uncertainty and ``hard'' facts in a single framework. A classic
example used in defeasible reasoning, ``Nixon diamond'' introduced by
Reiter and Criscuolo \cite{Reiter81}, can be modelled with a PD
framework as shown in Table~\ref{table:nixon}. We can see that both
arguments
\begin{center}
  $\argA = \argu{\{p, q, n\}}{p}$ and
  $\argB = \argu{\{\neg p, r, n\}}{\neg p}$
\end{center}
can be drawn \st{} they attack each other. Calculate their
probabilities, we have $\Pr(\argA) = \Pr(\argB) = 0.5$.

\begin{table}
  \begin{center}  
  \caption{Modelling the Nixon diamond example with
    PD.\label{table:nixon}}
  \begin{tabular}{|l|l|}
    \hline
    \multicolumn {2}{|c|}{Defeasible Knowledge} \\
    \hline
    usually, Quakers are pacifist & \prule{p \gets q}{0.5} \\
    usually, Republicans are not pacifist & \prule{\neg p \gets
      r}{0.5} \\
    \hline
    \multicolumn {2}{|c|}{``Hard'' Facts} \\
    \hline
    Richard Nixon is a Quaker & \prule{q \gets n}{1} \\
    Richard Nixon is a Republican & \prule{r \gets n}{1} \\
    Nixon exists & \prule{n \gets}{1}\\
    \hline    
  \end{tabular}
  \end{center}  
\end{table}

%

\subsection{PD Frameworks and Abstract Argumentation}
\label{subsec:PDAA}

To show relations between PD and AA~\cite{Dung95},
we first explore a few classic examples studied with AA to
illustrate PD's probabilistic semantics. We then present a few
intermediate results
(Proposition~\ref{prop:AA2PDOne2One}-\ref{prop:arg0Attacker1}). With
them, we show that PD generalises AA
(Theorem~\ref{thm:probCompleteLabel}).

We start by presenting a few
examples~\ref{exp:attackInTurn}-\ref{exp:stableExp}. Arguments and
attacks shown in these examples are used in \cite{Baroni11} to
illustrate differences between several classical (non-probabilistic)
argumentation semantics. Characteristics of these examples are
summarised in Table~\ref{table:expChar}.

\begin{table}
  \caption{Example Argumentation Frameworks Characteristics.
    \label{table:expChar}}
  \begin{tabular}{|l|l|l|}
    \hline
    Example & Description & Source \\
    \hline
    Example~\ref{exp:attackInTurn} &
    Three arguments and two attacks &
    Figure 2 in \cite{Baroni11} \\
    Example~\ref{exp:twoAttack} &
    Two arguments attack each other &
    Figure 3 in \cite{Baroni11} \\
    Example~\ref{exp:circleAttackAttack} &
    ``Floating Acceptance'' example &
    Figure 5 in \cite{Baroni11} \\
    Example~\ref{exp:threeCycle} &
    ``Cycle of three attacking arguments'' example &
    Figure 6 in \cite{Baroni11} \\
    Example~\ref{exp:stableExp} &
    Stable extension example &
    Figure 8 in \cite{Baroni11}\\
    \hline
  \end{tabular}
\end{table}

\begin{example}
  \label{exp:attackInTurn}
Let $F$ be a PD framework with three p-rules:
$$\{\prule{\sent_0 \gets }{1},
\prule{\sent_1 \gets \neg \sent_0}{1},
\prule{\sent_2 \gets \neg \sent_1}{1}
\}.$$
Let $\argA = \argu{\{\sent_0\}}{\sent_0}$,
$\argB = \argu{\{\sent_1, \neg \sent_0\}}{\sent_1}$,
$\argC = \argu{\{\sent_2, \neg \sent_1\}}{\sent_2}$.
Arguments and attacks are shown in Figure~\ref{fig:attackInTurn}.
The joint distribution $\pi$ is \st{}

\begin{center}
\begin{tabular}{llll}
$\pi(000) = 0$, &
$\pi(001) = 0$, &
$\pi(010) = 0$, &
$\pi(011) = 0$, \\
$\pi(100) = 0$, &
$\pi(101) = 1$, &
$\pi(110) = 0$, &
$\pi(111) = 0$.
\end{tabular}
\end{center}

\noindent
With these, we have $\Pr(\argA) = 1$, $\Pr(\argB) = 0$, and
$\Pr(\argC) = 1$.

\begin{figure}
  \[\xymatrix@1@C=26pt@R=20pt{
    *+[F]{\argA = \argu{\{\sent_0\}}{\sent_0}} \ar[r] &
    *+[F]{\argB = \argu{\{\sent_1, \neg \sent_0\}}{\sent_1}} \ar[r] &
    *+[F]{\argC = \argu{\{\sent_2, \neg \sent_1\}}{\sent_2}}
  }\]
  \caption{The PD framework shown in
    Example~\ref{exp:attackInTurn} \label{fig:attackInTurn}}  
\end{figure}

\end{example}

\begin{example}
  \label{exp:twoAttack}
  Let $F$ be a PD framework with two p-rules:
  $$\{\prule{\sent_0
    \gets \neg \sent_1}{1}, \prule{\sent_1 \gets \neg
    \sent_0}{1}\}.$$
  Let $\argA = \argu{\{\sent_0, \neg
    \sent_1\}}{\sent_0}$, and $\argB = \argu{\{\sent_1, \neg
    \sent_0\}}{\sent_1}$. The arguments and attacks are shown in
  Figure~\ref{fig:attackExp}. 
  A joint distribution $\pi$ is \st{}
  $$\pi(00) = 0,
  \pi(01) = 0.5,
  \pi(10) = 0.5,
  \pi(11) = 0.$$
  With these, we have $\Pr(\argA) = \Pr(\argB) = 0.5$. 
\end{example}

\begin{example}
  \label{exp:circleAttackAttack}
  
  Let $F$ be a PD framework with four p-rules:

  \begin{center}
  \begin{tabular}{llll}
  $\{\prule{\sent_0 \gets \neg \sent_1}{1}$,&
  $\prule{\sent_1 \gets \neg \sent_0}{1}$,&
  $\prule{\sent_2 \gets \neg \sent_0, \neg \sent_1}{1}$,&
  $\prule{\sent_3 \gets \neg \sent_2}{1}\}$.
  \end{tabular}
  \end{center}

  \noindent
  Let $\argA = \argu{\{\sent_0, \neg \sent_1\}}{\sent_0}$,
      $\argB = \argu{\{\sent_1, \neg \sent_0\}}{\sent_1}$,
      $\argC = \argu{\{\sent_2, \neg \sent_0, \neg \sent_1\}}{\sent_2}$, and
  $\argX{D} = \argu{\{\sent_3, \neg \sent_2\}}{\sent_3}$.
  The arguments and attacks are shown in
  Figure~\ref{fig:circleAttackAttack}.
  The joint distribution $\pi$ is \st{}

  \[ \pi(\omega_i) =
  \begin{cases}
    0.5  & \quad \text{if } \omega_i = \neg \sent_0 \wedge \sent_1
  \wedge \neg \sent_2 \wedge \sent_3, \\
    0.5  & \quad \text{else if } \omega_i = \sent_0 \wedge \neg \sent_1
  \wedge \neg \sent_2 \wedge \sent_3, \\
    0    & \quad \text{otherwise.}
  \end{cases}
  \]
  Compute argument probabilities, we have $\Pr(\argA) = \Pr(\argB) =
  0.5$, $\Pr(\argC) = 0$, and $\Pr(\argX{D}) = 1$.

  \begin{figure}
    \[\xymatrix@1@C=0pt@R=30pt{
      *+[F]{\argA = \argu{\{\sent_0, \neg \sent_1\}}{\sent_0}}
      *\ar@/^1.5pc/[rr] \ar[dr] & &
      *+[F]{\argB = \argu{\{\sent_1, \neg \sent_0\}}{\sent_1}}
      *\ar@/^1.5pc/[ll] \ar[dl] \\
      & *+[F]{\argC = \argu{\{\sent_2, \neg \sent_0, \neg
      \sent_1\}}{\sent_2}} 
      *\ar[d] \\
      & *+[F]{\argX{D} = \argu{\{\sent_3, \neg \sent_2\}}{\sent_3}}
    }\]
    \caption{The 0/1 PD framework shown in
    Example~\ref{exp:circleAttackAttack}. \label{fig:circleAttackAttack}}
\end{figure} 
\end{example}

\begin{example}
  \label{exp:threeCycle}
  
  Let $F$ be a PD framework with three p-rules:
  $$\{\prule{\sent_0 \gets \neg \sent_1}{1},
  \prule{\sent_1 \gets \neg \sent_2}{1},
  \prule{\sent_2 \gets \neg \sent_0}{1}\}.$$
  Let $\argA = \argu{\{\sent_0, \neg \sent_1\}}{\sent_0}$,
      $\argB = \argu{\{\sent_1, \neg \sent_2\}}{\sent_1}$, and 
  $\argC = \argu{\{\sent_2, \neg \sent_0\}}{\sent_2}$.
  Arguments and attacks are shown in Figure~\ref{fig:threeCycle}.  
  Although $F$ is Rule-PSAT, it is not P-CWA consistent. Thus, there
  is no consistent joint distribution over the CC set.
  $\Pr(\argA)$, $\Pr(\argB)$ and $\Pr(\argC)$ are undefined.

\begin{figure}
  \[\xymatrix@1@C=26pt@R=26pt{
    *+[F]{\argA = \argu{\{\sent_0, \neg \sent_2\}}{\sent_0}} \ar[r] &
    *+[F]{\argB = \argu{\{\sent_1, \neg \sent_0\}}{\sent_1}} \ar[r] &
    *+[F]{\argC = \argu{\{\sent_2, \neg \sent_1\}}{\sent_2}} \ar@/^1.8pc/[ll]
  }\]
  \caption{The PD framework shown in
    Example~\ref{exp:threeCycle}. \label{fig:threeCycle}}  
\end{figure}  
\end{example}

\begin{example}
  \label{exp:stableExp}
  
  Let $F$ be a PD framework with five p-rules:
  \begin{center}
  \begin{tabular}{lll}
  $\{\prule{\sent_0 \gets \neg \sent_1}{1},$ &
  $\prule{\sent_1 \gets \neg \sent_0}{1},$ &
  $\prule{\sent_2 \gets \neg \sent_1, \neg \sent_4}{1},$ \\
  $\prule{\sent_3 \gets \neg \sent_2}{1},$ &
  $\prule{\sent_4 \gets \neg \sent_3}{1}\}.$
  \end{tabular}
  \end{center} 
  Let
  $\argA = \argu{\{\sent_0, \neg \sent_1\}}{\sent_0}$,
  $\argB = \argu{\{\sent_1, \neg \sent_0\}}{\sent_1}$, 
  $\argC = \argu{\{\sent_2, \neg \sent_1, \neg \sent_4\}}{\sent_2}$,
  $\argX{D} = \argu{\{\sent_3, \neg \sent_2\}}{\sent_3}$, and
  $\argX{E} = \argu{\{\sent_4, \neg \sent_3\}}{\sent_4}$.
  Arguments and attacks are shown in Figure~\ref{fig:stableExp}.  
  The joint distribution $\pi$ is \st{}

  \[ \pi(\omega_i) =
  \begin{cases}
    1    & \quad \text{if } \omega_i = \neg \sent_0 \wedge \sent_1
  \wedge \neg \sent_2 \wedge \sent_3 \wedge \neg \sent_4, \\
    0    & \quad \text{otherwise.}
  \end{cases}
  \]
  Compute argument probabilities, we have $\Pr(\argA) = \Pr(\argC) =
  \Pr(\argX{E}) = 0$, and $\Pr(\argB) = \Pr(\argX{D}) = 1$.  

  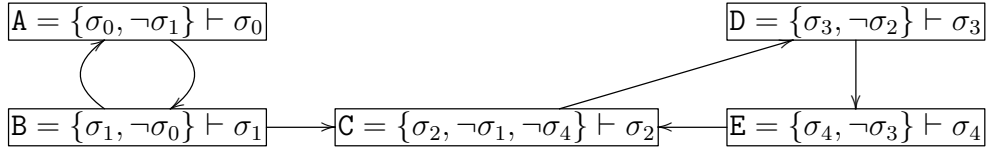
\begin{figure}
    \[\xymatrix@1@C=26pt@R=26pt{
      *+[F]{\argA = \argu{\{\sent_0, \neg \sent_1\}}{\sent_0}} \ar@/^1.8pc/[d] &&
      *+[F]{\argX{D} = \argu{\{\sent_3, \neg \sent_2\}}{\sent_3}} \ar[d] \\
      *+[F]{\argB = \argu{\{\sent_1, \neg \sent_0\}}{\sent_1}} \ar@/^1.8pc/[u]\ar[r] &
      *+[F]{\argC = \argu{\{\sent_2, \neg \sent_1, \neg \sent_4\}}{\sent_2}} \ar[ur] &
      *+[F]{\argX{E} = \argu{\{\sent_4, \neg \sent_3\}}{\sent_4}} \ar[l]
    }\]
    \caption{The PD framework shown in
      Example~\ref{exp:stableExp}. \label{fig:stableExp}}  
  \end{figure}  
\end{example}

We make two observations from these examples.
\begin{itemize}
  \item
Syntactically, it is straightforward to map AA frameworks to PD
frameworks \st{} a one-to-one mapping between arguments and attacks in 
an AA frameworks and their counterparts in the mapped PD framework
exist. Thus, for any AA framework $F$, there is a counterpart of it
represented as a PD framework. 

\item
Semantically, the probability semantics of PD frameworks in these
examples behaves intuitively, in the sense that:
\begin{itemize}
  \item
    winning arguments have probability 1;
  \item
    losing arguments have probability 0;
  \item
    arguments that can either win or lose have probability between 0
    and 1; and
  \item
    arguments cannot be labelled neither winning nor losing result in
    inconsistency.
\end{itemize}
With these observations, more formally, as we show below, when
argument probabilities are viewed as labelling, they represent a
complete labelling.
\end{itemize}
Starting with the syntactical aspect, we define a mapping from AA
frameworks to PD frameworks as follows.

\begin{definition}
  \label{dfn:AA2PD}

  The function \aapd{} is a mapping from AA frameworks to PD
  frameworks \st{} for an AA framework $\AAF$, $\aapd(\AAF)
  = \pdf$, where:
  \begin{itemize}
  \item
    $\mc{L} = \mc{A}$,
    
  \item
    $\mc{R} = \{\prule{\sent \gets \neg \sent_1, \ldots, \neg \sent_m}{1} |
    \sent \in \mc{A}$, $\{\sent_1, \ldots, \sent_m\} = \{\sent_i |
    (\sent_i, \sent) \in \mc{T}\}\}$.  
  \end{itemize}
\end{definition}

Proposition~\ref{prop:AA2PDOne2One} below sanctions that the arguments
and attacks in AA frameworks are mapped to their counterparts in
PD frameworks unambiguously. 

\begin{proposition}
  \label{prop:AA2PDOne2One}

  Given an AA framework $F = \AAF$, there exists a
  function $\mt{g}$ that maps arguments in $F$ to arguments in
  $\aapd(F)$ \st{} for any $(\argA, \argB) \in \mc{T}$,
  $\mt{g}(\argA)$ attacks $\mt{g}(\argB)$ in $\aapd(F)$.
\end{proposition}

To establish semantics connections between AA frameworks and PD
frameworks, we first introduce AA-PD frameworks as the set of PD
frameworks that are mapped from AA frameworks, i.e., let $\mc{F}$ be
the set of AA frameworks, then the set of AA-PD frameworks is
$\{\aapd(F)|F \in \mc{F}\}$. We observe the following with AA-PD
frameworks: 
\begin{enumerate}
\item
  unattacked arguments have probability 1 (note that this is the
  ``founded'' criterion in \cite{Hunter14}), and

\item
  if an argument $\argA$ with probability 1 attacks another argument
  $\argB$, $\argB$ will have probability 0.
\end{enumerate}  
Formally, we have propositions~\ref{prop:nonAttackedProb1} and
\ref{prop:zeroIfAttackedByOne} as follows. 

\begin{proposition}
  \label{prop:nonAttackedProb1}

  Given an AA-PD framework $F$, for an argument $\argA =
  \argu{S}{\sent}$ in $F$, if $\argA$ is not attacked in $F$, then 
  $\Pr(\argA) = 1$.
\end{proposition}

\begin{proposition}
  \label{prop:zeroIfAttackedByOne}

  Given an AA-PD framework $F$, for two arguments $\argA$ and $\argB$
  in $F$ \st{} $\argA$ attacks $\argB$. If $\Pr(\argA) = 1$ then
  $\Pr(\argB) = 0$.
\end{proposition}

Extending Propositions~\ref{prop:nonAttackedProb1} and
\ref{prop:zeroIfAttackedByOne}, we can show that if all attackers of 
an argument have probability 0, then the argument has probability 1;
moreover, if an argument that has been attacked has probability 1,
then all of its attackers must have probability 0. Formally, 

\begin{proposition}
  \label{prop:oneIfAttackedByZero}

  Given an AA-PD framework $F$, let $\argA$ be an argument in $F$ and
  $\args$ the set of arguments attacking $\argA$, $\argA \not\in
  \args$. 
  \begin{enumerate}
    \item
      If for all $\argB \in \args$, $\Pr(\argB) = 0$, then
      $\Pr(\argA) = 1$. 

    \item
      If $\Pr(\argA) = 1$, then for all $\argB \in \args$,
      $\Pr(\argB) = 0$. 
  \end{enumerate}  
\end{proposition}
\noindent
An argument has probability 0 \ifaf{} it has an attacker with
probability 1. 

\begin{proposition}
  \label{prop:arg0Attacker1}

  Given an AA-PD framework $F$, let $\argA$ be an argument in $F$ and 
  $\args$ the set of arguments attacking $\argA$, $\argA \not\in
  \args$. With maximum entropy reasoning, 
  \begin{enumerate}
  \item
    if $\Pr(\argA) = 0$, then there exists $\argB \in \args$, \st{}
    $\Pr(\argB) = 1$;

  \item
    if there exists $\argB \in \args$, \st{} $\Pr(\argB) = 1$, then
    $\Pr(\argA) = 0$. 
  \end{enumerate}  
\end{proposition}

Maximum entropy reasoning is a key condition for
Proposition~\ref{prop:arg0Attacker1}. This proposition does not
hold without it, as illustrate in the following example.
\begin{example}
  \label{exp:noME}

  Consider an AA-PD framework with five p-rules:
  \begin{center}
    \begin{tabular}{ccc}
      \prule{\sent_1 \gets \neg \sent_2}{1}, &
      \prule{\sent_2 \gets \neg \sent_1}{1}, &
      \prule{\sent_3 \gets \neg \sent_1, \neg \sent_4}{1}, \\
      \prule{\sent_4 \gets \neg \sent_5}{1}, &
      \prule{\sent_5 \gets \neg \sent_4}{1}.
    \end{tabular}
  \end{center}
  The arguments and attacks are shown in
  Figure~\ref{fig:noME}. Calculate the joint probability distribution
  with maximum entropy reasoning, we have the solution
  \[ \pi(\omega_i) =
  \begin{cases}
    0.25 & \quad \text{if } \omega_i \in \{01011, 01100, 10010, 10100\},\\
    0    & \quad \text{otherwise.}
  \end{cases}
  \]
  This solution gives $\Pr(\argX{A}) = \Pr(\argX{B}) =
  \Pr(\argX{C}) = \Pr(\argX{D}) = 0.5$ and $\Pr(\argX{E}) = 0.25$.

  Without maximum entropy reasoning, a possible solution is
  \[ \pi(\omega_i) =
  \begin{cases}
    0.33 & \quad \text{if } \omega_i \in \{01100, 10010, 10100\},\\
    0    & \quad \text{otherwise.}
  \end{cases}
  \]
  This joint distribution gives $\Pr(\argX{A}) = \Pr(\argX{C}) = 0.67,
  \Pr(\argX{B}) = \Pr(\argX{D}) = 0.33$ and $\Pr(\argX{E}) =
  0.$ Thus, $\Pr(\argX{E}) = 0$ even though neither of the two
  arguments ($\argX{A}$ and $\argX{C}$) attacking $\Pr(\argX{E})$ has
  probability 1. 

\begin{figure}
  \[\xymatrix@1@C=26pt@R=20pt{
    *+[F]{\argX{A} = \argu{\{\sent_1\, \neg \sent_2\}}{\sent_1}} \ar[r]\ar@/^1.5pc/[d] &
    *+[F]{\argX{E} = \argu{\{\sent_3, \neg \sent_1, \neg \sent_4\}}{\sent_3}}  &
    *+[F]{\argX{C} = \argu{\{\sent_4, \neg \sent_5\}}{\sent_4}} \ar[l]\ar@/^1.5pc/[d] \\
    *+[F]{\argX{B} = \argu{\{\sent_2\, \neg \sent_1\}}{\sent_2}} \ar@/^1.5pc/[u] &&
    *+[F]{\argX{D} = \argu{\{\sent_5, \neg \sent_4\}}{\sent_5}} \ar@/^1.5pc/[u]
  }\]
  \caption{The AA-PD framework shown in Example~\ref{exp:noME}. \label{fig:noME}}  
\end{figure}
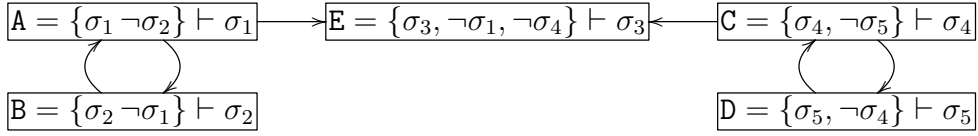

\end{example}

Propositions~\ref{prop:nonAttackedProb1} - \ref{prop:arg0Attacker1}
describe attack relations between PD arguments similar to
attacks in AA (or other non-probabilistic argumentation
frameworks). For instance, if we consider arguments with probability 0
\mt{out} and probability 1 \mt{in}, then we obtain a labelling-based
semantics as shown in \cite{Baroni11}. Formally,

\begin{theorem}
  \label{thm:probCompleteLabel}

  Given an AA PD framework $F$, let $\args$ be the set of arguments in
  $F$, with maximum entropy reasoning, the {\em Probabilistic
    Labelling} function $\Xi: \args \mapsto \{\mt{in}, \mt{out},
  \mt{undec}\}$ defined as 

  \[ \Xi(\argA) =
  \begin{cases}
    \mt{in}    & \quad \text{if } \Pr(\argA) = 1,\\
    \mt{out}   & \quad \text{if } \Pr(\argA) = 0,\\
    \mt{undec} & \quad \text{otherwise.}
  \end{cases}
  \]
  in which $\argA \in \args$, is a complete labelling.
\end{theorem}

Theorem~\ref{thm:probCompleteLabel} bridges PD and AA semantically as
relations between the complete labelling and argument extensions have
been studied extensively in e.g.,
\cite{Baroni11,Caminada17,Caminada09-labelling}. In short, labelling
can be mapped to extensions in the way that given some semantics $s$, 
arguments that are labelled \mt{in} with $s$-labelling belong to an
$s$-extension. Moreover, the complete labelling can be viewed at the
centre of defining labellings for other semantics \cite{Torre17}. For
instance, a grounded labelling is a complete labelling such that the
set of arguments labelled in is minimal with respect to set inclusion
among all complete labellings; a stable labelling is a complete
labelling such that the set of undecided arguments is empty; a
preferred labelling is a complete labelling such that the set of
arguments labelled in is maximal with respect to set inclusion among
all complete labellings \cite{Torre17}.

One last result we would like to present on AA-PD framework is the
following.
\begin{proposition}
  \label{prop:optimistic}

  Given an AA-PD framework $F$, let $\argA$ be an argument in $F$ and
  $\args$ the set of arguments attacking $\argA$, $\argA \not\in
  \args$. $\Pr(\argA) \geq 1 - \sum_{\argB \in \args} \Pr(\argB)$.
\end{proposition}
\noindent
This is the ``optimistic criterion'' introduced in
\cite{Hunter14,Hunter17}. We will discuss more on the relation between
AA-PD and probabilistic abstract argumentation in
Section~\ref{subsec:relation}.

In this section, we have introduced arguments built with p-rules and
attacks between arguments in PD frameworks. In PD frameworks,
arguments are deductions as they are in ABA frameworks, 
\cite{Cyras17}. Attacks are defined syntactically \st{} an argument
$\argA$ attacks another argument $\argB$ if the claim of $\argA$ is
the negation of some literal in $\argB$. We have compared PD with AA
and show that AA can be mapped to PD frameworks containing only rules
assigned with probability 1. The key insight is that the probability
semantics given by PD can be viewed as a complete labelling as defined
in AA.

\section{Probability Calculation}
\label{sec:solveRule-PSAT}

So far we have introduced the probability semantics of p-rules in
Section~\ref{sec:prules} and argument construction in
Section~\ref{sec:argAtt}. In both sections, we have assumed that the
joint probability distribution $\pi$ for the CC set can be
computed. In this section, we study methods for computing $\pi$ from
p-rules. We look at methods for computing exact solutions as well as
their approximations.

\subsection{Compute Joint Distribution with Linear Programming}
\label{subset:LP}

We begin with methods for computing exact solutions. 
Given a set of p-rules $\mc{R} = \{\rho_1, \ldots, \rho_m\}$ \st{}
$\mc{L}$ contains $n$ literals, to test whether $\mc{R}$ is
Rule-PSAT, we set up a linear system
\begin{equation}
  \label{eqn:APiB}
A\pi = B,
\end{equation}
where $A$ is an $(m + 1)$-by-$2^n$ matrix, $\pi = [\pi(\omega_1),
  \ldots, \pi(\omega_{2^n})]^T$, $B$ an $(m+1)$-by-$1$
matrix.\footnote{We let $\{\omega_1,\ldots,\omega_{2^n}\}$ be the CC
  set of $\mc{L}$. We consider elements in this set being ordered
  with their Boolean values. E.g., for $\mc{L} = \{\sent_0,
    \sent_1\}$, the four elements in the CC set are ordered such that
  $\{\omega_1 = \neg \sent_0 \wedge \neg \sent_1,
    \omega_2 = \neg \sent_0 \wedge \sent_1,
    \omega_3 = \sent_0 \wedge \neg \sent_1,
    \omega_4 = \sent_0 \wedge \sent_1\}$.\label{footnote:order}}
We construct $A$ and $B$ in a way such that $R$ is Rule-PSAT \ifaf{} $\pi$
has a solution in $[0,1]^{2^n}$, as follows.

For each rule $\rho_i \in \mc{R}$, if $\rho_i = \prule{\sent_0
  \gets}{\theta}$ has an empty body, then 
\begin{equation}
  \label{eqn:AijHead}
  A[i,j] =
  \begin{cases}
    1, & \mbox{if } \omega_j \models \sent_0 \mbox{;} \\
    0, & \mbox{otherwise};
  \end{cases}
\end{equation}
and
\begin{equation}
  \label{eqn:Bi}
B[i] = \theta.
\end{equation}
Otherwise, $\rho_i = \prule{\sent_0 \gets \sent_1, \ldots,
  \sent_k}{\theta}$, then
\begin{equation}
  \label{eqn:AijBody}
  A[i,j] =
  \begin{cases}
    \theta - 1, & \mbox{if } \omega_j \models \sent_0 \wedge \sent_1 \wedge \ldots \wedge \sent_k \mbox{;} \\
    \theta,     & \mbox{if } \omega_j \models \neg \sent_0 \wedge \sent_1 \wedge \ldots \wedge \sent_k \mbox{;} \\    
    0,              & \mbox{otherwise};
  \end{cases}
\end{equation}
\begin{equation}
  \label{eqn:Bi0}
B[i] = 0.
\end{equation}
Row $m+1$ in $A$ and $B$ are $1\ldots1$ and $1$, respectively.

\begin{example}
  \label{exp:solve}

  Consider a p-rule set containing two p-rules, $\rho_0, \rho_1$.   
  Let
  $$\rho_0 = \prule{\sent_0 \gets \sent_1}{\alpha},
  \rho_1 = \prule{\sent_1 \gets }{\beta}.$$
  Here, $m = 2$, $n = 2$.  
  From Equations~\ref{eqn:APiB}
  to \ref{eqn:Bi0}, we have
  \begin{equation*}
    \fixTABwidth{T}
    A = \bracketMatrixstack{
      0 & \alpha & 0 & \alpha - 1 \\
      0 & 1 & 0 & 1 \\
      1 & 1 & 1 & 1     
    },
  \end{equation*}
  $\pi = [\pi(\neg \sent_0 \wedge \neg \sent_1), 
    \pi(\neg \sent_0 \wedge \sent_1),
    \pi(\sent_0 \wedge \neg \sent_1),
    \pi(\sent_0 \wedge \sent_1)]^T$,
  and
    $B = [0, \beta, 1]^T.$
  It is easy to see that $\pi$ has solutions as shown in
  Example~\ref{exp:Rule-PSAT}.
\end{example}

\begin{theorem}
  \label{thm:Rule-PSAT}
  \cite{Fan22} 
  Given a set of p-rules $\mc{R}$ on some language $\mc{L}$, $\mc{R}$
  is Rule-PSAT \ifaf{} Equation~\ref{eqn:APiB} has a solution for
  $\pi$ in $[0,1]^{2^n}$.
\end{theorem}

\subsection{Compute Joint Distribution under P-CWA}
\label{subsec:ComputePCWA}

To reason with P-CWA, additional equations must be introduced as
constraints. To this end, we revise the construction of matrices $A$
and $B$ in Equation~\ref{eqn:APiB} as given in
Equations~\ref{eqn:AijHead} \& \ref{eqn:AijBody} and
Equations~\ref{eqn:Bi} \& \ref{eqn:Bi0}, respectively, to meet the
requirement given in Equation~\ref{eqn:sent_prob_cwa}. 

In revising constructions of these two matrices, one useful
observation we can make is that the P-CWA constraint 
$$\sum_{\omega_i \in \Omega, \omega_i \models x} \pi(\omega_i) =
\sum_{\omega_i \in \Omega, \omega_i \models S} \pi(\omega_i)$$
can be computed ``locally'' in the sense that one does not need to
explicitly identify $S$, the disjunction of conjunctions of literals
that are in deductions for $x$ (Equation~\ref{eqn:bigS}), 
when computing $\Pr(x)$ for each literal $x$. This is important as if
we were to identify $S$ explicitly upon computing $\Pr(x)$ for each
$x$, then we need to compute all deductions of $x$, which is both
repetitive and computationally expensive.
We first illustrate the ``local computation'' idea with two
examples and then present the algorithm along with a formal
proof.

Consider three p-rules: 

\begin{center}
  $\prule{\sent_0 \gets \sent_1}{\cdot},
  \prule{\sent_1 \gets \sent_2}{\cdot},
  \prule{\sent_2 \gets}{\cdot}$.
\end{center}
Directly applying the definition of P-CWA, we have
\begin{itemize}
  \item
    $\Pr(\sent_0) = \pi(\sent_0 \wedge \sent_1 \wedge \sent_2)$
    from the deduction $\argu{\{\sent_0, \sent_1,
      \sent_2\}}{\sent_0}$\footnote{To simplify the presentation, we
    use $\pi(s)$ to denote $\sum_{\omega \in \Omega, \omega \models s}
    \pi(\omega)$. E.g., $\pi(\sent_0 \wedge \sent_1 \wedge \sent_2)$
    denotes $\sum_{\omega \in \Omega, \omega \models \sent_0 \wedge
      \sent_1 \wedge \sent_2} \pi(\omega).$}, and
  \item
    $\Pr(\sent_1) = \pi(\sent_1 \wedge \sent_2)$   
    from the deduction $\argu{\{\sent_1, \sent_2\}}{\sent_1}$.  
\end{itemize}
However, if we were to take the ``global'' view and directly encode
$$\Pr(\sent_0) = \pi(\sent_0 \wedge \sent_1 \wedge \sent_2)$$ with the
equation
\begin{equation}
  \label{eqn:globalExp}
\sum_{\omega \in \Omega, \omega \models \sent_0, \omega \not \models
  \sent_0 \wedge \sent_1 \wedge \sent_2}\pi(\omega) = 0
\end{equation}
then we must traverse all three rules to find the deduction
$\arguD{\{\sent_0, \sent_1, \sent_2\}}{\sent_0}$. Instead of doing this
traversal, we can simply encode 
\begin{itemize}
  \item
    $\Pr(\sent_0) = \pi(\sent_0 \wedge \sent_1)$ from the p-rule
    $\prule{\sent_0 \gets \sent_1}{\cdot}$ with
    \begin{equation}
      \label{eqn:localExp1}
    \sum_{\omega \in \Omega, \omega \models \sent_0, \omega \not \models
      \sent_0 \wedge \sent_1}\pi(\omega) = 0,
    \end{equation}
    and 
  \item
    $\Pr(\sent_1) = \pi(\sent_1 \wedge \sent_2)$ from the p-rule
    $\prule{\sent_1 \gets \sent_2}{\cdot}$ with
    \begin{equation}
      \label{eqn:localExp2}
      \sum_{\omega \in \Omega, \omega \models \sent_1, \omega \not \models
        \sent_1 \wedge \sent_2}\pi(\omega) = 0.
    \end{equation}
\end{itemize}

The new equations \ref{eqn:localExp1} and \ref{eqn:localExp2} are
``local'' as given a rule $head \gets body$, we simply assert
$\Pr(head) = \pi(head \wedge body)$. There is no deduction
construction or multi-rule traversal, which is needed for constructing
Equation~\ref{eqn:globalExp}.  To see their equivalence, we
show that they assign the same set of $\omega \in \Omega$ to 0.

\begin{example}
  \label{exp:localForGlobal}
  
  We examine the $\omega$ assigned to 0 from each equation. To
  simplify the presentation, we again use the Boolean string
  representation introduced in Example~\ref{exp:twoRulesSameHead}
  for literals. E.g., ``110'' denotes $\sent_0 \wedge \sent_1 \wedge
  \neg \sent_2$.

  \begin{itemize}
  \item
    $\Pr(\sent_0) = \pi(\sent_0 \wedge \sent_1 \wedge \sent_2)$
    asserts that $\pi(100) = \pi(101) = \pi(110) = 0$.\footnote{These
      are easy to see as the first bit in the three-bit string must be
    1 so the conjunction represented by the string satisfies
    $\sent_0$; the remaining two bits cannot both be 1 as that would
    make the conjunction satisfies $\sent_0 \wedge \sent_1 \wedge
    \sent_2$. So we have 100, 101, and 110 produced in this case.}
    
  \item
    $\Pr(\sent_1) = \pi(\sent_1 \wedge \sent_2)$
    asserts that $\pi(010) = \pi(110) = 0$.\footnote{Similarly, in
      this case the second bit must be 1 to satisfy $\sent_1$, and the
    third bit must be 0 to not satisfy $\sent_1 \wedge \sent_2$. There
    is no constraint on the first bit, so we produce 010 and 110 in
    this case.}

  \item
    $\Pr(\sent_0) = \pi(\sent_0 \wedge \sent_1)$
    asserts that $\pi(100) = \pi(101) = 0$.
  \end{itemize}
  We see that the only difference between
  \begin{center}
  $\Pr(\sent_0) = \pi(\sent_0
  \wedge \sent_1 \wedge \sent_2)$ and $\Pr(\sent_0) = \pi(\sent_0
  \wedge \sent_1)$
  \end{center}
  is on setting $\pi(110) = 0$. Yet, this is asserted
  by $\Pr(\sent_1) = \pi(\sent_1 \wedge \sent_2)$, which is available
  in both the ``global'' and the ``local'' versions of constraint.
\end{example}

The above example illustrates the case where p-rules with different
heads are chained. When there are two p-rules with the same head,
e.g., there exist
\begin{center}
  \prule{\sent_0 \gets \sent_1}{\cdot} and
  \prule{\sent_0 \gets \sent_2}{\cdot},
\end{center}
then we assert
$$\Pr(\sent_0) = \pi((\sent_0 \wedge \sent_1) \vee (\sent_0 \wedge
\sent_2)) = \sum_{\omega \in \Omega, \omega \models (\sent_0 \wedge \sent_1) \vee (\sent_0 \wedge \sent_2)} \pi(\omega).$$

\begin{example}
  \label{exp:localPCWAMultiRules}

  Consider five p-rules:
  \begin{center}
    \prule{\sent_0 \gets \sent_1}{\cdot}, 
    \prule{\sent_0 \gets \sent_2}{\cdot},
    \prule{\sent_1 \gets \sent_3}{\cdot},
    \prule{\sent_2 \gets}{\cdot},
    \prule{\sent_3 \gets}{\cdot}.
  \end{center}
  With a direct application of P-CWA definition, we have
  \begin{center}
    $\Pr(\sent_0) = \pi((\sent_0 \wedge \sent_1 \wedge
    \sent_3) \vee (\sent_0 \wedge \sent_2))$ and
    $\Pr(\sent_1) = \pi(\sent_1 \wedge \sent_3)$.
  \end{center}
  We show that this is the same as asserting
  \begin{center}
    $\Pr(\sent_0) = \pi((\sent_0 \wedge \sent_1) \vee (\sent_0 \wedge
    \sent_2))$ and 
    $\Pr(\sent_1) = \pi(\sent_1 \wedge \sent_3)$.
  \end{center}


  \noindent
  Using the Boolean string representation, E.g., ``1001'' denotes
  $\sent_0 \wedge \neg \sent_1 \wedge \neg \sent_2 \wedge \sent_3$,

  \begin{itemize}
    \item
      with $\Pr(\sent_0) = \pi((\sent_0 \wedge \sent_1 \wedge
      \sent_3) \vee (\sent_0 \wedge \sent_2))$, we assert
      $$\pi(1000) = \pi(1001) = \pi(1100) = 0.$$

    \item
      With $\Pr(\sent_1) = \pi(\sent_1 \wedge \sent_3)$, we assert
      $$\pi(0100) = \pi(0110) = \pi(1100) = \pi(1110) = 0.$$

    \item
      With $\Pr(\sent_0) = \pi((\sent_0 \wedge \sent_1) \vee (\sent_0 \wedge
      \sent_2))$, we assert 
      $$\pi(1000) = \pi(1001) = 0.$$
  \end{itemize}
  \noindent
  We can see that the difference between the ``global'' constraint
  $$\Pr(\sent_0) = \pi((\sent_0 \wedge \sent_1 \wedge \sent_3) \vee
  (\sent_0 \wedge \sent_2))$$ and the ``local'' one
  $$\Pr(\sent_0) = \pi((\sent_0 \wedge \sent_1) \vee (\sent_0 \wedge
  \sent_2))$$ is on 
  asserting $\pi(1100) = 0$. However, this is asserted by
  $\Pr(\sent_1) = \pi(\sent_1 \wedge \sent_3)$. Thus the ``global''
  constraint is indeed satisfied by the ``local'' version.
\end{example}

Summarising these two examples, additional rows of matrices $A$ and
$B$ describing P-CWA can be constructed as follows.

Given a set of p-rules $\mc{R}$ \st{} $\sents = \{\sent_1, \ldots,
\sent_m\}$ are heads of p-rules in $\mc{R}$, for each $\sent \in
\sents$, let
$$\prule{\sent \gets \sent_1^1, \ldots, \sent_{l1}^1}{\cdot},
\ldots, \prule{\sent \gets \sent_1^m, \ldots, \sent_{lm}^m}{\cdot}$$
be the p-rules in $\mc{R}$ with head $\sent$. We construct 
\begin{equation}
  \label{eqn:smallS}
  s = (\sent \wedge \sent_1^1 \wedge \ldots \wedge \sent_{l1}^1) \vee
  \ldots \vee (\sent \wedge \sent_1^m \wedge \ldots \wedge
  \sent_{lm}^m).
\end{equation}
For each $\sent \in \sents$, append a new row $i$ to $A$ \st{}
\begin{equation}
  \label{eqn:AijPCWA}
  A[i,j] =
  \begin{cases}
    1, & \mbox{if } \omega_j \models \sent \mbox{ and }
    \omega_j \not\models s \mbox{,} \\
    0, & \mbox{otherwise};
  \end{cases}
\end{equation}
and
\begin{equation}
  \label{eqn:BiPCWA}
B[i] = 0,
\end{equation}
where $j = 1\ldots 2^n$ are the column indices of $A$, $\omega_j$ 
all atomic conjunctions in $\Omega$. Here, we again consider that $\Omega$
is ordered with the Boolean values of its elements as in footnote
\ref{footnote:order}. 

\begin{theorem}
  \label{thm:localForGlobal}

  Given a set of p-rules $\mc{R}$ on $n$ literals, if there is a
  solution $\pi \in [0,1]^{2^n}$ to $$A\pi=B$$ with $A$, $B$
  constructed with Equations~\ref{eqn:AijHead}, \ref{eqn:AijBody},
  \ref{eqn:AijPCWA} and~\ref{eqn:Bi}, \ref{eqn:Bi0}, \ref{eqn:BiPCWA},
  \respectively. Then $\mc{R}$ is P-CWA consistent, $\pi$ a P-CWA
  solution.
  
\end{theorem}

Theorem~\ref{thm:localForGlobal} sanctions the correctness of coding
the P-CWA criterion with local constraints. The proof of this theorem
shown in \ref{sec:proof} is long-winded. However, the idea is
simple. We first observe that the ``global'' constraints given by the 
P-CWA definition, which are defined \wrt{} deductions, require us
setting $\Pr(\omega_G) = 0$ for some $\omega_G \in \Omega$; and the
``local'' constraints, given by Equations~\ref{eqn:AijPCWA} and
\ref{eqn:BiPCWA}, which only need information in the level of p-rules,
also set $\Pr(\omega_L) = 0$ for some $\omega_L \in \Omega$.
This theorem states that $\omega_G$s and $\omega_L$s are the same set
of atomic conjunctions. This is the case as shown by our induction
proof that: 
\begin{enumerate}
\item
  when each deduction contains a single p-rule, it is obvious to see
  that the set of $\omega_L$s is the same set of $\omega_G$s;
\item
  when a deduction contains multiple p-rules, assume it is the
  case that $\omega_L$s = $\omega_G$s, then introducing any a new
  p-rule will not break the equality. This is the case because for any
  $\omega$ that is set to $\Pr(\omega)=0$ by the global constraint but
  not the local constraint defined by the new p-rule, we can find an
  existing p-rule that sets $\Pr(\omega) = 0$ for the same $\omega$.

  To see this, we observe that for a new p-rule of the form
  $$\rho = \prule{\sent_0 \gets \sent_1, \ldots, \sent_n}{\cdot}$$
  with deduction
  $$\argA = \arguD{\{\sent_0, \ldots, \sent_n, \sent_{n+1}, \ldots, \sent_{n+m}\}}{\sent_0},$$
  the $\omega$ that is set to $\Pr(\omega) = 0$ by $\argA$ but not $\rho$ is
  of the form
  $$\omega = \sent_0 \wedge \ldots \wedge \sent_n \wedge \ldots \wedge
  \neg \sent_k \wedge \ldots$$
  in which $\sent_k \in \{\sent_{n+1}, \ldots,
  \sent_{n+m}\}$. However, $\Pr(\omega) = 0$ will be set for such
  $\omega$ by the p-rule
  $$\rho' = \prule{\sent^* \gets \sent_k, \ldots}{\cdot}.$$
  $\sent^*$ and $\rho'$ must both exist in $\argA$ as without them,
  there would not be $\sent_k \in \argA$. With the local constraint,
  $\rho'$ will set $\Pr(\sent^* \wedge \neg \sent_k \wedge \ldots) = 0$.
\end{enumerate}

\subsection{Compute Maximum Entropy Solutions}
\label{subsec:ComputeMES}

To compute the maximum entropy distribution introduced in
Section~\ref{subsec:MES}, we use

{\em Maximize:} 
\begin{equation}
  \label{eqn:maxEntropyObj}
  H(\pi_1,\ldots,\pi_{2^n}) = -\sum_{i=1}^{2^n} \pi_i
  \log(\pi_i),   
\end{equation}

{\em subject to:}
\begin{align*}
  A\pi &= B, \\
  0    &\leq \pi_i.
\end{align*}
As the objective function
$H$ is
quadratic, it can no longer be solved with linear programming
techniques. Thus, one possibility is to use a quadratic programming
(QP) approach, such as a trust-region method \cite{Yuan15}, which
can optimise quadratic objective functions with linear constraints and
allow specification of variable bounds.


Alternatively, to maintain linearity hence reducing complexity,
instead of maximising the von Neumann entropy $H$
(Equation~\ref{eqn:maxEntropyObj}), we can use the {\em linear
  entropy} \cite{Buscemi07}, which approximates $\log(x)$ with $x-1$
(the first term in the Taylor series of $\log(x)$), and maximise
\begin{equation}
  \label{eqn:maxEntropyObj}
  H_l(\pi_1,\ldots,\pi_{2^n}) = -\sum_{i=1}^{2^n} \pi_i (1 - \pi_i),   
\end{equation}
with Lagrange multipliers \cite{Beveridge70} as follows.

Consider $A$ as a column vector of $m$ row vectors
\begin{equation*}
  A=\begin{bmatrix}
  \bm{a_1} \\
  \vdots   \\
  \bm{a_m}
  \end{bmatrix},
\end{equation*}
\noindent
and $B = [b_1,\ldots,b_m]^T$. Define an auxiliary function $L$:
\begin{equation*}
  L(\pi_1,\ldots,\pi_{2^n},\lambda_1,\ldots,\lambda_m) =
  H_l(\pi_1,\ldots,\pi_{2^n})-\sum_{i=1}^m\lambda_i(\bm{a_i}\cdot \pi - b_i).
\end{equation*}

\noindent
$\lambda_1,\ldots,\lambda_m$ are the Lagrange multipliers. We need to
solve
\begin{equation*}
  \nabla_{\pi_1,\ldots,\pi_{2^n},\lambda_1,\ldots,\lambda_m}L(\pi_1,\ldots,\pi_{2^n},\lambda_1,\ldots,\lambda_m)
  = 0.
\end{equation*} 
\noindent
This amounts to solve the following $m+2^n$ equations:

For $i = 1 \ldots m$:
\begin{equation}
  \label{eqn:APiB_Elem}
  \bm{a_i} \cdot \pi - b_i = 0.
\end{equation}

For $j = 1 \ldots 2^n$:
\begin{equation}
  \label{eqn:linearEntropy}
  \pi_j - \sum_{i=1}^m \lambda_i\bm{a_{i,j}} = 0.
\end{equation}
Together with the remaining $m$ equations
(Equation~\ref{eqn:APiB_Elem}), we have a new 
system that gives a
maximum linear entropy solution to $A\pi=B$.

In a matrix form, we have

\begin{equation}
    \fixTABwidth{T}
    \bracketMatrixstack{
      A      & 0 \\
      I_{2^n} & -A^T
    }
    \bracketMatrixstack{
      \pi \\
      \lambda
    }
    =
    \bracketMatrixstack{
      B \\
      0
    },
\end{equation}
where $I_{2^n}$ is the $2^n$-by-$2^n$ identity matrix, and $\lambda =
[\lambda_1, \ldots, \lambda_m]$.

\begin{example}
  \label{exp:linearEntropyError}
  Consider a p-rule set with two p-rules
  \begin{center}  
  $\prule{\sent_0 \gets \sent_1}{0.9}$ and $\prule{\sent_1 \gets}{0.8}$.
  \end{center}
  To find the
  maximum linear entropy solution of $\pi$ using Lagrange multipliers,
  we start by constructing matrices $A$ and $B$ as:
  \begin{equation*}
    \fixTABwidth{T}
    A = \bracketMatrixstack{
      0 & 0.9 & 0 & -0.1 \\
      0 & 1   & 0 & 1 \\
      1 & 1       & 1 & 1     
    },
  \end{equation*}
  $B = [0,0.8,1]^T$. Thus, $m=3$ and
  $$\bm{a_1} = [0,0.9,0,-0.1], \bm{a_2} = [0,1,0,1], \bm{a_3} = [1,1,1,1].$$

  Using Equations~\ref{eqn:linearEntropy} and \ref{eqn:APiB_Elem}, 
  we have a linear system with 7 equations to solve:
  \begin{align*}
    0.9\pi_2 - 0.1\pi_4 &= 0 \\
    \pi_2 + \pi_4 &= 0.8 \\    
    \pi_1 + \pi_2 + \pi_3 + \pi_4 &= 1 \\
    \pi_1 - \lambda_3 &= 0 \\
    \pi_2 - 0.9\lambda_1 - \lambda_2 - \lambda_3 &= 0 \\
    \pi_3 - \lambda_3 &= 0 \\
    \pi_4 + 0.1\lambda_1 - \lambda_2 - \lambda_3 &= 0     
  \end{align*}
  In matrix form, we have:
  \begin{equation*}
    \fixTABwidth{T}
    \bracketMatrixstack{
      0 & 0.9 & 0 & -0.1 & 0    & 0  & 0\\
      0 & 1   & 0 & 1    & 0    & 0  & 0\\
      1 & 1   & 1 & 1    & 0    & 0  & 0\\
      1 & 0   & 0 & 0    & 0    & 0  & -1\\      
      0 & 1   & 0 & 0    & -0.9 & -1 & -1\\
      0 & 0   & 1 & 0    & 0    & 0  & -1\\      
      0 & 0   & 0 & 1    & 0.1 & -1 & -1
    }
    \bracketMatrixstack{
      \pi_1 \\
      \pi_2 \\
      \pi_3 \\
      \pi_4 \\
      \lambda_1 \\
      \lambda_2 \\
      \lambda_3
    }
    =
    \bracketMatrixstack{
      0\\
      0.8\\
      1\\
      0\\
      0\\
      0\\
      0
    }.   
  \end{equation*}
  Solve these, we find:
  \begin{center}
    \begin{tabular}{llll}
      $\pi_1 = 0.1,$ & $\pi_2 = 0.08,$ &
      $\pi_3 = 0.1,$ & $\pi_4 = 0.72$, \\
      $\lambda_1 = -0.64,$ & $\lambda_2 = 0.556,$ & $\lambda_3 = 0.1.$
  \end{tabular}
  \end{center}
  Drop auxiliary variables $\lambda_1$, $\lambda_2$ and $\lambda_3$. 
  From $\pi_1$ to $\pi_4$, we find $\Pr(\sent_0) = 0.82$, and
  $\Pr(\sent_1) = 0.8$.
\end{example}

\begin{figure}
  \begin{center}
    \includegraphics[trim = 0cm 0cm 0cm 0cm, width=0.6\textwidth]{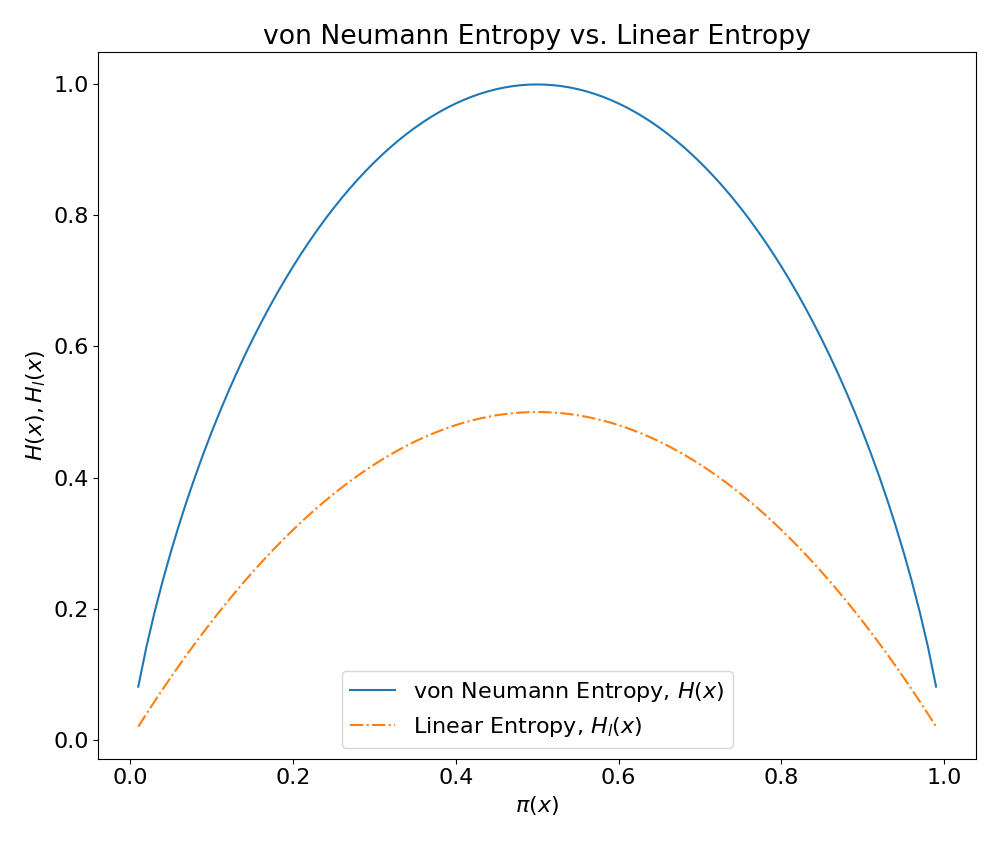}
    \caption{Linear Entropy vs. von Neumann Entropy Illustration for a
      language containing a single
      literal. \label{fig:linearVSvonNeumann}} 
  \end{center}  
\end{figure}

As illustrate in Figure~\ref{fig:linearVSvonNeumann}, linear entropy,
$H_l$, gives a lower bound to von Neumann entropy $H$. More
importantly, both $H$ and $H_l$ attain their maxima when the
probabilities are most equally distributed. It is easy to see that the
distribution that maximize linear entropy also maximizes von Neumann
entropy; thus we do not lose accuracy while maximizing linear entropy.

\subsection{Compute Solutions with Stochastic Gradient Descent}
\label{subsec:SGD}

So far, all discussions on calculating the joint distribution is
centered on solving the linear system
$$A\pi = B$$
with different constructions of $A$ and $B$.
Although linear programming with {\em linprog} computes solutions,
it is computationally expensive. To have a more efficient approach, we
consider a stochastic gradient descent (SGD) method for solving $A\pi
= B$ as follows. 

Let $A = [\bm{a_1}, \ldots, \bm{a_m}]^T$, for $i = 1,\ldots,m$, define
$$h_{\pi}(\bm{a}_i) = \bm{a}_i \cdot \pi.$$
 
\noindent
Consider the squared loss function $L$:
$$L(h_{\pi}(\bm{a}_i)) = \sum^m_{i=1}(\bm{a}_i \cdot \pi - b_i)^2.$$

\noindent
Then, solve $A\pi = B$ is to find $\pi^*$, \st{}
\begin{align*}
  \pi^* &= \argmin_\pi \sum_{i=1}^mL(h_{\pi}(\bm{a}_i))\\
        &= \argmin_\pi \sum^m_{i=1}(\bm{a}_i \cdot \pi - b_i)^2.
\end{align*}

\noindent
Use SGD to find the minimum point. For some
initial $\pi = [\pi_1, \ldots, \pi_{2^n}]$, loop $i$ in $1,\ldots,m$,
each $\pi_j$ in $\pi$ is updated iteratively with
\begin{equation}
  \label{eqn:update}
  \pi_j \Leftarrow \min(1,\max(0,\pi_j + \Delta\pi_j))
\end{equation}

\noindent
in which 
\begin{align*}
  \Delta\pi_j &= \learnRate \times \frac{\partial}{\partial \pi_j} (\bm{a}_i\cdot\pi - b_i)^2 \\
              &= \learnRate \times 2(\bm{a}_i\cdot\pi - b_i)\frac{\partial}{\partial \pi_j} (\bm{a}_i\cdot\pi - b_i)\\
              &= \learnRate \times 2(\bm{a}_i\cdot\pi - b_i)\bm{a}_{ij},
\end{align*}
where $\learnRate$ is the learning rate (a small positive number).

Such a root finding process can be viewed as training an unthresholded
perceptron model \cite{Russell2009} without activation function using
SGD. Each $\pi_i$ is bounded in $[0,1]$ in the updating step
(Equation~\ref{eqn:update}). Note that this is a generic method for
solving linear systems. It does not rely on any specific construction
of $A$ or $B$. Thus, to compute maximum entropy solutions, we can use
this approach to solve the linear system composed of
Equation~\ref{eqn:APiB_Elem} and \ref{eqn:linearEntropy}.

A prominent advantage of SGD based approach is the ability to control
the error $\zeta = A\pi - B$ at run time, so the gradient descent loop
terminates when $|\zeta|$ is ``small enough''. Moreover, as SGD can be
easily parallelized on GPU, its performance can be improved
significantly.

\subsection{Computational Performance Studies}
\label{subsec:sgdImplementation}

To study the performance of presented Rule-PSAT algorithms, we
experiment them on randomly generated p-rules. Given a language
$\mc{L}$, to ensure that p-rule sets defined over $\mc{L}$ are
Rule-PSAT, we first  generate a random distribution $\pi$ over the CC
set of $\mc{L}$ by drawing samples from a discrete uniform
distribution. Then we generate p-rules by randomly selecting literals
from $\mc{L}$ to be the head and body of p-rules. The length of each
p-rule is randomly selected between 1 and the size of the language.
The probability of each generated p-rule is then
computed from $\pi$ with Equations~\ref{eqn:dfnq3} and \ref{eqn:dfnq4}.   

We separate our experiments into two groups, approaches that find a
solution to $A\pi = B$ and approaches that find maximum
entropy solutions. Figure~\ref{fig:perfLChange} shows the solver
performances for the first gruop: LP, SGD (CPU) and SGD (GPU). In these
experiments, the size of the language ranges from $n=6$ to $n=20$; the
lengths of p-rule sets are 64, 128, 256 and 512, respectively. The
termination condition for SGD (with and without GPU) is $|\zeta| <
10^{-3}$. All experiments are conducted on a desktop PC with a Ryzen
2700 CPU (16 cores), 64GB RAM and a Nvidia 3090 GPU (24GB RAM). From
this figure, we observe that although LP is faster than SGD when the
size of the language is smaller than 10, SGD is significantly faster
as the size of $\mc{L}$ grows, especially when the GPU implementation
is used.

\begin{figure}
  \begin{center}
    \includegraphics[trim = 0cm 0cm 0cm 0cm, width=\textwidth]{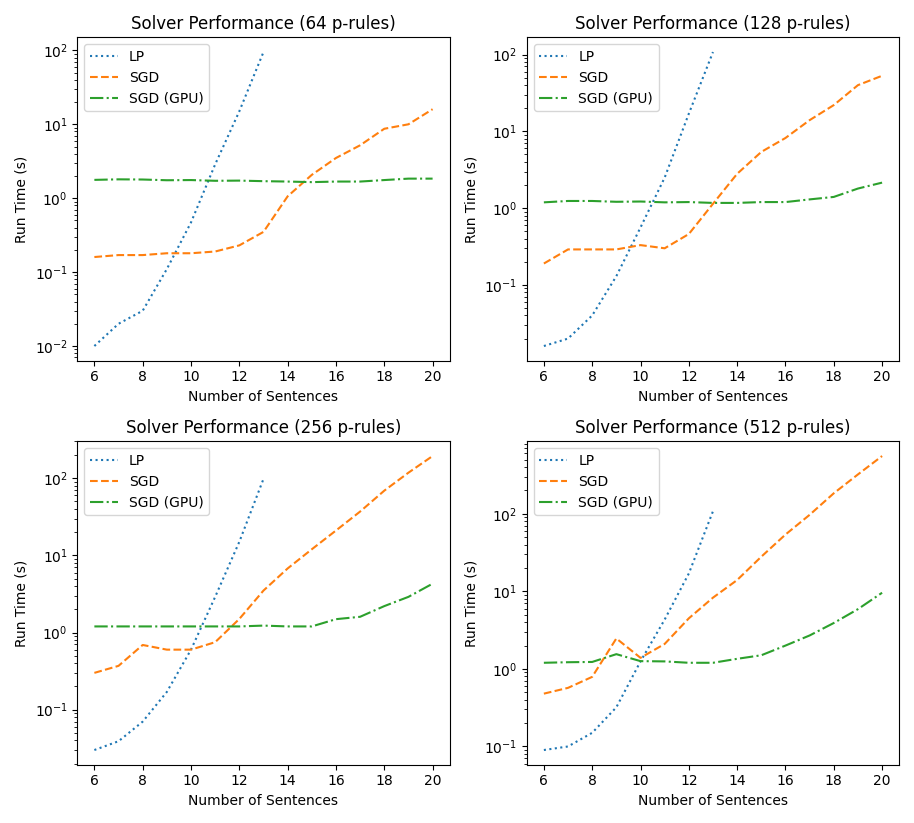}
    \caption{Solver Performance for algorithms that find a solution to
      $A\pi = B$. \label{fig:perfLChange}}
  \end{center}
\end{figure}

To study performances of approaches that compute maximum entropy
solutions, we first compare the entropy of solutions found by these
approaches with results shown in Table~\ref{table:entropy}. We observe
that approaches that optimize for linear entropy find the same
solutions as the QP approach that maximizes von Neumann entropy, as
expected. 
The small differences between these methods are likely caused by
numerical errors. For these experiments, the size of p-rule sets is 16.
Figure~\ref{fig:ME} presents the solver performance in terms of
speed. We see that although SGD is slower than LP and QP initially, it
over takes LP and QP as the size of the $\mc{L}$ grows.

\begin{table}
  \begin{center}
    \caption{Entropies of Solutions found by different
      Approaches with Languages of difference Sizes.\label{table:entropy}}
    \begin{small}
    \begin{tabular}{|c|c|c|c|c|c|}
      \hline
      Method & $|\mc{L}|$ = 6 & $|\mc{L}|$ = 7 & $|\mc{L}|$ = 8 & $|\mc{L}|$ = 9 & $|\mc{L}|$ = 10\\
      \hline
      QP maximize $-\sum \pi_i\log(\pi_i)$ & 4.101 & 4.820 & 5.529 & 6.235 & 6.928 \\
      LP maximize $-\sum \pi_i(\pi_i-1)$ & 4.103 & 4.821 & 5.529 & 6.235 & 6.928\\
      SGD maximize $-\sum \pi_i(\pi_i-1) $ & 4.101 & 4.824 & 5.529 & 6.235 & 6.929\\
      \hline
    \end{tabular}
    \end{small}    
  \end{center}
\end{table}

\begin{figure}
  \begin{center}
    \includegraphics[trim = 0cm 0cm 0cm 0cm, width=\textwidth]{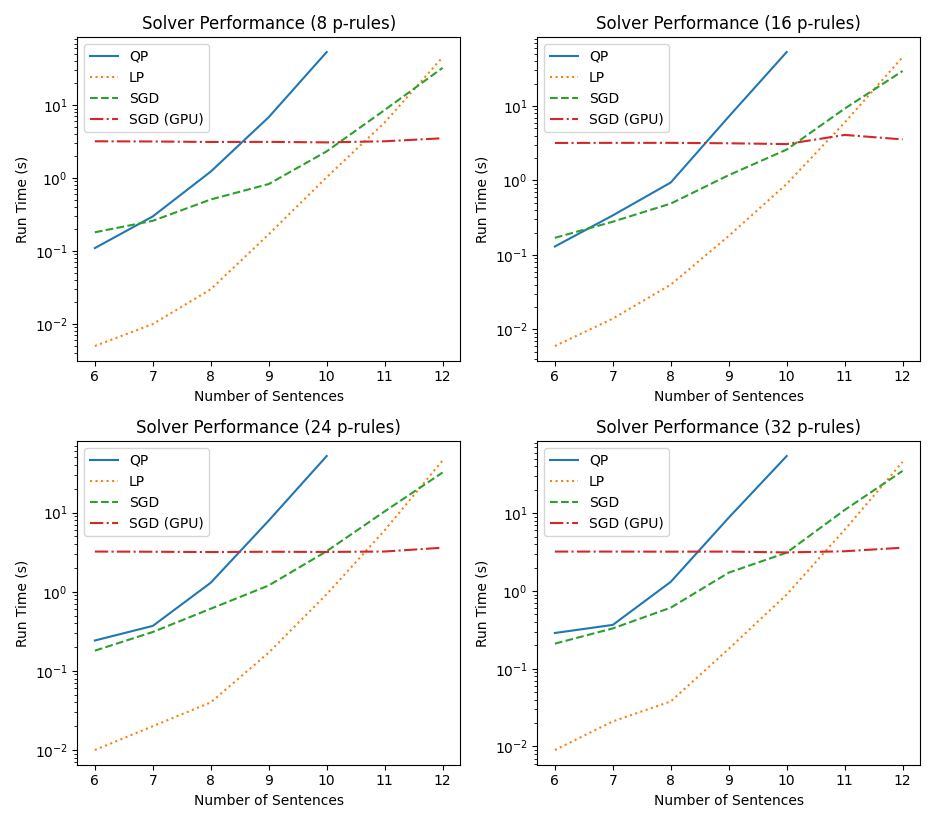}
    \caption{Solver Performance for algorithms that find maximum
      entropy solution to $A\pi = B$. \label{fig:ME}} 
  \end{center}
\end{figure}

In all experiments, the learning rate $\learnRate$ in our SGD
implementations are set to $1/2^{n}$, where $n$ is the size of the
language. A common technique, momentum \cite{Mitchell97}, in training
neural networks have been applied to speed up the SGD convergence
rate. Namely, in each iteration, $\Delta\pi_j$ is updated with
\begin{equation*}
  \Delta\pi_j \Leftarrow \learnRate \times 2(\bm{a}_i\cdot\pi - b_i)\bm{a}_{ij} + \alpha \Delta\pi_j,
\end{equation*}
where $\alpha = 0.99$ is the momentum used in all experiments. 
$\Delta\pi_j$ on the RHS is $\Delta\pi_j$ calculated in the previous
iteration.

In summary, Table~\ref{table:solverSummary} presents characteristics of the 
Rule-PSAT solving approaches studied in this work. We see that SGD
approaches with GPU implementation significantly outperform LP and QP
methods in terms of scalability. 

\begin{table}
  \begin{center}  
  \caption{Characteristics of Rule-PSAT Solving
    Approaches introduced in this section.\label{table:solverSummary}}
  \begin{tabular}{|c|c|c|}
    \hline
                    & Exact       & Maximum \\
    Method & Solution & Entropy Solution    \\
    \hline
    LP solve $A\pi=B$ & Yes & No  \\
    QP maximize $-\sum \pi_i\log(\pi_i)$ & Yes & Yes (von Neumann) \\
    LP maximize $-\sum \pi_i(\pi_i-1)$ & Yes & Yes (Linear) \\
    \hline    
    SGD solve $A\pi=B$ & No & No \\
    SGD maximize $-\sum \pi_i(\pi_i-1) $ & No & Yes (Linear) \\
    \hline
    SGD solve $A\pi=B$ (GPU) & No & No \\        
    SGD maximize $-\sum \pi_i(\pi_i-1)$ (GPU) & No & Yes (Linear)\\ 
    \hline
  \end{tabular}
  \end{center}  
\end{table}

\section{Discussion and Conclusion}
\label{sec:conclusion}

In this work, we have presented a novel probabilistic structured
argumentation framework, Probabilistic Deduction (PD). Syntactically,
PD frameworks are defined with probabilistic rules (p-rules) in the
form of 
$$\prule{\sent_0 \gets \sent_1, \ldots, \sent_n}{\probParameter},$$
which is read as conditional probabilities
$\Pr(\sent_0|\sent_1,\ldots,\sent_n) = \probParameter$. 
To reason with p-rules, we solve the rule probabilistic satisfiability
problem to find the joint probability distribution over the language
defining the p-rules and then compute literal probabilities for
literals in the language. We have introduced two different
formulations for this process, the probabilistic open-world assumption
(P-OWA) and probabilistic closed-world assumption (P-CWA). With P-OWA,
the joint probability is solved based on conditional probabilities
defined by p-rules; with P-CWA, additional constraints are introduced
to assert that the probability of a literal is the sum of all possible
worlds that contain a deduction to the literal. 

From p-rules, we build arguments as deductions in the way that the
leaves of a deduction are either literals that are heads of p-rules
with empty bodies or literals for which there are p-rules for their
negations. One argument attacks another when the claim of the former
is the negation of some literal in the latter. The main technical
achievement in this part is that we prove that with maximum entropy
reasoning, our probability semantics with P-CWA coincide with the
complete semantics defined for non-probabilistic argumentation. We
prove this abstract argumentation with mappings from AA frameworks to
PD presented. 

Solving Rule-PSAT is at the core of reasoning with PD. We have
investigated several different approaches for doing this using
linear programming, quadratic programming and stochastic gradient
descent. We have conducted experiments with these approaches on p-rule
sets built on different sizes of languages and with different numbers
of p-rules. We observe that stochastic gradient descent with GPU
implementation outperforms all other approaches as the size of the
language grows.

\subsection{Relations with some Existing Works}
\label{subsec:relation}

As discussed in \cite{Fan22}, Rule-PSAT is a variation of the
probabilistic satisfiability (PSAT) problem introduced by Nilsson in
\cite{Nilsson86}. Nilsson considered knowledge bases in Conjunctive Normal
Form.  A modus ponens example,\footnote{This example is used in
\cite{Nilsson86}. The figure on the left hand side of
Table~\ref{table:vsNilsson} is a reproduction of Figure 2 in
\cite{Nilsson86}.}

\begin{center}
If $\sent_1$, then $\sent_0$.
$\sent_1$.
Therefore, $\sent_0$.
\end{center}

\noindent
is shown in Table~\ref{table:vsNilsson}. The probabilities of the
conditional claim is $\alpha$, the antecedent $\beta$ and the
consequent $\gamma$. With Nilsson's probabilistic logic, this is
interpreated as: 

\begin{center}
\begin{tabular}{ccc}
  $\prule{\neg \sent_1 \vee \sent_0}{\alpha}$, &
  $\prule{\sent_1}{\beta}$, &
  $\prule{\sent_0}{\gamma}$,
\end{tabular}
\end{center}

\noindent
which gives rise to equations
\begin{align}
  \pi(\neg \sent_1 \wedge \sent_0)  + \pi(\sent_1 \wedge
  \sent_0)  + \pi(\neg \sent_1 \wedge \neg \sent_0) &= \alpha, \\
  \pi(\sent_1 \wedge \sent_0) + \pi(\sent_1 \wedge \neg \sent_0) &= \beta, \label{eqn:sent_1_beta}\\
  \pi(\sent_1 \wedge \sent_0) + \pi(\neg \sent_1 \wedge \sent_0) &= \gamma.\label{eqn:sent_0_gamma}
\end{align}

\noindent
With probabilistic rules discussed in this work, the interpreatation
to modus ponens is the three p-rules as follows.

\begin{center}
\begin{tabular}{ccc}
  $\prule{\sent_0 \gets \sent_1}{\alpha}$, &
  $\prule{\sent_1 \gets}{\beta}$, &
  $\prule{\sent_0 \gets}{\gamma}$,
\end{tabular}
\end{center}

\noindent
which gives rise to equations
\begin{align}
      \frac{\pi(\sent_0 \wedge \sent_1)}{
      \pi(\neg \sent_0 \wedge \sent_1) +
      \pi(\sent_0 \wedge \sent_1)} &= \alpha,
\end{align}
\ref{eqn:sent_1_beta} and \ref{eqn:sent_0_gamma}.
The two shaded polyhedrons shown in
Table~\ref{table:vsNilsson} illustrate probabilistic consistent regions
for $\alpha, \beta$ and $\gamma$, with probabilistic
logic and probabilistic rule, respectively, as defined by their
corresponding equations together with equations \ref{eqn:dfnq1} and
\ref{eqn:dfnq2}.  The consistent region in the probabilistic logic case
is a tetrahedron, with vertices (0,0,1), (1,0,0), (1,1,0) and
(1,1,1). The consistent region in the probabilistic rule case is an
octahedron, with vertices (0,0,0), (0,0,1), (0,1,0), (1,0,0), (1,1,0)
and (1,1,1). It is argued in \cite{Nilsson93} that the conditional
probability interpretation to modus ponens is more reasonable than the
probabilistic logic interpretation in practical settings.

\begin{table}
  \caption{Comparison of Consistent Probability Regions between
    Nilsson's Probabilistic Logic and Probabilistic Rules on an
    modus ponens instance. \cite{Fan22}\label{table:vsNilsson}}
  \begin{center}
    \begin{tabular}{|c|c|}
      \hline
      Probabilistic Logic & Probabilistic Rule \\
      \hline
      \hline
      $\prule{\neg \sent_1 \vee \sent_0}{\alpha}$, \hspace{0.3cm}
      $\prule{\sent_1}{\beta}$, \hspace{0.3cm}
      $\prule{\sent_0}{\gamma}$.
      &
      $\prule{\sent_0 \gets \sent_1}{\alpha}$, 
      $\prule{\sent_1 \gets}{\beta}$, 
      $\prule{\sent_0 \gets}{\gamma}$.\\      
      \hline      
      \includegraphics[trim = 0cm 0cm 0cm -1cm, width=5.7cm]{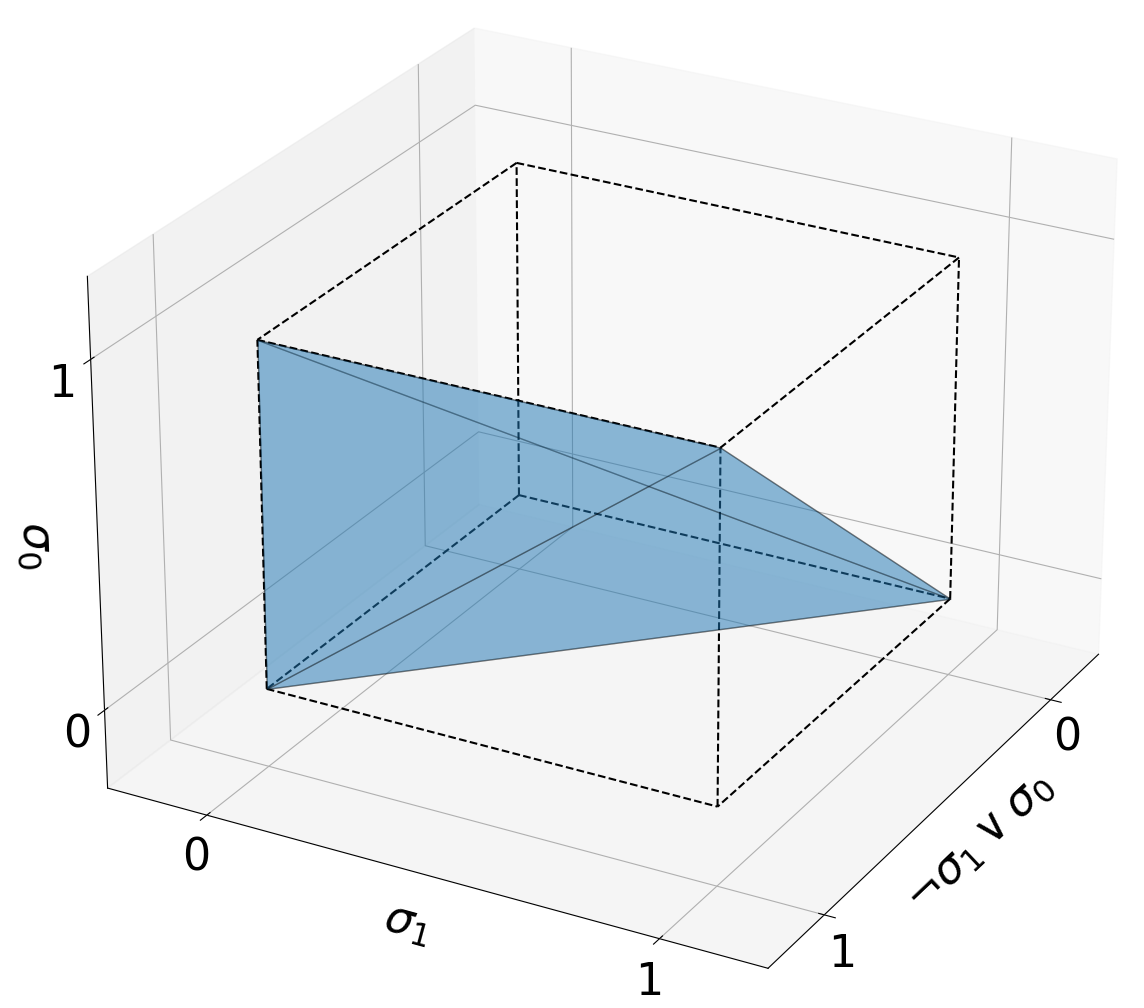}
      & \includegraphics[trim = 0cm 0cm 0cm 5cm, width=5.7cm]{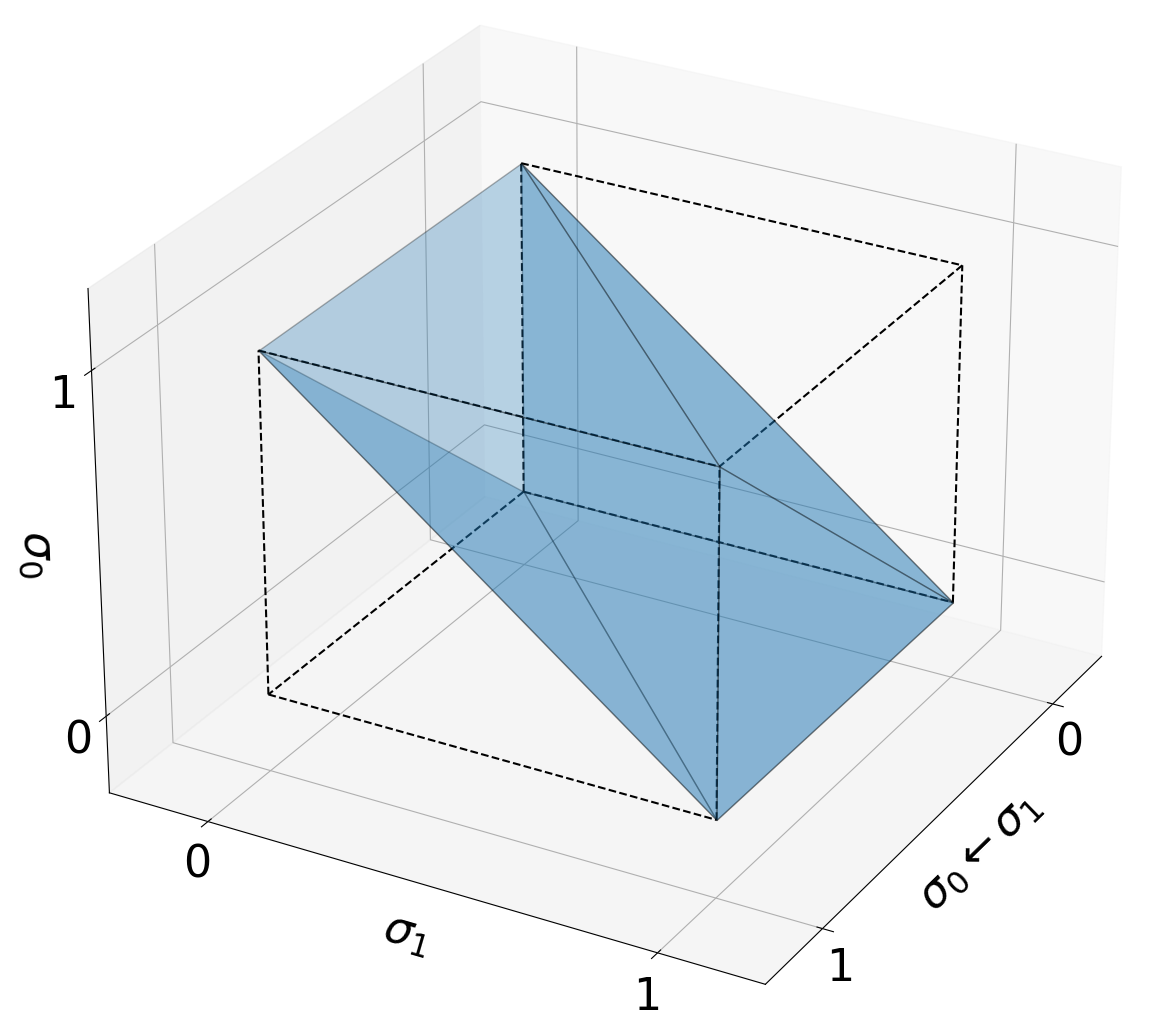}\\
      \hline
    \end{tabular}
  \end{center}  
\end{table}

From this example, we observe that both methods are nothing but
imposing constraints on the feasible regions of the spaces defined by
clauses (in the case of probabilistic logic) or p-rules (in the case
of probabilistic rules). In this sense, reasoning on such probability
and logic combined forms is about identifying feasible regions
determined by solutions to $\Pi$ in $A\Pi = B$.\footnote{Constructions
of $A$ differ between Nilsson's probabilistic logic and this
work. However, both are designed for solving the joint probability
distribution over the CC set.} 

Hunter and Liu~\cite{HunterL10} make an interesting observation on
representing scientific knowledge by combining probabilistic reasoning
with logical reasoning. Quoting from \cite{HunterL10}:
\begin{quote}
A key shortcoming of extending classical logic in order to handle
probabilistic or statistical information, either by using a possible
worlds approach or by adding a probability distribution to each model,
is the computational complexity that it involves.  
\end{quote}
They suggest one may circumvent the computation of joint
probability distribution by considering approaches such as Bayesian
networks. We certainly agree that computational difficulty is a major
challenge. On the other hand, as \cite{Georgakopoulos88} have shown
that the problem of PSAT is NP-complete, thus there does not exist a
shortcut that performs probabilistic reasoning ``correctly'' in
general cases. Thus, any probabilistic reasoning approach that does
not require the computation of the joint probabilistic distribution
either imposes probabilistic assumptions in the underlying model such
as independence e.g., \cite{Ma10,Henderson20}, or topological
constraints, e.g., conditional independence~\cite{philip79}, as in
Bayesian networks. In this work, we choose not to make such
assumptions and study computational approaches with optimization
techniques. 

In the landscape of probabilistic argumentation,
\cite{Thimm12,Hunter14,Hunter17} give detailed account on probabilistic
abstract argumentation with the epistemic approach. They have
described some ``desirable'' properties of probability semantics,
which can be viewed as properties imposed on the joint probability
distribution. As discussed in Section~\ref{Introduction}, the main difference
distinguishes this work with existing ones is that we do not assume a
given joint probability distribution. Yet, with our approach, we can
still compute argument probabilities and thus compare with some of the
properties they have proposed, as follows.

\begin{itemize}
\item
  {\bf COH} A probability distribution $P$ is {\em coherent} if for
  arguments $\argA$ and $\argB$, if $\argA$ attacks $\argB$, then
  $\Pr(\argA) + \Pr(\argB) \leq 1$.

  As shown in Proposition~\ref{prop:attackerAttackeeNoMoreThan1}, {\bf
    COH} holds in PD frameworks in general.

\item
  {\bf SFOU} $P$ is {\em semi-founded} if $\Pr(\argA) \geq 0.5$ for
  every $\argA$ not attacked.

  This is not true in general PD frameworks, as one can use a p-rule
  $$\prule{\sent_0 \gets}{0.2},$$
  to build an un-attacked argument $\argA =
  \argu{\{\sent_0\}}{\sent_0}$, $\Pr(\argA) = 0.2 \leq 0.5$. However,
  SFOU holds for AA-PD frameworks, as shown by
  Proposition~\ref{prop:nonAttackedProb1}. 

\item
  {\bf FOU} $P$ is {\em founded} if all un-attacked argument have
  probability 1.

  This is not true in general PD frameworks, but true for AA-PD
  frameworks, as shown by Proposition~\ref{prop:nonAttackedProb1}.

\item
  {\bf SOPT} $P$ is {\em semi-optimistic} if $\Pr(\argA) \geq 1 -
  \sum_{\argB \in \argsB} \Pr(\argB)$, where $\argsB \neq \{\}$ is the
  set of arguments attacking $\argA$.

  This is not true in general PD frameworks, as demonstrated in
  Example~\ref{exp:2ndCondPjustifiable}. This is true for AA-PD
  framework as shown by Proposition~\ref{prop:optimistic}.

\item
  {\bf OPT} $P$ is {\em optimistic} if $\Pr(\argA) \geq 1 -
  \sum_{\argB \in \argsB} \Pr(\argB)$, where $\argsB$ is the
  set of arguments attacking $\argA$.

  As in the previous case, this is not true in general PD framework
  but true for AA-PD frameworks by Proposition~\ref{prop:optimistic}.

\item
  {\bf JUS} $P$ is {\em justifiable} if $P$ is coherent and
  optimistic.

  Since PD frameworks are not optimistic in general, they are not
  justifiable. AA-PD frameworks are justifiable.

\item
  {\bf TER} $P$ is {\em ternary} if $\Pr(\argA) \in \{0, 0.5, 1\}$ for
  all $\argA$.

  This is not true in general PD frameworks as shown in e.g.,
  Example~\ref{exp:opening}. This is not true for AA-PD framework
  either, as illustrated in Example~\ref{exp:noME}.
\end{itemize}

This comparison is encouraging as one can take these properties
introduced by Hunter and Thimm as a benchmark for probabilistic
argumentation semantics. Observing AA-PD frameworks, a subset of PD
frameworks, confirm to these properties helps us to see the underlying
connection between PD frameworks and existing works on probabilistic
argumentation. At the same, since general PD frameworks do not confirm
to the founded and optimistic properties, we observe the hierarchical
structure as shown in Figure~\ref{fig:hierarchy}.

\begin{figure}
  \begin{center}
  \includegraphics[trim = 0cm 0cm 0cm 0cm, width=0.4\textwidth]{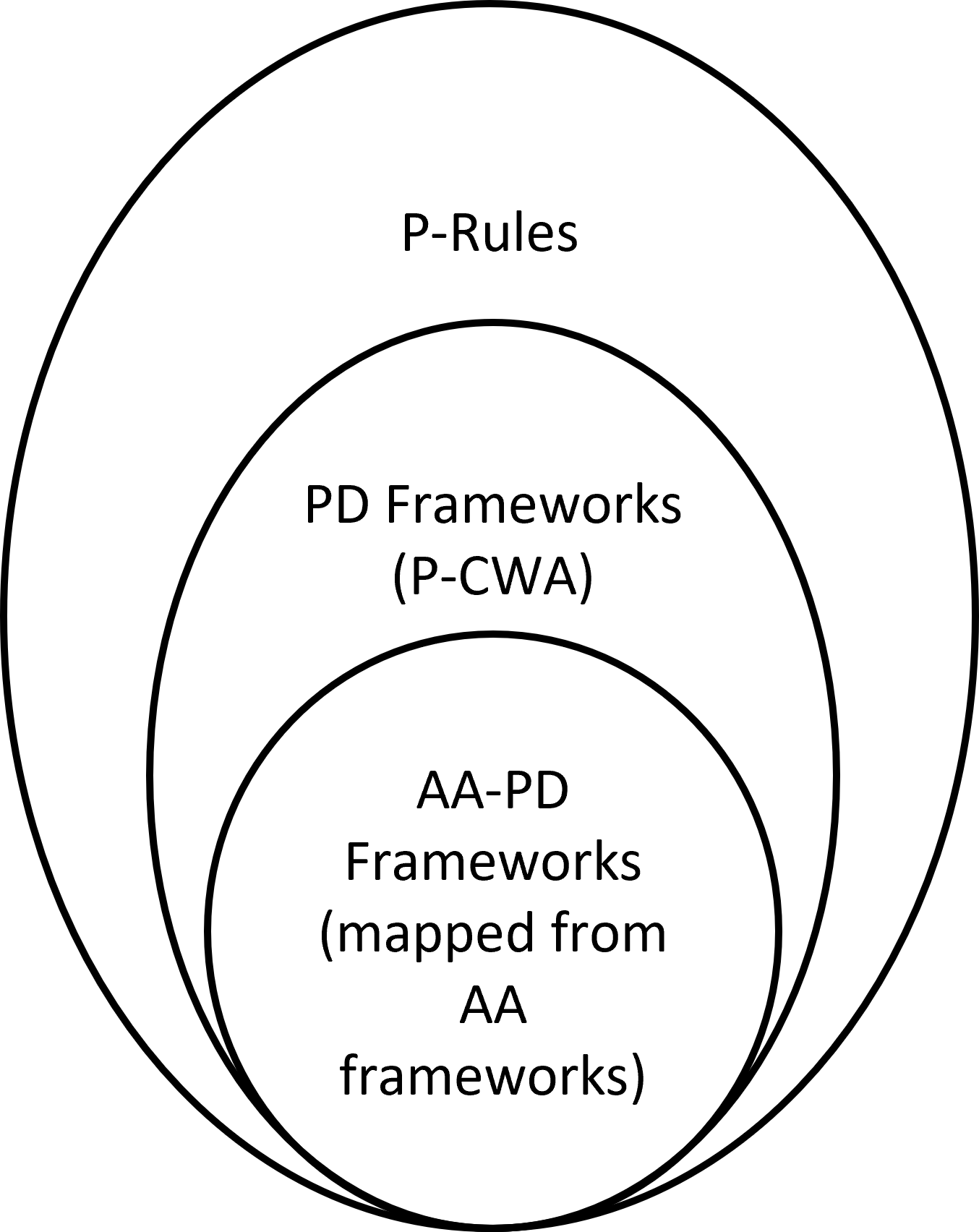}  
  \caption{The hierarchical structure between p-rules, PD frameworks
    and AA-PD frameworks.\label{fig:hierarchy}}
  \end{center}  
\end{figure}

PD frameworks share some similarities with Probabilistic
Assumption-based Argumentation (PABA)
\cite{Dung10,Hung17,Cyras21}, syntactically. \cite{Fan22} has
presented differences  between p-rules and PABA. Namely, PABA
disallows rules forming cycles and if two rules have the same head,
the body of one must be a subset of the other; whereas p-rules do not
have these constraints. PD frameworks and AA-PD frameworks introduced
in this work do not have these constraints. More fundamentally, PABA
is an constellation approach to probabilistic argumentation
\cite{Cyras21}, whereas PD is an epistemic approach.

\subsection{Future Work}
Moving forward, there are three main research directions we will
explore in the future. Firstly, as briefly explained in
Section~\ref{sec:inconsistPD}, the current approaches for computing
literal probability requires either a Rule-PSAT solution (for
reasoning with P-OWA) or a P-CWA consistent solution (for reasoning
with P-CWA) found on the joint probability distribution. However,
such requirement renders ``local'' reasoning impossible in the sense
that one cannot deduce the probability of any literal in an
inconsistent set of p-rules even if the literal of interest is
independent of the of subset of p-rules that are inconsistent. (This is
not much different from observing inconsistency in classical logic in
the sense that with a classical logic knowledge base, having both $p$
and $\neg p$ co-exist trivializes the knowledge base.) In the future,
we would like to explore probability semantics for such inconsistent
set of p-rules as well as their computational counterparts. 

Secondly, even though we have shown that solving Rule-PSAT with SGD is
a promising direction when compared with other approaches such as LP
and QP, we are aware that the number of unknowns grows exponentially
with respect to the size of the language. Thus, we would like to
explore techniques that do not explicitly compute the $2^n$
unknowns defining the joint probability distribution, as suggested by
e.g., \cite{HunterL10}. To this end, there are two directions we will
explore. (1) Inspired by the column generation method that is commonly
used in optimization, we would like to see whether a similar
technique can be developed for reasoning with p-rules; and (2)
investigate the existence of equivalent ``local'' semantics in
addition to the ``global'' semantics given in this work for literal
probability computation, especially in cases where the given PD
framework can be assumed to be P-CWA consistent and P-CWA can be
assumed. 

Lastly, we would like to explore applications of PD. As a generic
structured probabilistic argumentation framework, we believe the
practical limits of PD can be best understood by experimenting it with
applications from different domains. Just as ABA has seen its
applications in areas such as decision making and planning, we believe
PD with its ability in handling probabilistic information could be
suitable for solving problems in such domains. We would like to
explore these potentials in the future.

\bibliographystyle{plain}
\bibliography{bibliography}

\appendix

\section{Proofs for the Results}
\label{sec:proof}
\input{proofs.tex}

\section{Miscellaneous}

\subsection{Reasoning under Inconsistency}
\label{sec:inconsistPD}

As a ``global'' semantics, to reason with any literal in a set of
p-rules $R$, we require $R$ being Rule-PSAT for computing literal and
argument probabilities. However, such consistency requirement may be
unjustified if the literal or argument of interest is independent of
the part of $R$ that is inconsistent. 

Consider the following example.

\begin{example}
  \label{exp:inconsist}

  Consider a p-rule set $R$ with three p-rules:
  \begin{center}
    \prule{\sent_0 \gets}{0.5}, 
    \prule{\sent_0 \gets}{0.6},
    \prule{\sent_1 \gets}{1}.
  \end{center}
  Clearly, $R$ is inconsistent as we cannot have $\Pr(\sent_0) = 0.5 =
  0.6$ as $0.5 \neq 0.6$. However, if one is to query $\Pr(\sent_1)$,
  one would expect $\Pr(\sent_1) = 1$ can still be returned from this
  set of inconsistent p-rules.
\end{example}

Designing a comprehensive solution for solving this inconsistent
reasoning problem is beyond the scope of this work. However, a quick
yet still reasonable approach is to directly relax the condition $A\pi
= B$. Specifically, we formulate the following optimization problem.

{\em minimize:} 
\begin{equation}
\lVert  A\pi - B  \rVert_1
\end{equation}

{\em subject to:}
\begin{align*}
  \sum_{i=1}^{2^n} \pi_i &= 1.\\
  0     &\leq \pi_i.
\end{align*}
Clearly, $\pi$ obtained as such is a probability distribution over the
CC set. From which, one can compute literal probabilities as defined
in Equation~\ref{eqn:sentProb}.

Alternatively, one can also (1) find some maximum subset $R'$
(\wrt{} $\subseteq$) of $R$ \st{} $R'$ is Rule-PSAT and compute literal
probabilities on $R'$ (this is not unlike solving the MAXSAT problem
\cite{Hansen90} in a probabilistic setting) or (2) find new consistent
probabilities $\theta$s for p-rules in $R$ \st{} they are ``close'' to
the original probabilities (this is not unlike works that ``fix''
argumentation frameworks, e.g., \cite{Ulbricht19}). We will explore
these approaches in the future. 

\end{document}

%% file: proofs.tex
\paragraph{Proof of Proposition~\ref{prop:margin}}

Since $\pi$ is a consistent probability distribution, from
Equation~\ref{eqn:dfnq1}, we know that
\begin{equation*}
0 \leq \sum_{\omega_i \in \Omega, \omega_i \models \sent} \pi(\omega_i).
\end{equation*}
From Equation~\ref{eqn:dfnq2}, we know that
\begin{equation*}
\sum_{\omega_i \in \Omega, \omega_i \models \sent} \pi(\omega_i) \leq 1.
\end{equation*}

For $\omega_i \in \Omega$, either $\omega_i \models \sent$ or
$\omega_i \models \neg\sent$, since
\begin{equation*}
  \sum_{\omega_i \in \Omega} \pi(\omega_i) = 1,
\end{equation*}
Equation~\ref{eqn:prop2} holds. \qed

\paragraph{Proof of Proposition~\ref{prop:MEunique}}

$\pi^m$ exists as the set of p-rules is consistent. By
Definition~\ref{dfn:consistency}, there exists at least one solution
$\pi$. It is unique as the set of possible solutions is convex. 

\paragraph{Proof of Lemma~\ref{lemma:MEomega}}
Since $\alpha_\omega > 0$, there are
$\omega_1', \ldots, \omega_k' \in \Omega$, \st{}
$$\Pr(\omega_1')
+ \ldots + \Pr(\omega_k') = \alpha_\omega + \beta,$$ for some
$\beta \geq 0$ and $H(\omega, \omega_1', \ldots, \omega_k')$ attains
its maximum value when
$$\Pr(\omega) = \Pr(\omega_1') = \ldots
= \Pr(\omega_k') = \frac{\omega_\alpha + \beta}{k+1}.$$ Since
$\alpha_\omega > 0$,
$$\frac{\omega_\alpha + \beta}{k+1} > 0.$$ \qed

\paragraph{Proof of Corollary~\ref{coro:MEsent}}

This follows directly from Lemma~\ref{lemma:MEomega} and
definitions of $\Pr_c(\sent)$ and $\Pr_o(\sent)$
(Equations~\ref{eqn:sentProb} and \ref{eqn:sent_prob_cwa}).

\paragraph{Proof of Proposition~\ref{prop:inconsistentSent0Prob}}

Let $\sents = \{s_1, \ldots, s_k\}$. From
Equation~\ref{eqn:argProb}, the probability of an argument is the sum
of $\Pr(\omega_i)$ \st{} $\omega_i \models s_1 \wedge \ldots \wedge
s_k$. However, since $\sents$ contains both $\sent_i$ and
$\neg \sent_i$. So there is no $\omega_i$ \st{} $\omega_i \models
s_1 \wedge \ldots \wedge s_k$. Therefore $\Pr(\argA) = 0$. \qed

\paragraph{Proof of Proposition~\ref{prop:selfAttack0Prob}}

This is a special case of
Proposition~\ref{prop:inconsistentSent0Prob}. Since both $\sent$ and
$\neg \sent$ are in $\sents$, there is no $\omega_i$ \st{}
$\omega_i \models \ldots \wedge \sent \wedge \neg \sent \ldots$. Therefore
$\Pr(A) = 0.$ \qed

\paragraph{Proof of Proposition~\ref{prop:argProbLowerThanSentProb}}

This is the case by Equation~\ref{eqn:sent_prob_cwa}
and \ref{eqn:argProb}. From Equation~\ref{eqn:sent_prob_cwa}, we see
that
\begin{equation*}
\Pr(\sent)
= \sum_{\omega_i \in \Omega, \omega_i \models \sent} \pi(\omega_i). 
\end{equation*}

Let $\sents = \{\sent_1, \ldots, \sent_n\}$ (note that
$\sent \in \sents$). Since for any
$\omega_i \models \bigwedge_{j=1}^n \sent_j$, $\omega_i \models \sent$,
we have
\begin{equation*}
\sum_{\omega_i \in \Omega, \omega_i \models \bigwedge_{j=1}^n \sent_j} \pi(\omega_i)
\leq
\sum_{\omega_i \in \Omega, \omega_i \models \sent} \pi(\omega_i).
\end{equation*}
Therefore $\Pr(\argA) \leq \Pr(\sent)$. \qed

\paragraph{Proof of Proposition~\ref{prop:argProbEqSentProb}}

Assume that such $\argB$ exists, let 
$$\argB = \argu{\{\sent \wedge \sent_1^b \wedge \sent_n^b \}}{\sent}.$$
Since $\Pr(\argB) \neq 0$,
$\Pr(\sent \wedge \sent_1^b \wedge \sent_n^b) \neq 0$.
Let
$$\argA = \argu{\{\sent \wedge \sent_1^a \wedge \sent_m^a \}}{\sent}.$$
Since $\argA \neq \argB$,
$$\{\sent_1^b, \ldots, \sent_n^b\} \neq \{\sent_1^a, \ldots, \sent_m^a\}.$$
Thus, there exists some $\omega^* \in \Omega$ \st{}
\begin{center}
$\omega^* \models \sent \wedge \sent_1^b \wedge \sent_n^b$,
$\omega^* \not\models \sent \wedge \sent_1^a \wedge \sent_m^a$ and
$\Pr(\omega^*) \neq 0$.
\end{center}
Since $$\Pr(\argA) = \sum_{\omega \in
S1} \omega,$$ for some $S1 \subset \Omega$ \st{} $\omega^* \not\in S1$,
and $$\Pr(\sent) = \sum_{\omega \in S2} \omega,$$ for some $S2 \subseteq
\Omega$ \st{} $\omega^* \in S_{S2}$, we have
$$\Pr(A) \neq \Pr(\sent).$$ Contradiction. \qed 

\paragraph{Proof of Proposition~\ref{prop:attackerAttackeeNoMoreThan1}}

Let
$$\argA = \argu{\{\sent_0^a, \ldots, \sent_{ka}^a\}}{\sent_0^a}$$ and
$$\argB = \argu{\{\sent_0^b,\ldots,\sent_{kb}^b\}}{\sent_0^b}.$$
Since $\argA$ attacks $\argB$, $\sent_0^a = \neg \sent_j^b$, for
some $j \in \{0, \ldots, kb\}$. Thus,
$$\Pr(\argA) = \Pr(\sent_0^a \wedge \ldots \wedge \sent_{ka}^a)$$
and
$$\Pr(\argB) = \Pr(\ldots \wedge \neg \sent_0^a \wedge \ldots).$$

For all $\omega \in \Omega$, either
\begin{center}
$\omega \models \sent_0^a \wedge \ldots \wedge \sent_{ka}^a$
or
$\omega \models \ldots \wedge \neg \sent_0^a \wedge \ldots.$
\end{center}
Since
$\sum_{\omega \in \Omega} \Pr(\omega) = 1$,
$$\Pr(\argA) + \Pr(\argB) \leq 1.$$ \qed

\paragraph{Proof of Proposition~\ref{prop:AA2PDOne2One}}

This is easy to see as by Definition~\ref{dfn:AA2PD}, we know that the
size of $\mc{L}$ is the same as the size of $\mc{A}$. In other words,
an argument exists in $\mc{A}$ \ifaf{} there is its counterpart in
$\aapd(F)$. Specifically, for each argument $\sent \in \mc{A}$, let
$\{\sent_1, \ldots, \sent_m\}$ be the set of arguments attacking
$\sent$ in $F$, there is $\argu{\{\sent, \neg \sent_1, \ldots, \neg
\sent_m\}}{\sent}$ in $\aapd(F)$.

Furthermore, for two arguments $\sent_a$ attacking $\sent_b$ in $F$,
meaning that $(\sent_a, \sent_b) \in \mc{T}$, we have
arguments $\argA = \argu{\{\sent_a, \ldots\}}{\sent_a}$ and
$\argB = \argu{\{\sent_b, \neg \sent_a, \ldots\}}{\sent_b}$ in
$\aapd(F)$. Clearly, $\argA$ attacks $\argB$. \qed

\paragraph{Proof of Proposition~\ref{prop:nonAttackedProb1}}
This is trivially true as arguments in a AA-PD frameworks are either
of the two forms:
\begin{center}
$\argA = \argu{\{\sent_0\}}{\sent_0}$ or $\argB
= \argu{\{\sent_0, \neg \sent_1, \ldots\}}{\sent_0}$.
\end{center}
Arguments in the form of $\argA$ are not attacked whereas in the form
of $\argB$ are attacked by some arguments. Since arguments in the form
of $\argA$ are composed by a single p-rule $\prule{\sent_0 \gets}{1}$,
we have $\Pr(\argA) = 1$.

\paragraph{Proof of Proposition~\ref{prop:zeroIfAttackedByOne}}

Let $\argA = \argu{\{\sent_1^a, \ldots, \sent_n^a\}}{\sent_1^a}$ and
$\argB = \argu{\{\sent_1^b, \ldots, \sent_m^b\}}{\sent_1^b}$.

Since
$\Pr(\argA) = \Pr(\sent_1^a \wedge \ldots \wedge \sent_n^a) =
1,$
we have $$\Pr(\sent_i^a)=1$$ for
$\sent_i^a \in \{\sent_1^a, \ldots, \sent_n^a\}$. Since $\argA$
attacks $\argB$,
$\neg \sent_1^a \in \{\sent_1^b, \ldots, \sent_m^b\}$.

Let
$\neg \sent_1^a = \sent_k^b$, for some $k \in \{1,\ldots,m\}$. We have
$$\Pr(\sent_k^b) = 1 - \Pr(\sent_1^a) = 0.$$ Since
$\Pr(\argB) = \Pr(\ldots \wedge \sent_k^b \wedge \ldots) \leq
\Pr(\sent_k^b),$
we have
$$\Pr(\argB) = 0.$$
\qed

\paragraph{Proof of Proposition~\ref{prop:oneIfAttackedByZero}}
(Sketch.)
Let
\begin{center}
$\argA = \argu{\{\sent_1^a, \ldots, \sent_n^a\}}{\sent_1^a}$, and
$\argB = \argu{\{\sent_1^b, \ldots, \sent_m^b\}}{\sent_1^b}$.
\end{center}
Since $\argB$ attacks $\argA$, there is some $\sent_k^a$ in
$\argA$ \st{} $\sent_1^b = \neg \sent_k^a$.

To show the first part, since $\args$ contains all arguments attacking
$\argA$, $\args$ contains all arguments of the form
$\argu{\_}{\sent_1^b}$, for which $\argB$ is one of these. As all of
them have probability 0, with P-CWA, $\Pr(\sent_1^b) = 0$. Thus, we
have $\Pr(\sent_k^a) = 1$. This reasoning can be applied to all
$\sent^a$ in $\argA$ such that $\neg \sent^a$ is the claim of some
argument in $\args$. For all such $\sent^a$, we have that
$\Pr(\sent^a) = 1$.

To show the second part, again we notice that
$$\argA
= \argu{\{\sent_1^a, \neg \sent_{1b}^b, \ldots, \neg \sent_{nb}^b\}}{\sent_1^a},$$
\st{} $\{\argu{\_}{\sent_{1b}^b}, \ldots, \argu{\_}{\sent_{nb}^b}\}
= \args$.
Since $\Pr(\argA) = 1$, $\Pr(\neg \sent_{1b}^b) = \ldots
= \Pr(\neg \sent_{nb}^b) = 1$, therefore $\Pr(\sent_{1b}^b) = \ldots
= \Pr(\sent_{nb}^b) = 0$. Thus for all $\argB \in \args$, we have
$\Pr(\argB) = 0$. \qed

\paragraph{Proof of Proposition~\ref{prop:arg0Attacker1}}
To show (1), from Proposition~\ref{prop:nonAttackedProb1}, we know
that $\argA$ is attacked by some argument thus $\args$ is not
empty. We show it is the case that if for all $\argB \in \args$,
$\Pr(\argB) < 1$, then $\Pr(\argA) > 0$.

Let
$$\argA
= \argu{\{\sent_0, \neg \sent'_1, \ldots, \neg \sent'_n\}}{\sent_0},$$
\st{} $\args
= \{\argu{\_}{\sent'_1}, \ldots, \argu{\_}{\sent'_n}\}$. We 
see that 
$$\Pr(\argA) \leq
min(\Pr(\neg \sent'_1), \ldots, \Pr(\neg \sent'_n)).$$
Since $\Pr(\argB) < 1$ for all $\argB \in \args$,
$min(\Pr(\sent'_1), \ldots, \Pr(\sent'_n)) > 0$. By
Lemma~\ref{lemma:MEomega}, we know that $\Pr(\argA) > 0$.
Therefore, $\Pr(\argA) = 0$ only if these exists at least one
$\argB \in \args$ \st{} $\Pr(\argB) = 1$. 

(2) follows from Proposition~\ref{prop:zeroIfAttackedByOne}
directly. \qed

\paragraph{Proof of Theorem~\ref{thm:probCompleteLabel}}

This follows directly from the Proposition 2 in \cite{Baroni11},
Propositions~\ref{prop:oneIfAttackedByZero}
and \ref{prop:arg0Attacker1}. \qed

\paragraph{Proof of Proposition~\ref{prop:optimistic}}

Let $\argA
= \argu{\{\sent^a, \neg \sent^b_1, \ldots, \neg \sent^b_n\}}{\sent^a}$,
$\args
= \{\argB_1 = \argu{\_}{\sent^b_1}, \ldots, \argB_n 
= \argu{\_}{\sent^b_n}\}$, with $F$ containing the following p-rules:
\begin{center}
\prule{\sent^a \gets \neg \sent^b_1, \ldots, \neg \sent^b_n}{1},
\prule{\sent^b_1 \gets \_}{1}, 
\ldots,
\prule{\sent^b_n \gets \_}{1}.
\end{center}

This proposition is to show that 
$$Pr(\argA) + Pr(\argB_1) + \ldots + Pr(\argB_n) \geq 1.$$
Assume otherwise, then there exists $\omega \in \Omega$, $\Pr(\omega)
> 0$, \st{} 
$$\omega \not\models \argA, \omega \not\models \argB_1, \ldots, \omega \not\models \argB_n.$$
Through a case analysis, we show this is not possible.

\begin{itemize}
\item
Case 1: $\omega \models \sent^b_i, i \in \{1,\ldots,n\}$. By P-CWA,
either $\omega \models \argB_i$ or $\Pr(\omega) = 0$. 
Both contradict to the proposition. 

\item
Case 2: $\omega \models \neg \sent^b_i, i \in \{1,\ldots,n\}$. Then,
there are two sub-cases as follows.

\begin{itemize}
\item
Case 2a: if $\omega \models \sent^a$, then, by P-CWA, either
$\omega \models \argA$ or $\Pr(\omega) = 0$.

\item
Case 2b: if $\omega \not\models \sent^a$, then either

\begin{itemize}
\item
case 2b(i):
$\omega \models \neg \sent^a \wedge \neg \sent^b_1 \wedge \ldots \wedge \neg \sent^b_n$,
then $\Pr(\omega) = 0$ as from the p-rule
$$\prule{\sent^a \gets \neg \sent^b_1, \ldots, \sent^b_n}{1}$$
we have
$$\frac{\Pr(\sent^a \wedge \neg \sent^b_1 \wedge \ldots \wedge \neg \sent^b_n)}
{\Pr(\neg \sent^b_1 \wedge \ldots \wedge \neg \sent^b_n)} = 1,$$
so
$\Pr(\neg \sent^a \wedge \neg \sent^b_1 \wedge \ldots \wedge \neg \sent^b_n)
= 0$. 

\item
case 2b(ii): there exists $\sent^b_j$ \st{}
$\omega \models \neg \sent^a \wedge \ldots \wedge \sent^b_j \wedge \ldots$. In
this case, $\omega \models \sent^b_j$, which is Case 1 with $j$ in 
place of $i$.
\end{itemize}
\end{itemize}
\end{itemize}
Therefore, in all cases, either $\omega \models \argA$ or
$\omega \models \argB_i$ for some $\argB_i \in \args$ or $\Pr(\omega)
= 0$. \qed 

\paragraph{Proof of Theorem~\ref{thm:Rule-PSAT}}

  (Sketch.) Equations~\ref{eqn:dfnq1} to \ref{eqn:dfnq4} are satisfied
  by a $\pi$ solution in $[0,1]^{2^n}$ as follows.
  \begin{enumerate}
  \item
    If $\pi \in [0,1]^{2^n}$, then $0 \leq \pi(\omega_i) \leq 1$ for
    all $\omega_i$.
  \item
    Since Row $m+1$ in $A$ and $B$ are 1s, we have the sum of all
    $\pi(\omega_i)$s being 1.
  \item
    For each p-rule \prule{\sent_0 \gets}{\probParameter},
    Equations~\ref{eqn:AijHead} and \ref{eqn:Bi} ensure that
    Equation~\ref{eqn:dfnq3} is satisfied.
    
  \item
    For each p-rule \prule{\sent_0 \gets
      \sent_1,\ldots,\sent_k}{\probParameter},
    Equation~\ref{eqn:AijBody} and \ref{eqn:Bi0} ensure that
    Equation~\ref{eqn:dfnq4} is satisfied with simple algebra. 
  \end{enumerate}
  Thus, we see that Equation~\ref{eqn:APiB}, $A\pi = B$, is nothing
  but a linear system representation of
  Equations~\ref{eqn:dfnq1}-\ref{eqn:dfnq4}, which characterise
  probability distributions over the CC set of $\mc{L}_0$ with
  conditionals. \qed

\paragraph{Proof of Theorem~\ref{thm:localForGlobal}}

The key to this proof is on showing that the ``local'' constraints
given in Equations~\ref{eqn:AijPCWA} and \ref{eqn:BiPCWA} are
correct. In other words, we must generalize
Examples~\ref{exp:localForGlobal} and \ref{exp:localPCWAMultiRules} to
consider p-rule sets composed of arbitrary numbers of p-rules. To this
end, we use proof by induction.

Consider a language $\mc{L}$ and a set of p-rules $\mc{R}$ defined
with $\mc{L}$. Consider $\mc{R}$ as the union of $m$ sets of p-rules,
$$\mc{R} = \bigcup_{i=1}^m R^{(i)},$$
in which each set $R^{(i)} = \{\rho^{(i)}_1, \ldots, \rho^{(i)}_r\}$
contains $r$ p-rules with the same head
$\sent^{(i)}$\footnote{Although $r$ is parametrized on $(i)$, it is
omitted to simplify the notation. We do not force that all literals
which are heads of rules are heads of the same number of rules.}
\begin{center}
$\rho^{(i)}_1
= \prule{\sent^{(i)} \gets \sent_1^{(i),1}, \ldots, \sent_{r1}^{(i),1}}{\cdot},$ \\
\ldots \\
$\rho^{(i)}_r
= \prule{\sent^{(i)} \gets \sent_1^{(i),r}, \ldots, \sent_{rr}^{(i),r}}{\cdot}.$
\end{center}
From p-rules, we define
\begin{align*}
L(\sent^{(i)}) &=
(\sent^{(i)} \wedge \bigwedge\limits_{j=1}^{r1} \sent_j^{(i),1}) \vee \ldots \vee
(\sent^{(i)} \wedge \bigwedge\limits_{j=1}^{rr} \sent_j^{(i),r}), \\
f_L(\sent^{(i)}) &= \{\omega \in \Omega
| \omega \models \sent^{(i)}, \omega \not \models L(\sent^{(i)})\}, \\
\Omega_L^{(i)} &= \bigcup_{j=1}^i f_L(\sent^{(j)}).
\end{align*}
$L(\sent^{(i)})$ describes the local constraint defined by
p-rules with head $\sent^{(i)}$. $f_L(\sent^{(i)})$ is the set of
atomic conjunctions $\omega$ which are set to have $\Pr(\omega) = 0$
by considering p-rules with head $\sent^{(i)}$, and $\Omega_L^{(i)}$
the set of $\omega$s which are set to have $\Pr(\omega) = 0$ by
considering all p-rules with heads
$\sent^{(1)}, \ldots, \sent^{(i)}$.

Consider ``global'' constraints, we have deductions built from
p-rules. For each literal $\sent^{(i)}$, which is the head
of a rule, we define a set of deductions $D^{(i)}
= \{\delta_1^{(i)}, \ldots, \delta_d^{(i)}\}$, in which 
\begin{center}
$\delta_1^{(i)}
= \arguD{\{\sent^{(i)}, \sent_1^{(i),1}, \ldots, \sent_{d1}^{(i),1}\}}{\sent^{(i)}}$, \\
\ldots \\
$\delta_d^{(i)}
= \arguD{\{\sent^{(i)}, \sent_1^{(i),d}, \ldots, \sent_{dd}^{(i),d}\}}{\sent^{(i)}}$.\footnote{As
in the previous footnote, although $d$ is parametrized on $(i)$, it
is omitted to simplify the notation. We do not force all literals
which are claims of deductions having the same number of deductions.} 
\end{center}
From $\delta_1^{(i)},\ldots,\delta_r^{(i)}$, we define
\begin{align*}
G(\sent^{(i)}) &=
(\sent^{(i)} \wedge \bigwedge\limits_{j=1}^{d1} \sent_j^{(i),1}) \vee \ldots \vee
(\sent^{(i)} \wedge \bigwedge\limits_{j=1}^{dd} \sent_j^{(i),d}), \\
f_G(\sent^{(i)}) &= \{\omega \in \Omega | \omega \models \sent^{(i)},
\omega \not \models G(\sent^{(i)})\}, \\
\Omega_G^{(i)} &= \bigcup_{j=1}^i f_G(\sent^{(j)}) .
\end{align*}
$f_G(\sent^{(i)})$ is the set of atomic conjunctions
$\omega$, which are set to have $\Pr(\omega) = 0$ by considering
deductions for $\sent^{(i)}$, and $\Omega_G^{(i)}$ the set of
$\omega$s which are set to have $\Pr(\omega) = 0$ by considering all
deductions for $\sent^{(1)}, \ldots, \sent^{(i)}$. These are the
atomic conjunctions obtained from ``global'' constraints.

For the base case, $i=1$, by the constructions of $R^{(1)}$ and
$D^{(1)}$, which give $L(\sent^{(1)})$ and $G(\sent^{(1)})$, we see
that $$\Omega_L^{(1)} = \Omega_G^{(1)}.$$

Assume that $\Omega_L^{(n)} = \Omega_G^{(n)}$, we show it is the case
that $\Omega_L^{(n+1)} = \Omega_G^{(n+1)}$. We have
\begin{align*}
L(\sent^{(n+1)}) &=
(\sent^{(n+1)} \wedge \bigwedge\limits_{j=1}^{r1} \sent_j^{(n+1),1}) \vee \ldots \vee
(\sent^{(n+1)} \wedge \bigwedge\limits_{j=1}^{rr} \sent_j^{(n+1),r}), \\
G(\sent^{(n+1)}) &=
(\sent^{(n+1)} \wedge \bigwedge\limits_{j=1}^{d1} \sent_j^{(n+1),1}) \vee \ldots \vee
(\sent^{(n+1)} \wedge \bigwedge\limits_{j=1}^{dd} \sent_j^{(n+1),d}), \\
f_L(\sent^{(n+1)}) &= \{\omega \in \Omega
| \omega \models \sent^{(n+1)}, \omega \not \models L(\sent^{(n+1)})\}, \\
f_G(\sent^{(n+1)}) &= \{\omega \in \Omega | \omega \models \sent^{(n+1)},
\omega \not \models G(\sent^{(n+1)})\}.
\end{align*}

By the definition of deduction, $r \leq d$ and $r1 \leq d1$, $r2 \leq
d2$, \ldots. It is easy to see that for each $\sent^{(n+1)}$, it
holds that $$f_L(\sent^{(n+1)}) \subseteq f_G(\sent^{(n+1)}).$$ Thus, to
show $\Omega_L^{(n+1)} = \Omega_G^{(n+1)}$, we need to show that when 
$f_L(\sent^{(n+1)}) \subset f_G(\sent^{(n+1)})$, i.e., when there exists
$\omega \in f_G(\sent^{(n+1)})$ and $\omega \not\in
f_L(\sent^{(n+1)})$, there exists 
$\sent^+ \in \{\sent^{(1)}, \ldots,\sent^{(n)}\}$ \st{} 
\begin{center}
$\omega \in f_L(\sent^+)$ therefore $\omega \in \Omega_L^{(n)}$.
\end{center}
We observe that $\omega^* \in f_G(\sent^{(n+1)}) \setminus
f_L(\sent^{(n+1)})$ is of the form

$$\omega^*
= \sent^{(n+1)} \wedge \sent_1^{(n+1),k} \wedge \ldots \wedge \sent_{rk}^{(n+1),k} \wedge \neg \sent^*_1 \wedge \ldots \wedge \neg \sent^*_e \ldots$$ 
\st{}
\begin{enumerate}
\item
$\prule{\sent^{(n+1)} \gets \sent_1^{(n+1),k}, \ldots, \sent_{rk}^{(n+1),k}}{\cdot} \in
R^{(n+1)}$, and
\item
for all $\delta \in D^{(n+1)}$, there exists
$\sent^* \in \{\sent^*_1, \ldots, \sent^*_e\}$ \st{} $\sent^*$ is in
$\delta$.
\end{enumerate}
In other words, $\omega^* \models \sent^{(n+1)}, \omega^* \models
L(\sent^{(n+1)})$, and $\omega^* \not\models
G(\sent^{(n+1)})$. Therefore, 
There does not exist $\arguD{\sents}{\sent^{(n+1)}} \in D^{(n+1)}$
\st{}
\begin{center}
$\omega^* \models \bigwedge\limits_{\sent \in \sents} \sent$.
\end{center}
However, for such $\omega^*$, there must exist
$\sent^+ \in \{\sent_1^{(n+1),k}, \ldots, \sent_{rk}^{(n+1),k}\}$ \st{}
the following two conditions C1 and C2 are met:
\begin{itemize}
\item
{\bf (C1)}: $\omega^* \models \sent^+$, and

\item
{\bf (C2)}: $\omega^* \not\models L(\sent^+)$.
\end{itemize}
Therefore $\omega^* \in f_L(\sent^+)$.
Moreover, since
$\{\sent_1^{(n+1),k}, \ldots, \sent_{rk}^{(n+1),k}\} \subseteq \{\sent^{(1)},\ldots,\sent^{(n)}\}$,
we have $\sent^+ \in \{\sent^{(1)},\ldots,\sent^{(n)}\}$. Thus,
$$\omega^* \in \Omega_L^{(n)}.$$

Assume otherwise, i.e.,
\begin{center}
$\omega^* \models L(\sent_1^{(n+1),k})$, \ldots,
$\omega^* \models L(\sent_{rk}^{(n+1),k})$,
\end{center}
then there exists deductions
\begin{center}
$\arguD{\sents_1}{\sent_1^{(n+1),k}}, \ldots, \arguD{\sents_k}{\sent_{rk}^{(n+1),k}}$
\end{center}
\st{}
$$\omega^* \models \bigwedge\limits_{\sent \in \sents_1 \cup \ldots \cup
\sents_k} \sent.$$
Since $\omega^* \models \sent^{(n+1)}$, we also have
$$\omega^* \models \bigwedge\limits_{\sent \in \sents_1 \cup \ldots \cup
\sents_k \cup \{\sent^{(n+1)}\}} \sent.$$

Since there is a p-rule
$\prule{\sent^{(n+1)} \gets \sent_1^{(n+1),k}, \ldots, \sent_{rk}^{(n+1),k}}{\cdot} \in
R^{(n+1)}$, we have
$$\arguD{\sents_1 \cup \ldots \cup \sents_k \cup \{\sent^{(n+1)}\}}{\sent^{(n+1)}} \in
D^{(n+1)}.$$
Contradiction.

Therefore, there exists
$\sent^+ \in \{\sent_1^{(n+1),k}, \ldots, \sent_{rk}^{(n+1),k}\}$
meeting conditions {\bf C1} and {\bf C2}. Therefore
$\omega^* \in \Omega_L^{(n)}$ and $\Omega_L^{(n+1)}
= \Omega_G^{(n+1)}$. \qed

%% file: arXiv.bbl
\begin{thebibliography}{10}

\bibitem{Baroni11}
P.~Baroni, M.~Caminada, and M.~Giacomin.
\newblock An introduction to argumentation semantics.
\newblock {\em Knowl. Eng. Rev.}, 26(4):365--410, 2011.

\bibitem{Baroni2018}
P.~Baroni, D.~Gabbay, M.~Giacomin, and L.~Van~der Torre.
\newblock {\em Handbook of formal argumentation}.
\newblock College Publications, 2018.

\bibitem{tutor}
P.~Besnard, A.~Garcia, A.~Hunter, S.~Modgil, H.~Prakken, G.~Simari, and
  F.~Toni.
\newblock Special issue: Tutorials on structured argumentation.
\newblock {\em Argument \& Computation}, 5(1), 2014.

\bibitem{Beveridge70}
G.S.G. Beveridge, S.G. Beveridge, and R.S. Schechter.
\newblock {\em Optimization: Theory and Practice}.
\newblock Chemical Engineering Series. McGraw-Hill, 1970.

\bibitem{bongiovanni2018}
G.~Bongiovanni, G.~Postema, A.~Rotolo, G.~Sartor, C.~Valentini, and D.~Walton.
\newblock {\em Handbook of Legal Reasoning and Argumentation}.
\newblock Springer Netherlands, 2018.

\bibitem{Buscemi07}
F.~Buscemi, P.~Bordone, and A.~Bertoni.
\newblock Linear entropy as an entanglement measure in two-fermion systems.
\newblock {\em Physical Review A}, 75(3), mar 2007.

\bibitem{Caminada17}
M.~Caminada.
\newblock Argumentation semantics as formal discussion.
\newblock {\em {FLAP}}, 4(8), 2017.

\bibitem{Caminada09-labelling}
M.~Caminada and D.~Gabbay.
\newblock {A Logical Account of Formal Argumentation}.
\newblock {\em Studia Logica}, 93(2):109--145, December 2009.

\bibitem{Craven12}
R.~Craven, F.~Toni, C.~Cadar, A.~Hadad, and M.~Williams.
\newblock Efficient argumentation for medical decision-making.
\newblock In {\em Proc. of AAAI}. {AAAI} Press, 2012.

\bibitem{Cyras18}
K.~{\v Cyras}, B.~Delaney, D.~Prociuk, F.~Toni, M.~Chapman,
  J.~Dom{\'{\i}}nguez, and V.~Curcin.
\newblock Argumentation for explainable reasoning with conflicting medical
  recommendations.
\newblock In {\em Proc. of MedRACER 2018}, pages 14--22. CEUR-WS.org, 2018.

\bibitem{Cyras17}
K.~{\v Cyras}, X.~Fan, C.~Schulz, and F.~Toni.
\newblock Assumption-based argumentation: Disputes, explanations, preferences.
\newblock {\em IfCoLog JLTA}, 4(8), 2017.

\bibitem{Cyras21}
K.~{\v Cyras}, Q.~Heinrich, and F.~Toni.
\newblock Computational complexity of flat and generic assumption-based
  argumentation, with and without probabilities.
\newblock {\em Artificial Intelligence}, 293:103449, 2021.

\bibitem{Cyras0ABT21}
K.~{\v Cyras}, A.~Rago, E.~Albini, P.~Baroni, and F.~Toni.
\newblock Argumentative {XAI:} {A} survey.
\newblock In {\em Proc. of IJCAI}, pages 4392--4399. ijcai.org, 2021.

\bibitem{philip79}
A.~P. Dawid.
\newblock Conditional independence in statistical theory.
\newblock {\em Journal of the Royal Statistical Society: Series B
  (Methodological)}, 41(1):1--15, 1979.

\bibitem{Doder14}
D.~Doder and S.~Woltran.
\newblock Probabilistic argumentation frameworks - {A} logical approach.
\newblock In {\em Proc. of SUM}, pages 134--147. Springer, 2014.

\bibitem{Dondio14}
P.~Dondio.
\newblock Multi-valued and probabilistic argumentation frameworks.
\newblock In {\em Proc. of COMMA}, volume 266, pages 253--260. {IOS} Press,
  2014.

\bibitem{Dung95}
P.~M. Dung.
\newblock On the acceptability of arguments and its fundamental role in
  nonmonotonic reasoning, logic programming and n-person games.
\newblock {\em Artificial Intelligence}, 77(2):321--357, 1995.

\bibitem{Dung10}
P.M. Dung and P.M. Thang.
\newblock Towards (probabilistic) argumentation for jury-based dispute
  resolution.
\newblock In {\em Proc. of COMMA}, pages 171--182. IOS Press, 2010.

\bibitem{Eemeren17}
F.~H.~Van Eemeren and B.~Verheij.
\newblock Argumentation theory in formal and computational perspective.
\newblock {\em {FLAP}}, 4(8), 2017.

\bibitem{Fan22}
X.~Fan.
\newblock Rule-psat: Relaxing rule constraints in probabilistic
  assumption-based argumentation.
\newblock In {\em Proc. of COMMA}, 2022.

\bibitem{Fan13-CLIMA}
X.~Fan, R.~Craven, R.~Singer, F.~Toni, and M.~Williams.
\newblock Assumption-based argumentation for decision-making with preferences:
  A medical case study.
\newblock In {\em Proc. of CLIMA}, pages 374--390, 2013.

\bibitem{FazzingaFP16}
B.~Fazzinga, S.~Flesca, and F.~Parisi.
\newblock On efficiently estimating the probability of extensions in abstract
  argumentation frameworks.
\newblock {\em Int. J. Approx. Reason.}, 69:106--132, 2016.

\bibitem{Fazzinga16}
B.~Fazzinga, S.~Flesca, F.~Parisi, and A.~Pietramala.
\newblock Computing or estimating extensions' probabilities over structured
  probabilistic argumentation frameworks.
\newblock {\em {FLAP}}, 3(2):177--200, 2016.

\bibitem{Fox07}
J.~Fox, D.~Glasspool, D.~Grecu, S.~Modgil, M.~South, and V.~Patkar.
\newblock Argumentation-based inference and decision making--a medical
  perspective.
\newblock {\em IEEE Intelligent Systems}, 22(6):34--41, 2007.

\bibitem{Gainsburg2016}
J.~Gainsburg, J.~Fox, and L.~M. Solan.
\newblock Argumentation and decision making in professional practice.
\newblock {\em Theory Into Practice}, 55(4):332--341, 2016.

\bibitem{Georgakopoulos88}
G.~F. Georgakopoulos, D.~J. Kavvadias, and C.~H. Papadimitriou.
\newblock Probabilistic satisfiability.
\newblock {\em Journal of Complexity}, 4(1):1--11, 1988.

\bibitem{Hansen90}
P.~Hansen and B.~Jaumard.
\newblock Algorithms for the maximum satisfiability problem.
\newblock {\em Computing}, 44(4):279--303, 1990.

\bibitem{Henderson20}
T.~C. Henderson, R.~Simmons, B.~Serbinowski, M.~Cline, D.~Sacharny, X.~Fan, and
  A.~Mitiche.
\newblock Probabilistic sentence satisfiability: An approach to {PSAT}.
\newblock {\em Artificial Intelligence}, 278, 2020.

\bibitem{Hung17}
N.~D. Hung.
\newblock Inference procedures and engine for probabilistic argumentation.
\newblock {\em International Journal of Approximate Reasoning}, 90:163--191,
  2017.

\bibitem{Hunter12}
A.~Hunter.
\newblock Some foundations for probabilistic abstract argumentation.
\newblock In {\em Proc. of COMMA}, volume 245, pages 117--128. {IOS} Press,
  2012.

\bibitem{Hunter13}
A.~Hunter.
\newblock A probabilistic approach to modelling uncertain logical arguments.
\newblock {\em International Journal Approximate Reasoning}, 2013.

\bibitem{Hunter20}
A.~Hunter.
\newblock Reasoning with inconsistent knowledge using the epistemic approach to
  probabilistic argumentation.
\newblock In {\em Proc. of KR}, pages 496--505, 2020.

\bibitem{Hunter22}
A.~Hunter.
\newblock Argument strength in probabilistic argumentation based on defeasible
  rules.
\newblock {\em Int. J. Approx. Reason.}, 146:79--105, 2022.

\bibitem{HunterL10}
A.~Hunter and W.~Liu.
\newblock A survey of formalisms for representing and reasoning with scientific
  knowledge.
\newblock {\em Knowl. Eng. Rev.}, 25(2):199--222, 2010.

\bibitem{Hunter20b}
A.~Hunter, S.~Polberg, and M.~Thimm.
\newblock Epistemic graphs for representing and reasoning with positive and
  negative influences of arguments.
\newblock {\em Artificial Intelligence}, 281:103236, 2020.

\bibitem{Hunter14}
A.~Hunter and M.~Thimm.
\newblock Probabilistic argumentation with incomplete information.
\newblock In {\em Proc. of ECAI}, pages 1033--1034. {IOS} Press, 2014.

\bibitem{Hunter17}
A.~Hunter and M.~Thimm.
\newblock Probabilistic reasoning with abstract argumentation frameworks.
\newblock {\em J. Artif. Intell. Res.}, 59:565--611, 2017.

\bibitem{Jaynes57}
E.~T. Jaynes.
\newblock Information theory and statistical mechanics.
\newblock {\em Phys. Rev.}, 106:620--630, May 1957.

\bibitem{jaynes2003}
E.T. Jaynes and G.L. Bretthorst.
\newblock {\em {Probability Theory: The Logic of Science}}.
\newblock Cambridge University Press, 2003.

\bibitem{Kafer22}
N.~K{\"{a}}fer, C.~Baier, M.~Diller, C.~Dubslaff, S.~Alice Gaggl, and
  H.~Hermanns.
\newblock Admissibility in probabilistic argumentation.
\newblock {\em J. Artif. Intell. Res.}, 74, 2022.

\bibitem{Kokciyan2021}
N.~K{\"o}kciyan, I.~Sassoon, E.~Sklar, S.~Modgil, and S.~Parsons.
\newblock Applying metalevel argumentation frameworks to support medical
  decision making.
\newblock {\em IEEE Intelligent Systems}, 36(2):64--71, 2021.

\bibitem{Labrie2014}
N.~Labrie and P.~J. Schulz.
\newblock Does argumentation matter? a systematic literature review on the role
  of argumentation in doctor--patient communication.
\newblock {\em Health communication}, 29(10):996--1008, 2014.

\bibitem{Li11}
H.~Li, N.~Oren, and T.~Norman.
\newblock Probabilistic argumentation frameworks.
\newblock In {\em Proc. of TAFA}, 2011.

\bibitem{Ma10}
J.~Ma, W.~Liu, and A.~Hunter.
\newblock Inducing probability distributions from knowledge bases with
  (in)dependence relations.
\newblock In {\em Proc. of AAAI}. {AAAI} Press, 2010.

\bibitem{Mitchell97}
T.~M. Mitchell.
\newblock {\em Machine Learning}.
\newblock McGraw-Hill, Inc., New York, NY, USA, 1 edition, 1997.

\bibitem{Nilsson93}
N.~Nilsson.
\newblock Probabilistic logic revisited.
\newblock {\em Artificial Intelligence}, 59(1-2):39--42, 1993.

\bibitem{Nilsson86}
N.~J. Nilsson.
\newblock Probabilistic logic.
\newblock {\em Artificial Intelligence}, 28(1):71--87, 1986.

\bibitem{paris1994}
J.~B. Paris.
\newblock {\em The uncertain reasoner's companion: a mathematical perspective}.
\newblock Cambridge University Press, 1994.

\bibitem{Polberg14}
S.~Polberg and D.~Doder.
\newblock Probabilistic abstract dialectical frameworks.
\newblock In Eduardo Ferm{\'{e}} and Jo{\~{a}}o Leite, editors, {\em Proc. of
  JELIA}, volume 8761, pages 591--599. Springer, 2014.

\bibitem{Reiter77}
R.~Reiter.
\newblock On closed world data bases.
\newblock In {\em Logic and Data Bases, Symposium on Logic and Data Bases,
  Centre d'{\'{e}}tudes et de recherches de Toulouse, France, 1977}, pages
  55--76, New York, 1977. Plemum Press.

\bibitem{Reiter81}
R.~Reiter and G.~Criscuolo.
\newblock On interacting defaults.
\newblock In {\em Proc. of IJCAI}, pages 270--276. William Kaufmann, 1981.

\bibitem{Rienstra12}
T.~Rienstra.
\newblock Towards a probabilistic dung-style argumentation system.
\newblock In {\em Proc. AT}, volume 918, pages 138--152. CEUR-WS.org, 2012.

\bibitem{Russell2009}
S.~Russell and P.~Norvig.
\newblock {\em Artificial Intelligence: A Modern Approach}.
\newblock Prentice Hall Press, Upper Saddle River, NJ, USA, 3rd edition, 2009.

\bibitem{Sun15b}
X.~Sun and B.~Liao.
\newblock Probabilistic argumentation, a small step for uncertainty, a giant
  step for complexity.
\newblock In {\em Proc. of EUMAS}, pages 279--286. Springer, 2015.

\bibitem{Thimm12}
M.~Thimm.
\newblock A probabilistic semantics for abstract argumentation.
\newblock In {\em Proc. of ECAI}, 2012.

\bibitem{Toni14}
F.~Toni.
\newblock A tutorial on assumption-based argumentation.
\newblock {\em Argument \& Computation, Special Issue: Tutorials on Structured
  Argumentation}, 5(1):89--117, 2014.

\bibitem{Torre17}
L.~V.~D. Torre and S.~Vesic.
\newblock The principle-based approach to abstract argumentation semantics.
\newblock {\em {FLAP}}, 4(8), 2017.

\bibitem{Ulbricht19}
M.~Ulbricht and R.~Baumann.
\newblock If nothing is accepted - repairing argumentation frameworks.
\newblock {\em J. Artif. Intell. Res.}, 66:1099--1145, 2019.

\bibitem{Williamson02}
J.~Williamson.
\newblock {\em Handbook of the Logic of Argument and Inference: the Turn Toward
  the Practical}, chapter Probability Logic, pages 397--424.
\newblock Elsevier, 2002.

\bibitem{WilsonLopez20}
A.~Wilson-Lopez, A.~R. Strong, C.~M. Hartman, J.~Garlick, K.~H. Washburn,
  A.~Minichiello, S.~Weingart, and J.~Acosta-Feliz.
\newblock A systematic review of argumentation related to the
  engineering-designed world.
\newblock {\em Journal of Engineering Education}, 109(2):281--306, 2020.

\bibitem{Yuan15}
Y.~Yuan.
\newblock Recent advances in trust region algorithms.
\newblock {\em Math. Program.}, 151(1):249--281, 2015.

\end{thebibliography}
